\documentclass[11pt,a4paper]{article}
\usepackage[margin=2.3cm]{geometry}
\usepackage{array}
\usepackage[T1]{fontenc}
\usepackage{lmodern}
\usepackage{amsmath,amssymb,amsthm,bm}
\usepackage{graphicx}
\usepackage{booktabs,multirow,rotating}
\usepackage{siunitx}
\usepackage{natbib}
\usepackage{tikz}
\usetikzlibrary{arrows.meta,positioning,calc,fit}
\usepackage{xcolor}
\usepackage[hidelinks]{hyperref}
\usepackage{longtable}
\usepackage{caption}
\usepackage[section]{placeins} %

\graphicspath{{figs/}}
\setcitestyle{authoryear,round,aysep={,}}
\newcommand{\CaRT}{CaRT}

\newcommand{\CaRTp}{CaRT$^{+}$}
\newcommand{\dd}{\mathrm{d}}
\newcommand{\sg}{\operatorname{sig}}

\DeclareMathOperator{\Var}{Var}
\DeclareMathOperator{\Cov}{Cov}

\newtheorem{law}{Law}
\newtheorem{corollary}{Corollary}
\newcommand{\cA}{\mathcal{A}}
\newcommand{\ein}{e^{\rm in}}
\newcommand{\cf}{c_m}

\title{\textbf{From hydrological laws to catchment models: A process-based model matches deep-learning accuracy on three continents}}
\author{Xilin Xia$^{1}$ \\[2pt]
\small $^{1}$School of Engineering, University of Birmingham, Birmingham, UK \\
\small x.xia.1@bham.ac.uk}
\date{4 November 2026}

\begin{document}
\maketitle

\begin{abstract}
Dooge asked whether hydrology has laws at the catchment scale. We ask whether a low-dimensional system built under physical laws can match the accuracy of neural networks that outperform process-based models on the same inputs. Three laws constrain a lumped catchment model. Conservation makes it a flux-connected storage network, dissipation fixes the sign of every flux and aggregation determines how the unresolved composition enters the fluxes, through effective laws over threshold distributions and a memory of the forcing. These laws admit four kinds of refinement. A parsimonious model under them, \CaRT{} (11 parameters per catchment), is refined in a loop assisted by AI agents. Each refinement addresses an observed deficit through a hypothesis from the literature. The result, \CaRTp, has fourteen states, 35 parameters per catchment and no daily neural network. Most refinements follow from aggregation over thresholds or from variables that carry no water but modulate fluxes. On 602 British and 593 US gauged catchments, single \CaRTp{} models reach median Nash--Sutcliffe efficiencies of 0.863 and 0.765 in held-out years. A stateful LSTM reaches 0.869 and 0.751 and an entity-aware LSTM 0.862 and 0.754, while three-seed LSTM ensembles remain ahead of the \CaRTp{} ensemble by 0.003--0.012 in median paired NSE. On 505 drier Australian gauged catchments unused in development, single \CaRTp{} models reach 0.778 against 0.762 and 0.753 for the two LSTMs. The results suggest that catchment-scale laws exist as laws of form, with the constitutive families shared and only their parameters learned.
 \end{abstract}

\section*{Plain Language Summary}
Predicting how much water a river carries after rain is a long-standing problem in hydrology. The physics of water moving through soil and over hillslopes is well understood at small scales, yet no equally clear description exists for a whole catchment, where these processes combine over varied soils, slopes and vegetation. Researchers have long debated whether laws exist at the catchment scale at all. The debate sharpened when deep-learning models, trained only on rainfall, temperature and a few catchment properties, began to predict streamflow much better than models built from physical processes, which suggests that those models miss something. This paper derives three laws for catchment models. Water must be conserved, flow must run down gradients and lose energy, and when varied land is averaged into a few stores, the variability that is lost reappears as a memory of past weather, which a few additional variables can carry. These laws decide what a catchment model may look like and how it may be improved. Within these constraints, AI agents helped construct and test refinements of a simple model. The resulting model, with no neural network inside, matches deep learning on about 1900 catchments in Britain, the United States and Australia.

\noindent\textbf{Key Points}
\begin{itemize}\setlength\itemsep{0pt}
\item Conservation, dissipation and aggregation fix the structure of a lumped catchment model and define four admissible ways to refine it.

\item Most refinements of the low-dimensional model follow from threshold aggregation or from fluxes modulated by variables that carry no water.

\item One structure with 35 parameters per catchment matches the single-model median NSE of two same-input LSTMs on three continents.

\end{itemize}

\section{Introduction}\label{sec:intro}
Predictions of daily streamflow are essential for flood-risk assessment and water-resources planning. The processes that generate streamflow are understood in detail at the scale of a soil column or a hillslope, yet models built from these processes have not matched the predictive
accuracy of neural networks that represent none of them explicitly \citep{lees2022}. Are processes missing from our understanding, or are we missing the laws by which known processes
combine at the catchment scale?

Findings of the last decade frame this question. Large-sample datasets such as CAMELS and CAMELS-GB \citep{addor2017camels,newman2015,coxon2020camelsgb,addor2020large,kratzert2023caravan} made it possible to test rainfall--runoff models on hundreds of catchments at once. On these samples deep-learning models trained jointly across catchments outperform conceptual and process-based models (PBMs) by a wide margin \citep{kratzert2018,kratzert2019wrr,kratzert2019hess,lees2021,mai2022gripgl,nearing2024nature}. These models use nothing beyond daily catchment-mean forcing and a handful of static attributes, and \citet{nearing2021wrr} concluded that large-sample observations contain more information about catchment behaviour than hydrological theory has used. Attempts to locate the missing ingredient in a relaxed water balance \citep{frame2022extremes,frame2023} or the training protocol \citep{kratzert2024neverTrain} account for small parts of the gap, and the gap is not an artefact of the metric \citep{gauch2023metrics}. A mass-conserving LSTM comes within 0.02 in median NSE of the LSTM \citep{hoedt2021mclstm}. Differentiable models, in which neural networks supply the parameters of a conceptual structure \citep{tsai2021,shen2023,song2025}, have approached the accuracy of the LSTM \citep{feng2022,feng2023}, as have neural ODEs in which networks replace the fluxes of such a structure \citep{hoge2022}. On CAMELS-US, a differentiable HBV comes within 0.02 NSE of the LSTM \citep{feng2022}, but that model is sixteen copies of HBV run in parallel and averaged, and most of the gain comes from the mixture. Daily parameters from the network add only 0.018 in median NSE to it \citep{feng2022}, and sixteen copies with static parameters from a multilayer perceptron score 0.71 where one copy scores 0.63 \citep{poudel2026}. However, a mixture of sixteen models is no longer a low-dimensional structure. The networks that parameterise such hybrids fit deliberately damaged structures almost as well \citep{acuna2024}. On CAMELS-GB the conceptual constraint adds little to what the network has to predict and can make the prediction harder \citep{alvarezchaves2026physics}. Such hybrids show that the inputs carry the information but not which processes a single structure would need to extract it \citep{kirchner2006}.

These findings suggest that whatever current PBMs miss is a function of the inputs they already receive, and that a flexible parameterisation can supply it. It is open whether one fixed, low-dimensional structure with static parameters can supply it and what the structure must then contain. This points to the form of the laws that map storage and forcing to fluxes, and revives the old question of whether there are hydrological laws at the catchment scale \citep{dooge1986,sivapalan2003,sivapalan2005,mcdonnell2007,bloschl2019} or whether every catchment is unique \citep{beven2000,beven2020dl}. \citet{kirchner2009} showed that some catchments behave as first-order dynamical systems, whereas the balance equations of representative elementary watersheds need closure relations that are nonlinear, hysteretic and scale dependent \citep{reggiani1998,reggiani1999,beven2006}. \citet{clark2008,fenicia2011,knoben2019} developed modular frameworks assemble conceptual models from interchangeable stores and fluxes. Stepwise improvement guided by diagnostics \citep{fenicia2008}, genetic programming over such components \citep{matin2026hydrologically} and mass-conserving networks trained for parsimony \citep{wang2025using} could be used to search model structures within the assemblies. Agents built on large language models have been proposed to assist this search \citep{eythorsson2025toward}. However, which assemblies are admissible, and which component is needed for a given deficit, remain open.

We therefore take up the question of \citet{dooge1986}, and his search for simplicity \citep{dooge1997searching,dooge2005bringing}, in a form that large samples of catchments can test. We ask whether a low-dimensional system built under physical laws can predict daily streamflow as accurately as a neural network trained on the same inputs. We use the neural network as a measure of the information in the inputs. If such a system exists, the laws that produce it are candidates for the catchment-scale laws Dooge looked for. If it does not, either the laws are incomplete or the catchment response has no low-dimensional form. This work provides evidence of the existence of such a low-dimensional system. Section~\ref{sec:laws} states three laws, conservation, dissipation and aggregation, from which the admissible refinements of a lumped structure follow as a corollary. The aggregation law turns local threshold laws into exact effective laws and makes unresolved heterogeneity appear as memory of the forcing that decays with the catchment's own relaxation times. Section~\ref{sec:model} presents a parsimonious model consistent with these laws, a mass-conserving structure with six rain--runoff parameters and five snow parameters (\CaRT), and Section~\ref{sec:refine} refines it iteratively. A deficit is diagnosed from the observations, and a refinement is derived from a stated fine-scale assumption through the operations admitted by the laws, in most cases nesting the model it refines. It is implemented as an exact daily law and kept if paired comparisons on a development split (Section~\ref{sec:loop}) show an improvement overall, or in some classes of catchments without a loss overall. The loop then starts again on the refined model, with AI agents assisting it under the author's instructions and the constraints of the laws. Removing each group of refinements from the final model and retraining the model measures what the group contributes. The result is \CaRTp, a single model for both continents without any daily network.

Section~\ref{sec:results} evaluates the models on 671 CAMELS-GB and 671 CAMELS-US catchments under one protocol and tests the transfer to 561 catchments of CAMELS-AUS, which were not used in model development. Training uses 1989--2008 and model selection 2009--2010, while the test period 2011--2019 was not used to fit the parameters or to select checkpoints. In addition, 10\,\% of the catchments of each continent are held out from training. The comparison includes HBV regionalised in the same way as \CaRT, the differentiable HBV of \citet{feng2022}, a stateful LSTM and an entity-aware LSTM (EA-LSTM) trained on the same inputs. Section~\ref{sec:discussion} discusses whether existing theory can describe the catchment response and in what sense universal hydrological laws may exist, and Section~\ref{sec:conclusions} concludes the paper.
 
\section{Hydrological laws}\label{sec:laws}
This section states what mathematical form a lumped daily catchment model must satisfy. It then shows how such a model is obtained from local hydrological mechanisms by aggregation and how it can be refined to include missing details. We show that effective catchment-scale relations between states and fluxes emerge from the aggregation of established local mechanisms under conservation. These effective relations are conditioned by the distribution, connectivity and interactions of heterogeneous elements and may therefore be nonlinear and scale dependent, and may carry memory. What is universal is the set of principles that produce these relations, namely the hydrological laws defined below.

\medskip
\noindent\fbox{\parbox{\dimexpr\linewidth-2\fboxsep-2\fboxrule}{\textbf{Definitions (hydrological law, constitutive family, constitutive relation).} A \emph{hydrological law} is a statement about catchment-scale storages and fluxes that (i) holds for every catchment; (ii) follows from a physical principle; and (iii) constrains the admissible \emph{form} of a model. Within the forms a law admits, a \emph{constitutive family} is one functional form chosen under a stated fine-scale assumption, such as a Pareto distribution of storage capacities or a power-law release, and can be shared by many catchments. The \emph{constitutive relation} of a catchment is the member of the family selected by catchment parameters, such as soil capacity and the release-law exponent, and must come from data. }}
\medskip

\noindent The laws are universal by construction, but whether the constitutive families are shared and how the parameters vary are empirical questions. Three laws of this kind and one corollary are stated below, and Table~\ref{tab:lawsummary} lists the form each statement imposes on a model and the choices left to the constitutive family and to data.

\begin{table}[tbp]\centering\small
\caption{The three hydrological laws of this section and their corollary: the form each imposes on a lumped model, and the choices left to constitutive families (assumptions) and parameters (data). The physical basis of each statement is summarised in Section~\ref{sec:lawsum}.}\label{tab:lawsummary}
\begin{tabular}{>{\raggedright\arraybackslash}p{2.3cm}>{\raggedright\arraybackslash}p{9.4cm}>{\raggedright\arraybackslash}p{3.4cm}}
\toprule
Statement & Form & Left to assumptions and data \\
\midrule
1 Conservation & Storage changes only through fluxes,\newline\hspace*{1em}$\dot{\bm x}=\bm B\bm q+\bm p-\bm e-\bm o+\bm j_{\rm ext}$, \eqref{eq:network}.\newline Internal fluxes conserve water, $\bm 1^{\!\top}\bm B=\bm 0^{\!\top}$.\newline Every correction acts on a flux. & the number of volumes; the family and parameters of each flux law \\[2pt]
2 Dissipation & Potential rises with storage, $\partial H_i/\partial x_i\ge0$.\newline Water flows down the potential, $q_{ij}=L_{ij}\,\Delta H_{ij}$, $L_{ij}\ge0$, \eqref{eq:dissip}.\newline Larger volumes exchange the sum of these fluxes,\newline\hspace*{1em}$q_{IJ}=\sum L_{km}(H_k-H_m)=L_{IJ}\,\Delta H_{IJ}$, \eqref{eq:dissipIJ}.\newline Phase change is bounded, $E_a=E_{\rm dem}\,\varepsilon$, $0\le\varepsilon\le1$. & the families of the conductances and of the demand function; their exponents and constants; the lumped potentials, a closure (Law~3) \\[2pt]
3 Aggregation & The aggregates obey $\dot{\bm z}=\bar{\bm f}(\bm z,\bm u)+\mathcal R$.\newline The closed law is the mean over the compositions,\newline\hspace*{1em}$\bar{\bm f}=\mathrm E\big[\bm\Pi\bm f(\bm\xi,\bm u)\,\big|\,\bm\Pi\bm\xi=\bm z,\ \bm u\big]$, \eqref{eq:fbar}.\newline The remainder is a memory of the history,\newline\hspace*{1em}$\mathcal R(t)=\mathcal R\big[\bm y(0);\,\bm z|_{[0,t]},\,\bm u|_{[0,t]}\big]$, \eqref{eq:elim}.\newline Only compositions that change the future of $\bm z$ need states. & the families of the threshold distribution; the states that carry the memory, by local structure (Table~\ref{tab:memory}); the amplitude and time constants of the correction \\[2pt]
\midrule
Corollary: Admissible refinement & (a) Split a volume, $\bm\Pi'\bm B'=\bm B\bm K$, \eqref{eq:PiB}.\newline (b) Add a connection.\newline (c) Change the form of a flux law.\newline (d) Modulate its coefficients by $\bm u$ or $\bm\chi$. & which refinement a catchment needs; the family of each refinement and its parameters \\
\bottomrule
\end{tabular}
\end{table}

\subsection{General laws}\label{sec:laws1}

\paragraph{Conservation.} Let $\theta(\bm r,t)$ be the local water content, $\bm j$ the water flux and $s_w$ the local net source. Partition a fixed catchment domain into $n$ non-overlapping control volumes $\Omega_i$, $i=1,\dots,n$, and define their storages as $x_i=A^{-1}\int_{\Omega_i}\theta\,\mathrm d\mathcal V$, where $A$ is the catchment area. The storages are collected in the storage vector $\bm x=(x_1,\dots,x_n)^{\!\top}$. Integrating $\partial_t\theta+\nabla\!\cdot\bm j=s_w$ over each volume and applying the divergence theorem yields
\begin{equation}\label{eq:network}
\dot{\bm x} = \bm B\,\bm q + \bm p - \bm e - \bm o + \bm j_{\rm ext},\qquad \bm 1^{\!\top}\bm B=\bm 0^{\!\top}.
\end{equation}
Here $\bm q$ contains the net internal fluxes with prescribed reference directions, and $\bm B$ is their incidence matrix, in which each column has $-1$ at the source volume and $+1$ at the receiving volume. The vectors $\bm p$, $\bm e$ and $\bm o$ hold, for each volume, the precipitation it receives, its evaporation (including transpiration and sublimation) and its outflow to the outlet, all non-negative. The vector $\bm j_{\rm ext}$ holds the other exchanges of each volume across the catchment boundary, such as regional groundwater flow, positive into the volume. Each boundary flux is assigned to one volume, although boundary fluxes can also be represented by an augmented incidence matrix that includes the atmosphere, the outlet and the exterior as boundary nodes. 

Storages and fluxes are expressed as catchment-equivalent depths and depth rates, so that a volume covering an area fraction $a_i$ and holding a mean local water depth $h_i$ has storage $x_i=a_i h_i$. An internal transfer leaves one volume and enters another at the same rate, so each column of $\bm B$ sums to zero. Consequently, the total catchment-equivalent storage $V=\bm 1^{\!\top}\bm x$ satisfies $\dot V=P-E_a-Q+J_{\rm ext}$ for any such partition, where the terms on the right are the sums of the corresponding external contributions.

Conservation fixes this balance structure but does not give a closure for the internal fluxes. A lumped model represents them as $\bm q=\bm q(\bm x,\bm u,\bm\chi)$, where $\bm u$ is the forcing and $\bm\chi$ denotes conditional variables, such as connectivity or vegetation activity, that affect the fluxes without contributing to water storage. Whether these variables suffice for an exact closure is discussed in Section~\ref{sec:laws2} and Section~\ref{sec:laws3}.

\begin{law}[Conservation]\label{law:cons}
A lumped catchment model is a storage vector with $\dot{\bm x}=\bm B\bm q+\bm p-\bm e-\bm o+\bm j_{\rm ext}$, $\bm 1^{\!\top}\bm B=\bm 0^{\!\top}$.
\end{law}

 The state is the set of stores, and each internal flux removes water from one store and adds it to another (each column of $\bm B$ sums to zero), so water is conserved whatever the flux laws are. A model is changed only by changing, adding or removing fluxes and stores.

\paragraph{Bounds.} Storages are non-negative and some have a capacity ($0\le S\le C$), and the state must not leave the admissible set ($\dot x_i\ge0$ at $x_i=0$, $\dot S\le0$ at $S=C$).

\paragraph{Dissipation.} Water in the pores and in the channels obeys the Navier--Stokes equations, under which the free energy of the water in any fixed region changes by the work done at its boundary minus a viscous dissipation that is never negative. In the pores the momentum relaxes in much less than a second, and on hillslopes and in channels in minutes to hours, both fast compared with the daily step of lumped models. Averaging Stokes flow over the pores then gives Darcy's law, $\bm q=-\bm K\nabla H$ with $\bm K$ symmetric and positive definite \citep{tartar1980,allaire1989,hornung1997}. On hillslopes and in channels the Saint-Venant equations reduce to the kinematic or diffusive wave at small Froude number \citep{lighthill1955,chen1994}. In both cases the flux follows the gradient of the head $H$, in the kinematic wave through its bed-slope part alone. The finest partition is a modelling level bounded by two scales. Its units must be larger than the representative elementary volume, so that Darcy's law or the Saint-Venant equations hold for them \citep{bear1972}. They must also be small enough for the head within a unit to even out faster than the forcing changes, $\ell\ll\sqrt{D\,T}$, with $\ell$ the size of the unit in the direction considered and $T$ the time scale of the forcing. The hydraulic diffusivity $D$ is the conductivity divided by the change of storage per unit change of head, and $\ell^2/D$ is the time over which the head within a unit evens out. For daily forcing this upper bound is about a decimetre to a metre vertically in an unsaturated soil of moderate wetness, metres to tens of metres laterally in an unconfined sandy aquifer and kilometres along a channel. We take the finest partition to be K-orthogonal, with the line between the centres of two neighbours normal to their interface in the metric of $\bm K$, as a Voronoi partition is for isotropic $\bm K$. The flux across the interface between two volumes is then approximated by a conductance times the difference of their heads. This two-point flux of finite volumes is consistent on such a partition and converges to the local flux as the partition is refined, while on other partitions it also depends on the heads of further neighbours. The distributed system of Section~\ref{sec:laws2} is this fine model, and the decompositions below are exact for it. On this partition each volume $i$ has an effective potential $H_i(x_i,\bm\chi)$, which is the head that drives water out of it. This potential does not decrease as the storage increases, since a water table rises with storage and the retention curve has a positive capillary capacity. The boundary nodes have prescribed potentials. The outlet has a fixed datum and the exterior a regional head. With $\Delta H_{ij}=H_i-H_j$, the flux between two volumes is driven by the potential difference and does work against resistance, $q_{ij}\Delta H_{ij}\ge0$:
\begin{equation}\label{eq:dissip}
q_{ij} = L_{ij}(\bm x,\bm u,\bm\chi,\Delta H_{ij})\,\Delta H_{ij},\qquad L_{ij}\ge 0.
\end{equation}
Here the effective conductance may depend on storage, on the forcing, on the potential difference and on $\bm\chi$. Two larger volumes $I$ and $J$, each made of finer volumes, exchange the sum of the fluxes across the interfaces they share,
\begin{equation}\label{eq:dissipIJ}
q_{IJ}=\sum L_{km}(H_k-H_m)=L_{IJ}\,\Delta H_{IJ},\qquad L_{IJ}=\sum L_{km},\qquad \Delta H_{IJ}=\frac{\sum L_{km}(H_k-H_m)}{L_{IJ}},
\end{equation}
where the sums run over the pairs of finer volumes $k$ in $I$ and $m$ in $J$ that share an interface. The total conductance is non-negative, and $\Delta H_{IJ}$ is the conductance-weighted mean of the potential differences across these interfaces, which the potentials of the finer volumes at the interface determine. Where $L_{IJ}=0$, every flux across the interfaces vanishes and $\Delta H_{IJ}$ is not needed. Like Law~\ref{law:cons}, this relation holds in every catchment and on every partition made of finer volumes. Where the potential of every finer volume of $I$ exceeds that of every finer volume of $J$, the flux runs from $I$ to $J$ whatever the composition, as for outflow to the outlet, spill and percolation to a deeper water table. Fluxes that change the phase of water (evaporation, sublimation, melt) are driven by a vapour pressure deficit or an energy supply and are bounded by the energy available and the water present. Representing atmosphere and vegetation by a demand $E_{\rm dem}$ gives $E_a=E_{\rm dem}\,\varepsilon(\bm x,\bm\chi)$ with $0\le\varepsilon\le1$, $\varepsilon(0)=0$, $\varepsilon'\ge0$. Melt and sublimation take the smaller of their energy-limited rate and the snow available.

\begin{law}[Dissipation]\label{law:dissip}
Water flows down its potential against a resistance. On the finest partition each volume has a potential that does not decrease with its storage, and the flux between two volumes, or between a volume and a boundary node of prescribed potential, is a non-negative conductance times their potential difference, \eqref{eq:dissip}. Between larger volumes the flux is the sum of these fluxes, \eqref{eq:dissipIJ}. Phase-change fluxes are bounded by the available energy and water.
\end{law}
\noindent The difference $\Delta H_{IJ}$ is in general not a function of the lumped storages, and writing it as one is a closure, the subject of Law~\ref{law:closure}. At the reference state of Section~\ref{sec:laws2} the units of each aggregate that are neither empty nor full share one potential, and where the units at the interface are of this kind, $\Delta H_{IJ}$ is the difference of the two lumped potentials.

\noindent A flux law may depend on the states of both volumes it connects and on conditional variables. The flux laws of this paper have a further property beyond the sign fixed by \eqref{eq:dissip}. A flux out of a volume does not decrease as the storage of that volume increases and does not increase as the storage of the receiving volume increases. This monotonicity is constitutive. It holds where the conductance at each interface is taken from the source and does not decrease with the storage of the source, as the conductivity of unsaturated soil grows with its water content. It restricts the exchange of Section~\ref{sec:derive} to a loss and a constant inflow and, under the locality, drainage and coupling conditions of Appendix~\ref{app:lose}, lets the memory of initial states fade. A flux law reduces to a single-state storage--discharge relation $q=f(x_i)$ only if the downstream head is fixed and the upstream head and conductance are functions of $x_i$ alone. The single-state reservoir, which underlies almost all conceptual models, is therefore conservation plus a one-state hydraulic closure.

\subsection{Aggregation and closure}\label{sec:laws2}

\paragraph{The closure condition.} The distributed system is \eqref{eq:network} written on the finest partition (grid cells). We write it as $\dot{\bm\xi}=\bm f(\bm\xi,\bm u)$, with $\bm\xi=(\bm x,\bm\chi)$ the fine state, made of the vector $\bm x$ of all local storages and the vector $\bm\chi$ of all local conditional variables on which the fluxes depend. The conditional variables are of $K$ kinds, such as the wetness of hillslope units or the activity of vegetation patches. Kind $k$ is carried by $n_k$ units with area weights $a_{ki}$ that sum to one, and $\bm\chi=(\bm\chi_1,\dots,\bm\chi_K)$ lists their local values $\chi_{ki}$ kind by kind, so that $\bm\chi$ has $n_1+\dots+n_K$ components,
\[
\dot{\bm x}=\bm B\,\bm q(\bm x,\bm u,\bm\chi)+\bm p(\bm u)-\bm e(\bm x,\bm u,\bm\chi)-\bm o(\bm x,\bm\chi)+\bm j_{\rm ext},\qquad \dot{\bm\chi}=\bm\gamma(\bm x,\bm u,\bm\chi),
\]
the first being \eqref{eq:network} on that partition, with $\bm u$ the catchment-mean precipitation, temperature and potential evaporation, and the second the balances of $\bm\chi$, which carry no water and do not enter $\bm B$. A lumped model keeps only aggregates $\bm z=\bm\Pi\bm\xi$ (for example, a few storage tanks representing the whole catchment), with each row of $\bm\Pi$ summing the local storages that make up one aggregate or taking the area mean of one kind, $\bar\chi_k=\sum_ia_{ki}\chi_{ki}$. The aggregates $\bm z$ thus contain the lumped stores and the $K$ conditional variables of the lumped model, $\bar{\bm\chi}=(\bar\chi_1,\dots,\bar\chi_K)$. As for $\bm x$, the laws of Section~\ref{sec:laws1} hold on any partition, with $\bm\chi$ the conditional variables of that partition. A closed lumped model $\dot{\bm z}=\bar{\bm f}(\bm z,\bm u)$ exists if and only if
\begin{equation}\label{eq:closurecond}
\bm\Pi\,\bm f(\bm\xi,\bm u)=\bar{\bm f}(\bm\Pi\bm\xi,\bm u)\quad\text{for all reachable }\bm\xi,
\end{equation}
together with the same condition for the outlet flow, so that all internal configurations with the same aggregates produce the same aggregate change. This is exact lumpability \citep{wei1969,li1989}, which rarely holds. However, the unresolved contribution can still be expressed without approximation. Define the composition as the deviation $\bm y=\bm\xi-\bar{\bm\xi}(\bm z)$ of the fine state from the reference state $\bar{\bm\xi}(\bm z)$, in which each kind of conditional variable sits at its area mean and the water of each aggregate is distributed over its units so as to minimise the stored potential energy $\sum_i\int_0^{x_i}H_i\,\dd x$ for the given aggregates. For fixed conditional variables, internal exchange cannot increase this energy (Law~\ref{law:dissip}). At the minimum the units that are neither empty nor full share one potential, while an empty unit has a potential no lower and a full unit one no higher. The minimum exists because the potentials are continuous and the storages of given aggregates lie in a bounded set. It is unique where the potentials increase strictly, and it puts units with the same relation between potential and storage and the same capacity at the mean of their aggregate. The composition is the part of the fine state that the aggregates do not resolve. Then
\begin{equation}\label{eq:split}
\dot{\bm z}=\underbrace{\bm\Pi\bm f(\bar{\bm\xi}(\bm z),\bm u)}_{\bm F_0(\bm z,\bm u)}+\underbrace{\bm\Pi\big[\bm f(\bar{\bm\xi}(\bm z)+\bm y,\bm u)-\bm f(\bar{\bm\xi}(\bm z),\bm u)\big]}_{\Delta\bm F(\bm z,\bm y,\bm u)} .
\end{equation}
The aggregate rate of change is thus the sum of the rate at the reference state ($\bm F_0$) and the correction caused by the actual unresolved composition ($\Delta\bm F$). The contribution of each flux between two aggregates to $\Delta\bm F$ is the departure of $L_{IJ}\,\Delta H_{IJ}$ from its value at the reference state, at which $\Delta H_{IJ}$ is the difference of the lumped potentials where the units at the interface are neither empty nor full. Since $\bm z$ contains $\bar{\bm\chi}$, the reference state $\bar{\bm\xi}(\bm z)$ puts the units of each aggregate at one potential, except those at a bound, and the conditional variables at the area means $\bar\chi_k$. Hence $\bm F_0$ contains the lumped fluxes as functions of the lumped stores, $\bm u$ and $\bar{\bm\chi}$, the form $\bm q(\bm x,\bm u,\bm\chi)$ of Section~\ref{sec:laws1}. For the conditional variables, \eqref{eq:split} reads
\begin{equation}\label{eq:chisplit}
\dot{\bar\chi}_k=\sum_ia_{ki}\,\gamma_{ki}\big(\bar{\bm\xi}(\bm z),\bm u\big)+\sum_ia_{ki}\big[\gamma_{ki}\big(\bar{\bm\xi}(\bm z)+\bm y,\bm u\big)-\gamma_{ki}\big(\bar{\bm\xi}(\bm z),\bm u\big)\big],
\end{equation}
the balance with every unit at the mean and the change due to the deviations $\chi_{ki}-\bar\chi_k$, as in \eqref{eq:meandev} for the hillslope wetness. For example, two soil units with the same total storage but different wetness drain more than the same two units at their mean wetness, because drainage grows faster than linearly with wetness. The aggregate storage alone cannot tell these two cases apart, and their difference is $\Delta\bm F$. Equation~\eqref{eq:split} is therefore the formal starting point of refinement.

$\bm F_0$ is the rate of change of the aggregates when the units sit in the reference state. The same aggregates can however arise from many internal states, which give different rates. A closed lumped model can at best return their average for given aggregates and forcing, with each state weighted by how often it occurs in the catchment under the forcing of the period in which the model is learned,
\begin{equation}\label{eq:fbar}
\bar{\bm f}(\bm z,\bm u)=\mathrm E\big[\bm\Pi\bm f(\bm\xi,\bm u)\,\big|\,\bm\Pi\bm\xi=\bm z,\ \bm u\big]=\bm F_0(\bm z,\bm u)+\mathrm E\big[\Delta\bm F\,\big|\,\bm z,\bm u\big],
\end{equation}
which is the closed model of \eqref{eq:closurecond} where that exists. The remainder $\mathcal R=\dot{\bm z}-\bar{\bm f}$ contains the part due to the composition, $\Delta\bm F-\mathrm E[\Delta\bm F\,|\,\bm z,\bm u]$, which has zero mean for given aggregates and forcing. The rest of Section~\ref{sec:laws} discusses the form of $\mathcal R$.

A direct way to reduce $\mathcal R$ is to explicitly represent part of the fine states by splitting the lumped stores. Alternatively, $\mathcal R$ can be reduced by aggregation of local thresholds, which removes it where the threshold distribution is resolved, or represented by memory (Section~\ref{sec:laws3}, Appendix~\ref{app:moments}).

Part of $\mathcal R$ is not an effect of the composition of the stores. The forcing $\bm u$ is a catchment mean, and the fine state also responds to how the day's rain and melt are spread over the catchment, which neither the aggregates nor their history determine. Fluxes that people control, such as abstractions, discharges and reservoir releases, are boundary fluxes in Law~\ref{law:cons} that the forcing does not drive, and they enter $\mathcal R$ too. These parts form a residual that no lumped structure recovers and that remains after the three ways above have reduced the rest of $\mathcal R$.

\paragraph{Effective laws from local thresholds.} Most local hydrological laws are threshold laws, as when a point produces runoff once its storage capacity is exceeded. If each unit contributes to the outlet through a local law $q_{\rm loc}(v;c)$ of a driving variable $v$ shared by the units and a threshold $c$ that varies in space with distribution $F_c$, the aggregate is the Lebesgue--Stieltjes integral,
\begin{equation}\label{eq:aggthr}
\bar q(v)=\int q_{\rm loc}(v;c)\,\dd F_c(c).
\end{equation}
 The aggregate has no free shape parameters beyond those of $F_c$ and is exact when the state of every unit is set by $v$ and its own threshold, so that two units with the same $c$ are in the same state. Where the units have instead reached different values $v_i$, the aggregate is $\sum_ia_iq_{\rm loc}(v_i;c_i)$, an integral over the joint distribution of the thresholds and the driving values. Its departure from \eqref{eq:aggthr} is the effect of the spread of the $v_i$, which is the composition $\bm y$ of \eqref{eq:split} for a threshold law.

As an example, capacities $c$ of units in parallel with a Pareto distribution, $F_c=1-(1-c/c_{\max})^b$, give the probability-distributed store \citep{moore1985,zhao1992}. Under uniform rain from empty, every unit that is not yet full holds the same depth, which is the shared variable, and the exact increase in storage under a net rain depth $P_n$ is
\begin{equation}\label{eq:pareto}
p_s=C\big[D-\big(D^{1/(b+1)}-P_n/c_{\max}\big)_+^{\,b+1}\big],\qquad D=1-S/C,\quad c_{\max}=(b+1)C.
\end{equation}
 The saturation-excess runoff is $P_n-p_s$. The GR4J partition \citep{perrin2003}, with runoff fraction $\eta=s^2$ of the net rain and relative storage $s=S/C$, is a member of the same class, with $F_c(c)=\tanh^2(c/C)$. Evaporation between events lowers the units that were full to depths that depend on their capacity. The units then no longer share one depth, and \eqref{eq:pareto} holds for filling only.

The other threshold laws of this paper are the same integral with different driving variables $v$, which are the rain for the infiltration excess, the quick store for the release exponent (Section~\ref{sec:derive}), the wetness for the connected fraction of the hillslope (Appendix~\ref{app:memory}) and the total storage along the path for units that spill into one another (Section~\ref{sec:laws3}). Once the units cease to share $v$, the aggregate depends on how they came to differ, and Section~\ref{sec:laws3} shows that this dependence is a memory of the forcing.

\paragraph{Effective laws from smooth local laws.} Let $w_i$ be the change of the outflow of the aggregate per unit change of the local depth or conditional variable of unit $i$, divided by its area. The first-order term of $\Delta\bm F$ is a sum over $w_i$ times the deviation of each unit, and it vanishes for all compositions when the $w_i$ are equal. This defines units in parallel, each of which acts on the outflow in the same way, and units in series, of which only those near the outlet act on it directly. A conditional variable that acts on the release of every unit alike, as the wetness of the hillslope units does, is in parallel, and one that acts only through the units below it is in series. Units in parallel with smooth laws therefore have a correction of second order (Appendix~\ref{app:lose}), whereas for units in series the first-order term remains. Let local units have states $s_i=\bar s+\delta s_i$ and properties $\kappa_i=\bar\kappa+\delta\kappa_i$ with zero-mean deviations of order $\varepsilon$, and outflow $q_i=f(s_i,\kappa_i)$. Expanding $f$ about the means and averaging over the units (Appendix~\ref{app:lose}) gives
\begin{equation}\label{eq:taylor}
\bar q=f(\bar s,\bar\kappa)+\tfrac12 f_{ss}\Var(s)+f_{s\kappa}\Cov(s,\kappa)+\tfrac12 f_{\kappa\kappa}\Var(\kappa)+O(\varepsilon^3).
\end{equation}
The expansion is about the arithmetic means. Where the units share one relation between potential and storage and differ only in their conductances, the reference state puts every unit at the mean storage, the terms in $\Var(s)$ and $\Cov(s,\kappa)$ are $\Delta\bm F$ and the term in $\Var(\kappa)$ belongs to $\bm F_0$. Where the properties enter the potential, the reference state has a spread of its own, and $\Delta\bm F$ is the change of these terms from their values at the reference state. This correction can still be large although it is of second order. Two units with storages $\bar s(1\pm\varepsilon)$ and $f\propto s^5$ release $1+10\varepsilon^2+5\varepsilon^4$ times the single-store flow, 41\,\% more for $\varepsilon=0.2$. Heterogeneity thus makes the mean outflow depend on both the mean storage and the spread of the storages. But the spread is not a function of the current mean. Both integrals, \eqref{eq:aggthr} and \eqref{eq:taylor}, depend on the distribution of the units at the given moment, which the aggregates do not fix, and Section~\ref{sec:laws3} treats how this distribution depends on the history.

\subsection{Spatial composition and memory}\label{sec:laws3}

 Eliminating the composition from the equations leaves a functional of the history of the aggregates and the forcing, plus the contribution of the initial composition, which fades where all water leaves through paths that respond to storage and the variables that modulate the fluxes do not feed back on the water that drives them (Appendix~\ref{app:lose}, Eq.~\eqref{eq:elim}). This elimination underlies the Mori--Zwanzig projection \citep{mori1965,zwanzig1973,chorin2002}. A lumped model can therefore keep the composition as extra states or replace it by a memory of the history of the aggregates and the forcing, and the two are equivalent. The memory takes different forms for threshold and smooth laws and for units in parallel and in series.

\paragraph{Thresholds on the storage of each unit.} Where the threshold acts on the storage of each unit, the memory is a hysteresis of the deficits. Let $d_i=c_i-s_i$ be the deficit of unit $i$ and $F_d$ its area distribution at the start of a wet spell. Under uniform rain without evaporation, a unit in parallel is full once the rain $P$ accumulated since then exceeds its deficit. The fraction of the area that is full, and passes the rain on, is then the connected fraction $F_d(P)$, and the storage is
\begin{equation}\label{eq:deficit}
z=z_0+\int_0^P\big[1-F_d(P')\big]\,\dd P'.
\end{equation}
Since $z$ grows with $P$, the connected fraction is a function of $z$ for a given $F_d$. The storage fixes only the mean deficit, $\bar d=\bar c-z$, and the flux depends on the whole distribution. Filling and drying change this distribution in different ways. Rain lowers each deficit until it reaches zero, $d_i\to\max(d_i-P,0)$, and erases the differences among the units that fill. Uniform evaporation $E$ raises each deficit until the unit is empty, $d_i\to\min(d_i+E,c_i)$, and keeps the differences among the units that still hold water. From empty, $F_d=F_c$, and the storage follows the integral \eqref{eq:pareto}. After drying by $E$ from full, every unit that kept water has the deficit $E$, and these units reconnect together when $z$ returns to its value before drying. The same storage thus corresponds to different deficit distributions, and to different fluxes, after a wet and after a dry spell. Along a path a unit connects once it and every unit below it are full. With immediate spill and uniform rain from empty, the outflow is the rain on this connected part, $Q=p\,F_c(z)$, where $c$ is now the total storage $z$ at which a unit connects. The distribution of $c$ is set by the order of the capacities along the path, since small capacities upstream merge into the thresholds of the larger ones below them \citep{shook2011,tromp2006}. After drying, the connected fraction depends on the deficits along the path in the same way. The deficit distribution is therefore the composition that the stores do not resolve. With losses that do not depend on the state of the unit and no outflow between empty and full, each unit is a stop operator of the accumulated net forcing. The deficit distribution of a store is the superposition of these operators, a Prandtl--Ishlinskii hysteresis operator of the forcing \citep{okane2007,brokate1996}. The state depends only on the path of the accumulated forcing and on its turning points, so the memory has no time scale. An event that fills or empties all units erases it. A lumped model can carry this memory as the deficit distribution or as the current accumulated forcing and the turning points that generate it (Appendix~\ref{app:realise}).

\paragraph{Thresholds on a shared driving variable.} Thresholds can also carry a memory that relaxes. Where the units share a driving variable that the stores do not resolve, or lateral exchange brings them to share it (Appendix~\ref{app:memory}), that variable has its own balance with a drainage term. It enters the integral \eqref{eq:aggthr} through its argument, $\bar q=\int q_{\rm loc}(\bar\chi_k;c)\,\dd F_c(c)$, or through the scale of $F_c$, with the shape of $F_c$ fixed. The flux keeps the sign of the local law and can start from zero. In \CaRTp{} the hillslope wetness $\theta$ shifts the connection thresholds of the hillslope units, and the near-surface wetness scales the infiltration capacities (Appendix~\ref{app:memory}). Where the integral is a fraction bounded away from zero acting on a positive flux, the result is a multiplier of that flux, as for the connected fraction $A\ge A_{\min}$ that multiplies the release.

\paragraph{Parallel units with smooth fluxes.} For units in parallel as defined in Section~\ref{sec:laws2}, inputs create the spread in \eqref{eq:taylor} (units that receive more, or drain more slowly, pull ahead), while drainage removes it (the wetter units drain faster). The current spread therefore remembers the inputs of the recent past, over the time drainage takes to remove it (moment equations in Appendix~\ref{app:lose}). A lumped model that represents the correction keeps the spread as an extra state or as a memory of the forcing. For a conditional variable in parallel the same holds, with the contraction of its own balance in place of drainage, as for the wetness in \eqref{eq:meandev}. Where the properties differ, the term in $\Cov(s,\kappa)$ is linear in the imbalance of the storages, and this term has an exact description in the simplest case of two heterogeneous reservoirs.

\paragraph{A worked example: two heterogeneous reservoirs.} Take two units of equal area that receive the same recharge $G(t)$ and release at rates $r_0(1\pm\varepsilon)m(W)$, with $m(W)$ a shared factor of the total storage $W=(s_1+s_2)/2$. The units differ only in their release rates, so the reference state puts both at $W$. With the imbalance $D=(s_1-s_2)/2$ and its scaled form $h=-D/\varepsilon$ ($\varepsilon\ne0$), the pair is represented exactly by (Appendix~\ref{app:prop1}):
\begin{equation}\label{eq:hstate}
\dot W=G-r_0m(W)\,[W-\varepsilon^2h],\qquad \dot h=r_0m(W)\,[W-h].
\end{equation}
The total obeys the single-store law plus a term in $h$. If both units start with the same storage, $h$ starts at zero, and the linear equation for $h$ makes it a weighted average of the past values of $W$. An initial imbalance adds a transient that is forgotten after a few residence times. The correction is therefore a memory of the store's own history, of amplitude $\varepsilon^2$, with weights that decay over the residence time of the store. Appendix~\ref{app:prop1} presents the derivation and the daily form (an exponential moving average).

The correction of units in parallel starts at second order and acts on the flux of the reference state. In \eqref{eq:hstate} the release is $r_0m(W)W\,(1-\varepsilon^2h/W)$, the single-store release times a multiplier close to one, which is the form of operation (d) of Section~\ref{sec:laws4}.

\paragraph{Units in series with smooth fluxes.} When the units drain into one another, $w_i=0$ for all but the last, and the composition is the position of the water along the path. This composition arises even for identical units. Under steady input every store passes the input on, the total storage is $z=Tp$ with $T$ the mean travel time, and the outflow is $z/T$. This quasi-steady closure reads the outflow from the total storage, but the outflow is set by the last store alone. After a rise of the input the new water is still upstream, so $z$ has increased while the outflow has not, and the outflow falls below $z/T$. After a fall the upstream stores empty first and the outflow stays above it. For input that varies slowly compared with $T$, the outflow of $n$ identical stores falls below $z/T$ by the fraction $\frac{n-1}{2n}\,T\dot p/p$ (Appendix~\ref{app:realise}). The memory of units in series is therefore a memory of the changes of the input. It is present for identical units, whereas the memory of units in parallel is not driven when the units are identical and fades as their initial differences decay, and a lumped model of a chain has to keep the water in transit.

The water in transit can be kept as stores, one for each unit, for any flux laws. For linear, time-invariant units it can also be kept as a convolution. The outflow is then exactly the convolution $q=h*p$ of the input with the travel-time distribution $h$ of the network, the distribution of the time that water takes from the input to the outlet, which has mean $T$ and is also the response of the outflow to a unit pulse of input. The kernel is causal, non-negative and of unit mass. Unit mass conserves water (Law~\ref{law:cons}), with the water in transit as a store, and a non-negative kernel keeps the sign of the flux (Law~\ref{law:dissip}). A chain of $n$ identical linear stores has a gamma kernel \citep{nash1957}. Where the network is not known, a kernel family with few parameters approximates $h$, as the unit hydrographs of Section~\ref{sec:model} do with one parameter.

\paragraph{Reduction to a few states.} The composition itself is a set of states that carries $\mathcal R$ exactly except for the spatial heterogeneity of forcing and human influence. A state needs to carry a difference in composition only while that difference can still change the future of the aggregates. Under the monotonicity of Section~\ref{sec:laws1} and the drainage and coupling conditions of Appendix~\ref{app:lose}, this influence fades. Every flux is driven by a potential difference, so a unit loses more water when its potential (head) is higher than that of its neighbours, and differences between units shrink at a rate set by how fast the losses grow with storage and by the conductance of the exchange between units (Appendix~\ref{app:lose}). In catchments these rates fall into a few separated time scales, hours for the redistribution of water within an event, days to weeks for soils and hillslopes and months to years for groundwater. A difference that relaxes within the day follows the current state and forcing after a short transient and enters the effective drift of the slower states, so only the slower differences that the forcing drives need a state. Where units in parallel share the forcing, the contraction also keeps their spread small, since inputs pull them apart only through their differences in properties, and for smooth laws the low moments of the spread approximate its effect. Where the rate of a unit does not depend on its storage between its bounds, Law~\ref{law:dissip} gives no contraction. A turning point of the accumulated forcing then stays in the memory until the forcing passes it, and an event that fills or empties all units erases all of them. In Section~\ref{sec:results}, one state for each conditional variable and the water in transit are enough, and Appendix~\ref{app:realise} gives the conditions under which so few states keep the error small. Table~\ref{tab:memory} lists the states that remain for each local structure.

\begin{law}[Aggregation]\label{law:closure}
The aggregates obey $\dot{\bm z}=\bar{\bm f}(\bm z,\bm u)+\mathcal R$, with $\bar{\bm f}$ the mean of the local laws over the compositions for given $\bm z$ and $\bm u$, \eqref{eq:fbar}, and $\mathcal R$ the remainder, which contains the effect of the composition that $\bm z$ does not resolve. Apart from the response to the spatial pattern of the forcing and to the fluxes that people control, $\mathcal R$ is a functional of the history of $\bm z$ and $\bm u$ and of the initial composition, \eqref{eq:elim}. A closed model need keep only the differences in composition that change the future of the aggregates under some future forcing.
\end{law}
\noindent The response to the pattern of the forcing and to the controlled fluxes is the residual of Section~\ref{sec:laws2}.

\begin{table}[t]
\caption{Memory of the unresolved composition by local structure. The states realise the history functional \eqref{eq:elim}. Op.\ is the operation of Corollary~\ref{cor:refine} through which the memory enters.}\label{tab:memory}
\small
\begin{tabular}{p{2.6cm}p{4.0cm}p{2.9cm}p{3.9cm}l}
\toprule
Local structure & States that carry the memory & Enter the flux as & Size and time scale & Op. \\
\midrule
Smooth laws, units in parallel & slow modes of the composition, no water (Eqs.~\ref{eq:hstate}, \ref{eq:vardyn}) & multiplier of a conductance & second order in the heterogeneity of states and properties together; relaxes as the spread drains & d \\[2pt]
Smooth laws, units in series & stores for water in transit & convolution of the input (linear laws), or a chain of stores & first order in the rate of change of the input, present for identical units; the travel-time distribution \eqref{eq:series} & c, a \\[2pt]
Thresholds with a shared driving variable that the stores do not resolve & the driving variable itself, no water & shift or scale of the threshold distribution in \eqref{eq:aggthr}, its shape fixed & leading order, through the threshold distribution; the relaxation of the driving variable & d \\[2pt]
Units in parallel that fill to a capacity and spill & the current accumulated forcing and its turning points since the last event that filled or emptied all units & shape of the threshold distribution in \eqref{eq:aggthr}, which is the deficit distribution & leading order, through the deficit distribution; no time scale, erased by that event & d \\
\bottomrule
\end{tabular}
\end{table}

\subsection{Admissible refinement}\label{sec:laws4}

The derivation above starts from the fine system, but conceptual models are usually built the other way, from a few stores to which detail is added where a closure fails. Laws 1--3 admit four kinds of refinement. Two change the network and two change a flux law, and they can be combined.

Splitting a volume (a) represents its composition explicitly without adding a process. Where $\bm\Pi'$ aggregates the new volumes into the old ones, the structural relation is
\begin{equation}\label{eq:PiB}
\bm\Pi'\bm B'=\bm B\bm K,
\end{equation}
with $\bm K$ a flux-aggregation matrix of entries 0 or 1 that maps each new flux to the old flux it is part of, or to none for an exchange between sub-volumes. The extension only reorganises the old connections.

Adding a connection (b) is an extension for which no such $\bm K$ exists. It adds a flux between two volumes that exchanged no water before, or between a volume and a boundary node it was not connected to, such as uptake from groundwater to the atmosphere, evaporation from depressions or drainage straight to the outlet. A new flux between volumes adds a column to $\bm B$, and a new boundary flux adds to the boundary terms of \eqref{eq:network}. A new volume, such as a snowpack on the precipitation or depressions on the path of direct runoff, combines the two. Its fluxes that map onto old ones split a volume, and the others are new connections.

Changing the form of a flux law (c) leaves $\bm B$ unchanged and includes adding a second flux between two volumes that are already connected. Where the local law has a threshold, the new flux law is the integral of Law~\ref{law:closure} over the threshold distribution \eqref{eq:aggthr}, as for the Pareto store \eqref{eq:pareto}.

Modulating the coefficients of a flux law (d) by the forcing or by conditional variables also leaves $\bm B$ unchanged, since the conditional variables carry no water. By Law~\ref{law:dissip} the modulated flux must keep the sign of the potential difference, so a modulation is written as a multiplier, $q=q_0M$ with $M\ge0$, which does not change the sign of the flux and keeps it zero when its store is empty. Where the modulated flux stays positive wherever $q_0$ is, $M=e^{m}$, with $m$ a function of the states, the forcing and the conditional variables. A modulation that moves a threshold changes where the flux vanishes, and it belongs to the threshold distribution. Writing a lumped flux as a conductance times the difference of the lumped potentials is itself a closure. Where an exchange runs in different directions between different units, the conditional mean \eqref{eq:fbar} need not have this form. The model then keeps the form and accepts the difference from \eqref{eq:fbar} as a structural error, which, unlike the part of $\mathcal R$ due to the composition, need not have zero mean for given aggregates and forcing. In return, the lumped model keeps the sign structure of the fine system for any values of its parameters, with no outflow from an empty store. Where its flux laws are also monotone in the sense of Section~\ref{sec:laws1}, a constitutive property beyond Law~\ref{law:dissip}, and the drainage and coupling conditions of Appendix~\ref{app:lose} hold, it forgets its initial state.

\begin{corollary}[Admissible refinement]\label{cor:refine}
By Laws 1--3, a lumped model is refined in four ways: (a) split a volume, with $\bm\Pi'\bm B'=\bm B\bm K$; (b) add a connection, a flux between two volumes that exchanged no water before or a flux between a volume and the boundary; (c) change the form of a flux law; (d) modulate the coefficients of a flux law (its conductance, its demand or its threshold distribution) by the forcing or by conditional variables $\bm\chi$, which are states that carry no water and leave $\bm B$ unchanged. Operations (a) and (b) change the network and (c) changes the form of a flux law, all within Law~\ref{law:cons}, with the sign of each new flux fixed by Law~\ref{law:dissip}, and (d) keeps the sign of the law it modulates. The memory of Law~\ref{law:closure} can enter through three of them, as a chain of stores (a), as a convolution kernel (c), or as a multiplier or a shift of a threshold distribution driven by conditional variables (d). Each such refinement conserves water by construction.
\end{corollary}
\noindent\emph{Rule.} We require each refinement to nest the unrefined model in a limit of its parameters where such a limit exists, allowing the added value of the refinement to be measured by a direct comparison.

\subsection{Summary}\label{sec:lawsum} Laws 1--3 and Corollary~\ref{cor:refine} are formally derived from basic principles that do not require new observations, namely conservation of mass, negligible inertia at the time scales of catchment hydrology and, for Law~\ref{law:closure}, a fine system whose rates are locally Lipschitz in its state. Together they restrict a lumped model to a class of dynamical systems, a compartmental system on a flux network \citep{jacquez1993} driven by differences of lumped potentials,
\begin{equation}\label{eq:class}
\dot{\bm x}=\bm B\,\bm q+\bm p-\bm e-\bm o+\bm j_{\rm ext},\qquad q_k=L_k(\bm x,\bm u,\bm\chi)\,\Delta H_k,\ \ L_k\ge0,\qquad \dot{\bm\chi}=\bm\gamma(\bm x,\bm u,\bm\chi),
\end{equation}
with $\bm x$ and $\bm\chi$ the stores and the conditional variables of the lumped model, which make up its aggregates $\bm z$ (Section~\ref{sec:laws2}), and the water balance a balance of $\bm x$ alone, in which the lumped flux laws follow from the aggregation of local laws and the unresolved composition enters through added stores or through the conditional variables $\bm\chi$, which carry no water. The lumped potentials are exact at the reference state of Section~\ref{sec:laws2} for interfaces whose units are neither empty nor full, and a closure elsewhere, since the weighted differences of Law~\ref{law:dissip} need not derive from one potential for each store. Within this structure, specific constitutive families are hypotheses that must be tested against observations. 
\section{A base model}\label{sec:model}
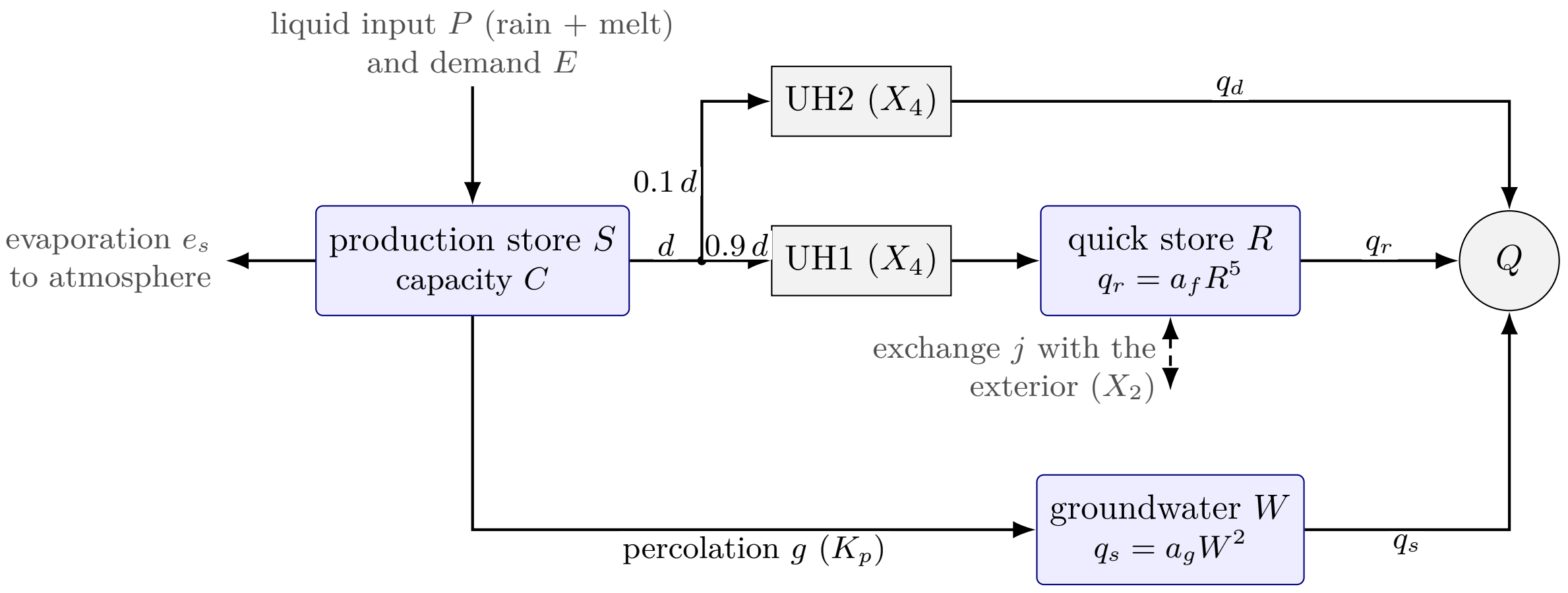
\begin{figure}[tbp]\centering
\begin{tikzpicture}[>=Latex, x=1cm, y=1cm, every node/.style={font=\small},
  store/.style={draw, rounded corners=2pt, minimum width=26mm, minimum height=11mm, align=center, fill=blue!7, draw=blue!50!black},
  uh/.style={draw, minimum width=16mm, minimum height=7mm, align=center, fill=gray!10},
  flux/.style={->, thick}, lab/.style={font=\footnotesize, inner sep=1pt, fill=white}, bnd/.style={font=\footnotesize, text=black!70}]
  \node[store] (S) at (0,0) {production store $S$\\[-1pt]\footnotesize capacity $C$};
  \coordinate (X) at (2.3,0);
  \node[uh] (U1) at (3.9,0) {UH1 ($X_4$)};
  \node[uh] (U2) at (3.9,1.6) {UH2 ($X_4$)};
  \node[store] (R) at (7.0,0) {quick store $R$\\[-1pt]\footnotesize $q_r=a_fR^{5}$};
  \node[store] (W) at (7.0,-2.7) {groundwater $W$\\[-1pt]\footnotesize $q_s=a_gW^{2}$};
  \node[draw, circle, minimum size=10mm, fill=gray!10] (Q) at (10.4,0) {$Q$};
  \draw[flux] (0,1.75) node[above, bnd, align=center] {liquid input $P$ (rain $+$ melt)\\and demand $E$} -- (S.north);
  \draw[flux] (S.west) -- ++(-0.9,0) node[left, bnd, align=right] {evaporation $e_s$\\to atmosphere};
  \draw[thick] (S.east) -- node[above, lab] {$d$} (X); \fill (X) circle (1.3pt);
  \draw[flux] (X) -- node[above, lab] {$0.9\,d$} (U1.west);
  \draw[flux] (X) |- node[pos=0.25, left, lab] {$0.1\,d$} (U2.west);
  \draw[flux] (U1.east) -- (R.west);
  \draw[flux] (U2.east) -| node[pos=0.25, above, lab] {$q_d$} (Q.north);
  \draw[flux] (R.east) -- node[above, lab] {$q_r$} (Q.west);
  \draw[flux] (S.south) |- node[pos=0.75, below, lab] {percolation $g$ ($K_p$)} (W.west);
  \draw[flux] (W.east) -| node[pos=0.25, below, lab] {$q_s$} (Q.south);
  \draw[<->, thick, densely dashed] (R.south) -- ++(0,-0.75) node[left, bnd, align=right, yshift=6pt] {exchange $j$ with the\\exterior ($X_2$)};
\end{tikzpicture}
\caption{The structure of \CaRT, without the snow store.}\label{fig:schematic}
\end{figure}

We now write a parsimonious model consistent with Laws~\ref{law:cons}--\ref{law:closure}. For a catchment represented by soil, fast-response and slow-response stores, Laws~\ref{law:cons} and \ref{law:dissip} require four classes of constitutive law: the partition of rain and the evaporation of soil water, drainage from the soil to groundwater, quick release and groundwater release. They also require a routing kernel for the water in transit to the outlet \citep{rodrigueziturbe1979,rinaldo1991}. We add the GR4J exchange with the exterior, restricted to losses, to increase accuracy. The model, \CaRT{} (Capacity--Residence-Time model), has exactly one store for each class that needs a state (the soil, the quick path and the groundwater) and uses one parameter per law and one routing scale. It descends from the GR4J family \citep{perrin2003,santos2018} but differs in the connection matrix and in the decoupling of parameters (Figure~\ref{fig:schematic}, Table~\ref{tab:laws}).

\subsection{Production, routing and storage--discharge laws}\label{sec:loss}

The production store converts liquid input $P$ and potential evaporation $E$ into direct runoff, actual evaporation and percolation, using the GR4J net-input convention $P_n=\max(P-E,0)$ and $E_n=\max(E-P,0)$. We use the partition of Section~\ref{sec:laws2} with $\eta=s^2$, the evaporation $E_a=E_{\rm dem}\,\varepsilon$ of Law~\ref{law:dissip} with $\varepsilon=2s-s^2$ and a power-law drainage. With $s=S/C$, these laws are
\begin{equation}\label{eq:production}
d=P_n\,s^2,\qquad e_s=E_n\,(2s-s^2),\qquad g=K_p\,s^5.
\end{equation}
The stores are advanced by operators applied in sequence within a day. Coefficients that depend on state or forcing are evaluated once per day and held fixed, so that conservation and the storage bounds hold exactly in the daily model. Where a law has a closed-form one-day solution, it is used, as for $\dot V=-aV^p$ with the update $V^+=V^-[1+(p-1)a(V^-)^{p-1}]^{-1/(p-1)}$, which for $p=5$ is the update of the GR4J routing store. The hyperbolic-tangent production formulae of GR4J are the exact integrals of $\dd s/\dd c=(1-s^2)/C$ (Supporting Information S2.16), with the complete daily update given in Supporting Information S1. In GR4J the percolation scale is tied to the capacity, $K_p=C(4/9)^4/4$. But Law~\ref{law:dissip} provides no reason for this tie, and the calibrated ratio $K_p/K_{p,{\rm GR4J}}$ ranges over two orders of magnitude (0.11--7.3 between the 10th and 90th percentiles), so we free the percolation scale. The direct runoff is routed by the two GR4J unit hydrographs with time base $X_4$, which send 0.9$d$ into a quick store $R$ and 0.1$d$ directly to the outlet. The quick store releases by a power law and exchanges water across the boundary,
\begin{equation}\label{eq:quick}
q_r=a_f\,R^5,\qquad j=X_2\,(R/X_3)^{3.5},
\end{equation}
with $a_f=1/(4X_3^4)$ in terms of the GR4J routing capacity $X_3$ and $X_2\le0$. This is the exchange of GR4J restricted to losses, applied once (0.9 to the quick store, 0.1 to the direct flow), with the realised flux included in the balance. The gain branch $X_2>0$ of GR4J violates Law~\ref{law:dissip} at high storage, where an inflow that grows with the storage it feeds flows against the head difference for any fixed exterior head. This positive feedback has no bound and can create water. The loss branch is the exchange with an exterior of zero head, whereas Section~\ref{sec:derive} introduces a positive exterior head. Percolation feeds a groundwater store with $q_s=a_gW^2$, $W^+=W^-/(1+a_gW^-)$, parameterised by the groundwater residence time $\tau$ at the mean recharge $\bar g$. At steady state the outflow equals $\bar g$ and the storage is $W_{\rm ss}=\bar g\tau$, so $a_g=\bar g/W_{\rm ss}^2=1/(\bar g\tau^2)$. The residence time $\tau$, in days, is then comparable between catchments whatever their recharge. The quadratic form follows from the threshold aggregate \eqref{eq:aggthr}, because groundwater leaves the aquifer through outlets such as springs and stream reaches with beds at different heights $x$ above the aquifer base. Each releases water in proportion to the head above its bed, $q_{\rm loc}=a\,(h-x)_+$. All share the water-table height $h$, to which the storage is proportional. Beds spread uniformly over the range of $h$ give $\bar q=\int_0^h a\,(h-x)\,\dd x/x_{\max}=a\,h^2/(2x_{\max})\propto W^2$, because a rising water table raises the flow of each active outlet and activates more of them, whereas beds all at the base would give the linear store. Each local flux runs down the head difference between the aquifer and a bed of fixed height and stops when the water table falls below it, so the aggregate is dissipative (Law~\ref{law:dissip}). The same exponent is the late-time drainage law of an unconfined aquifer \citep{brutsaert1977} and fits the recessions of many catchments \citep{wittenberg1999hp}.

\begin{table}[tbp]\centering\footnotesize
\caption{Components of \CaRT: the general class required by Laws~\ref{law:cons}--\ref{law:dissip}, the choice made and the parameter that controls it. The second column is fixed by the principles, and $f_R'\ge0$ is assumed. The routing kernel is the travel-time memory of units in series (Section~\ref{sec:laws3}).}\label{tab:laws}
\begin{tabular}{@{}llll@{}}
\toprule
Component & Required class & \CaRT{} choice & Parameter \\
\midrule
Rain partition & $d=P_n\,\eta(s,\dots)$, $0\le\eta\le1$ & $\eta=s^2$ & $C$ \\
Evaporation & $e_s=E_{\rm dem}\,\varepsilon(s,\dots)$, $0\le\varepsilon\le1$ & $\varepsilon=2s-s^2$ & $C$ \\
Soil drainage & $g=\mathcal G(S,W,\dots)$ & $g=K_ps^5$ & $K_p$ \\
Quick release & $q_r=f_R(R,\dots)$, $f_R\ge0$, $f_R'\ge0$ & $f_R=a_fR^5$, $a_f=1/(4X_3^4)$ & $X_3$ \\
Groundwater & $q_s=f_W(W,\dots)$ & $f_W=a_gW^2$, $a_g=1/(\bar g\tau^2)$ & $\tau$ \\
Routing & causal, non-negative, unit-mass kernel & UH1/UH2 with scale $X_4$ & $X_4$ \\
Exchange & explicit boundary term $j_{\rm ext}(\bm z,\dots)$ & $X_2(R/X_3)^{3.5}$, $X_2\le0$ & $X_2$ \\
Snow & inserted volume (Corollary~\ref{cor:refine}a, b) & five-band degree-day store & $T_T$, CFMAX, SFCF, $\Gamma$, $\kappa_s$ \\
\bottomrule
\end{tabular}
\end{table}

\paragraph{Snow.} On all three continents the liquid input $P$ comes from a degree-day snow store, a volume inserted on the precipitation path (operation a) that loses water to the atmosphere through sublimation (operation b). Because the composition of the snowpack across elevation is an $O(1)$ heterogeneity with a threshold, the store is split (operation a) into five elevation bands $k$ with temperatures $T_k=T+\Gamma(\bar h-h_k)$. In band $k$ precipitation falls as snow, $\mathrm{SFCF}\,P$, when $T_k<T_T$, and melt is $\min(N_k,\mathrm{CFMAX}\,(T_k-T_T)_+)$. Sublimation $u_k=\min(\kappa_sE,N_k)$ removes snow to the atmosphere at a fraction $\kappa_s$ of the potential evaporation, which on the prairies represents sublimation losses from blowing snow \citep{pomeroy1993}. This gives five parameters per catchment, $T_T$, CFMAX, SFCF, $\Gamma$ and $\kappa_s$. The snowfall correction SFCF (between 0.5 and 1.5) multiplies the snowfall and is therefore a correction of the input. It changes the precipitation by a median of less than 0.1\,\% on CAMELS-GB and CAMELS-US, but by up to about $\pm30\,\%$ in individual CAMELS-US basins. The bands are equal-area on CAMELS-GB, whereas on CAMELS-US they are built from the 100~m elevation bands of the dataset and are generally not equal-area (Supporting Information S1.1).

\subsection{Parameters and regionalisation}\label{sec:regional}

Each process of \CaRT{} is controlled by one parameter (Table~\ref{tab:laws}): $C$ the partition and evaporation, $K_p$ the soil drainage, $X_3$ the quick release, $\tau$ the groundwater release, $X_4$ the routing kernel and $X_2$ the exchange. The snow store adds five parameters, giving 11 per catchment. The catchment parameters are given by a static network with three hidden layers of 128 units, taking standardised catchment attributes as inputs (Supporting Information S1.7). It is trained jointly on all training catchments with the mean of $1-\mathrm{NSE}$ as loss, and the same network predicts the parameters of catchments held out from training.
 
\section{Refining the base model}\label{sec:refine}
Law~\ref{law:closure} and Corollary~\ref{cor:refine} say how \CaRT{} may be refined, and we identify in a loop from the observations where it fails and what a refinement needs. A deficit is diagnosed by conditioning the residual on inputs and model states (Section~\ref{sec:diag}), and a refinement is then derived with one admissible operation (Corollary~\ref{cor:refine}). Each refinement has one of five origins (Table~\ref{tab:refine}) and is exact when it aggregates a fine-scale assumption without approximation and asymptotic when the aggregate holds only in a limit. It is postulated when its form, together with any new state and its balance, is chosen by analogy with a known law. It is physical when it takes the form of a known physical law and heuristic when only the expected sign and bounds of its form are specified. Each refinement nests the model it refines where a natural limit exists (Section~\ref{sec:derive}). Where no such limit exists, as for the Pareto partition and the viscosity term, it is assessed by the paired comparison alone. The refinement is kept if paired multi-seed comparisons on a development split show that it improves the model overall, or improves it in some classes of catchments without degrading it overall. The diagnosis is then repeated on the new model (Section~\ref{sec:loop}). All models use the static attribute network of Section~\ref{sec:regional} for their per-catchment parameters and follow a common training protocol (Appendix~\ref{app:numerics}).

\subsection{Diagnosing deficits on observations}\label{sec:diag}

\paragraph{Conditioning.} We condition the residual $Q_{\rm obs}-Q_{\rm sim}$ on input quantities and model states, namely season, antecedent wetness, event rain depth, air temperature and the class of catchment (aridity, snow fraction, baseflow index). %
A conditional bias became a structural target when it appeared in the training period as well as the development period and was reproduced across training seeds. It also had to be predictable out of sample from the inputs, which we assessed with a per-catchment linear corrector of the residual, fitted on the training period and applied to the development period.

\paragraph{Diagnostics.} We use five diagnostics:
\begin{enumerate}
\item the seasonal volume bias by month and class;
\item the exponent of the observed dry-weather recessions, $-\dd Q/\dd t=aQ^b$, fitted on the same days to simulation and observation;
\item the volume bias of the low-flow segment of the flow-duration curve (the lowest 30\,\%), and its log-space form FLV of \citet{yilmaz2008};
\item the event runoff ratio against rain depth and antecedent wetness, with events selected on rain;
\item the sensitivity of melt timing to air temperature, from the shift of the spring flow centroid per degree of spring warming between years and between periods.
\end{enumerate}

\paragraph{Findings.} We located each deficit by continent and class of catchment. On both continents the error of the quick flow depended on antecedent wetness and on the season, and large events had approximately correct volumes but flatter peaks than observed. On CAMELS-US the quick-flow recession was slower than observed, and the rain-fed basins had a seasonal phase error and recessions that did not steepen with the evaporative demand. Flow was overestimated in the prairie basins, as were low flows in the dry basins, and snowmelt was too sensitive to air temperature. On CAMELS-GB, once connectivity had been added to the quick release, runoff from small events was overestimated and the event runoff ratio rose too slowly with rain depth. Flow was overestimated in summer in the high-baseflow catchments, and the spring-fed chalk and limestone streams lacked water that reaches them from beyond their divide.

\subsection{Refining \CaRT{} under the laws}\label{sec:derive}

\begin{figure}[tbp]\centering
\includegraphics[width=\textwidth]{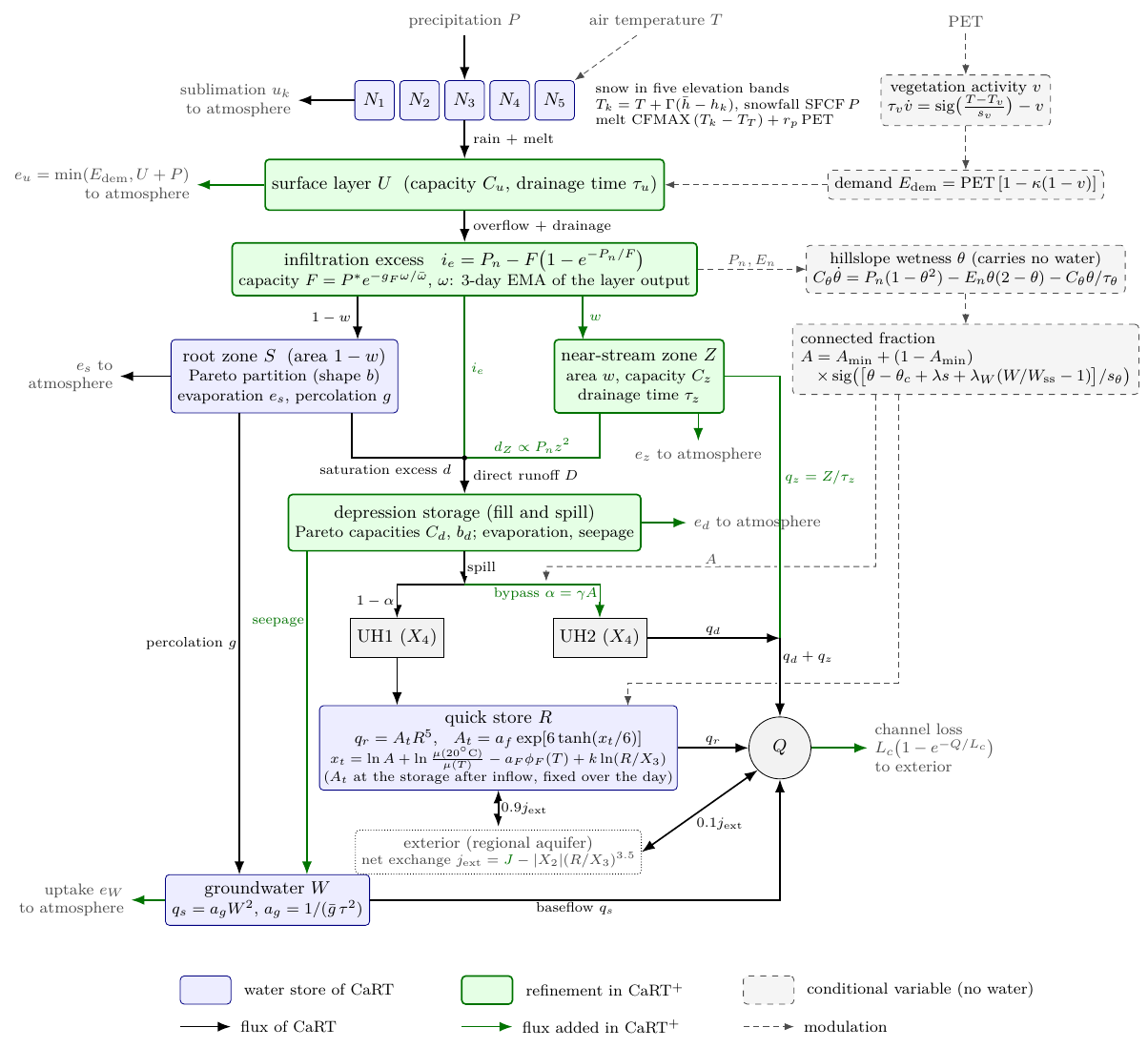}
\caption{The structure of \CaRTp.}\label{fig:cartp}
\end{figure}

Each refinement below starts from a deficit diagnosed as in Section~\ref{sec:diag} on the model before it, takes a fine-scale hypothesis from the literature and derives its catchment-scale law with one admissible operation (Corollary~\ref{cor:refine}). Where a parameter limit of the refinement recovers the model before it, that limit is given (Table~\ref{tab:refine}), and a single static attribute network supplies all parameters. Figure~\ref{fig:cartp} shows the resulting structure, and Table~\ref{tab:refine} lists the refinements and states which parts are exact aggregates and which are postulated. The connection matrix is in Appendix~\ref{app:matrix}, while Supporting Information S2 and S1 provide the derivation of each refinement and the daily update used in the implementation. %

\paragraph{Connectivity of the quick release (operations c, d).} The quick-flow recession of \CaRT{} was slower than observed on CAMELS-US, and on both continents the event-scale error depended on antecedent wetness, with quick flow under-simulated after dry spells and over-simulated after wet spells. Field studies and syntheses describe hillslopes that release water to the channel only once their wetness exceeds a threshold, with thresholds that vary from place to place, so that the connected fraction of the hillslope grows with wetness \citep{western2001,tromp2006,bracken2007,jencso2009,penna2011}. Let the hillslope consist of units $i$ that share the quick-flow depth $R$ (lateral redistribution is faster than release to the channel) and release only when their wetness exceeds a unit threshold $\theta_{c,i}$. A fraction $A_{\min}$ next to the channel is always connected. Units with the same balance under the same forcing converge to a shared wetness $\theta$, and lateral exchange keeps their deviations small where the forcing or the units vary in space (Appendix~\ref{app:memory}). With the shared wetness, summing the units yields the exact relation $q_r=a_fR^5[A_{\min}+(1-A_{\min})F_\theta(\theta)]$, i.e.\ $m=\ln A(t)$. We take a logistic threshold distribution with a centre that shifts with soil wetness $s$, $A=A_{\min}+(1-A_{\min})\sg((\theta-\theta_c+\lambda s)/s_\theta)$, with $\lambda\ge0$ because a wetter soil can only lower the thresholds. For wetness, a conditional variable that carries no water, we postulate a balance of the same form as that of the production store, driven by the net input and demand that reach the root zone, with a loss term,
\begin{equation}\label{eq:theta}
C_\theta\dot\theta=P_n(1-\theta^2)-E_n\theta(2-\theta)-C_\theta\theta/\tau_\theta .
\end{equation}
Linearised about its mean, $\theta$ is an exponential moving average of net rain and demand, weighted by the mean wetness, with a time constant set by the catchment and its mean forcing (Appendix~\ref{app:memory}). The wetness $\theta$ therefore carries the memory of the history of rain and demand. %

\paragraph{Shape and temperature of the release (operations c, d).} On both continents large events had about the right volume but flatter peaks than observed, and quick flow was overestimated in winter and underestimated in summer, errors that we attribute to the release law. Aggregating the GR4J release $a(R-x)_+^5$ over depth thresholds $x$ with $F_x\propto x^k$ at small depth, the integral \eqref{eq:aggthr}, gives the exponent $5+k$ within the power-law range, with $k$ the small-depth power of the threshold distribution. The release $q_r=a_fe^mR^5(R/X_3)^k$ with $k\ge0$ steepens at high storage and sharpens the peaks. The hydraulic conductivity of the quick-flow pathways is inversely proportional to the viscosity of water, $K=\kappa\rho g/\mu$ with $\kappa$ the intrinsic permeability, so the log-multiplier gains $-\ln[\mu(T)/\mu(\SI{20}{\celsius})]$ without free parameters \citep{constantz1982,huber2009}. Air temperature stands in for water temperature and is clamped at \SI{0}{\celsius}. Seasonally frozen soil reduces infiltration and the conductivity of shallow subsurface paths, with effects on runoff that depend on soil, snow and climate \citep{shanley1999,stahli1999,bayard2005,iwata2010,alaaho2021}. In the model, frost raises the thresholds of the quick release by the frozen depth. We represent this rise by a storage-independent step of size $a_F\ge0$ in the log-multiplier, $-a_F\sg((T_F-T)/s_F)$, so that frost can only suppress the release. Here $T$ is the air temperature and $T_F$, $s_F$ are global constants (learned near \SI{0}{\celsius} and \SI{2.3}{\celsius}). %

\paragraph{Near-stream saturated zone (operations a, b, parallel split).} On both continents the runoff ratio of events after dry spells was underestimated, as was that of small events in the arid basins of CAMELS-US. The model needed a part of the catchment that responds to each rainfall event regardless of hillslope state. This is the variable source area, a saturated zone next to the channel with an extent that varies with wetness \citep{hewlett1967,dunne1970,freeze1972,bevenkirkby1979}. A fraction $w$ of the soil volume next to the channel is therefore given a small capacity $C_z$ and a water table close to the surface, with storage governed by $\dot Z=P_n(1-z^2)-E_nz(2-z)-Z/\tau_z$, $z=Z/C_z$. Saturation excess $P_nz^2$ is routed as direct runoff, and drainage $Z/\tau_z$ goes straight to the outlet. The limit $w\to0$ recovers \CaRT, and the limiting cases of this family include an impervious area, a linear interflow store and the variable source area. The zone's flux laws are postulated by analogy with the production store.

\paragraph{Surface layer (operation a, series split).} On CAMELS-GB, once the release was connected, runoff from small events was overestimated. The first millimetres of each event are held by canopy, litter and the top soil and returned to the atmosphere at the demand rate by interception and stage-one soil evaporation \citep{rutter1971,ritchie1972}. We represent these processes by splitting the soil volume into a thin surface layer $U$ and the root zone $S$ below it. The layer receives the liquid input, evaporates at the demand rate, passes on water exceeding the capacity $C_u$ and drains the rest linearly with time constant $\tau_u$. Overflow and drainage enter the root zone with the remaining demand. Merging $U$ and $S$ recovers the columns of \CaRT, and $\tau_u\to0$ recovers the \CaRT{} same-day interception $\min(P,E)$ exactly once the layer is empty (from the second day on) \citep[cf.\ the interception store of][]{ficchi2019}. %

\paragraph{Partition and infiltration excess (operations c, d).} On CAMELS-GB the event runoff ratio rose too slowly with rain depth. Where storage capacities and infiltration capacities are distributed within the catchment, saturation excess \citep{moore1985,zhao1992} and infiltration excess \citep{horton1933,philip1957,hawkins1987} grow gradually with rain depth. Section~\ref{sec:laws2} derives exact expressions for both aggregates when rain is uniform over the catchment and the day. The root-zone partition $\eta=s^2$ is replaced by the Pareto aggregate \eqref{eq:pareto} with shape $b$, a different member of the same class that does not nest $\eta=s^2$. Infiltration excess over exponentially distributed capacities, $i_e=P_n-F(1-e^{-P_n/F})$, is taken off the water that leaves the surface layer and routed with the direct runoff. This excess is the integral \eqref{eq:aggthr} over the capacities and is exact for rain uniform in space and in time within the day. The mean capacity falls with a near-surface wetness $\omega$ shared by the catchment, and because $\omega$ scales all local capacities alike, the aggregate stays exact (Appendix~\ref{app:memory}). The wetness $\omega$ is an exponential moving average of the water leaving the surface layer with a time constant of 3 days, which is the exact daily solution of a linear balance driven by that water. The scale function in $F=P^*e^{-g_F\omega/\bar\omega}$ is heuristic, with $\bar\omega$ the scale of $\omega$, and this construction is an example of how memory effects enter the threshold distribution function. %

\paragraph{Demand and groundwater uptake (operations d, b).} In the rain-fed basins of CAMELS-US flow was underestimated in spring and overestimated in autumn, a phase error that follows the greening and senescence of the vegetation. The observed dry-weather recessions steepened with the evaporative demand, whereas the simulated ones did not. We made the evaporative demand follow the seasonal activity of vegetation, which lags air temperature \citep{chuine2000}, and added groundwater uptake by riparian vegetation \citep{loheide2005}. The demand becomes $E_{\rm dem}=\mathrm{PET}[1-\kappa(1-v)]$, where the vegetation-activity variable $v$ is a conditional variable that relaxes towards $\sg((T-T_v)/s_v)$ with time constant $\tau_v$. Groundwater uptake adds a flux from $W$ to the atmosphere, $e_W=\lambda_EE_rW/(W+W_{\rm ss})$, with $\lambda_E\le1$, where $E_r$ is the remaining demand and $W_{\rm ss}$ the steady storage of the store. The demand is the exact area average over vegetation patches that switch at a common rate towards thermal thresholds (Appendix~\ref{app:memory}), and the groundwater-uptake relation is a heuristic closure. Sublimation and the melt term below use the potential evaporation itself.

\paragraph{Routing of direct runoff (operation d).} GR4J sends a fixed 10\,\% of direct runoff past the quick store. However, under the connectivity hypothesis above, surface runoff reaches the channel only from connected units. On a connected unit it reaches the channel directly with probability $\gamma$ and otherwise infiltrates into the unit's quick-flow layer, while on a disconnected unit it re-infiltrates downslope. If runoff production is independent of the connection thresholds, the bypass fraction aggregates exactly to $\alpha_t=\gamma A(\theta_t)$ for hillslope runoff and to $\gamma$ for the near-stream zone, which is always connected. In \CaRTp{} the runoff of all sources is mixed in the depressions, and $\gamma A(\theta_t)$ is applied to the whole spill. The values $A\equiv1$ and $\gamma=0.1$ recover the GR4J split, so the constant is replaced by one parameter and this connected fraction now controls both the release and the routing of quick flow. Because $A$ is shared with the release, this limit recovers the previous model where its release is fully connected. 

\paragraph{Depression storage (operations a, b, series split).} Streamflow was overestimated in the low-relief prairie basins of CAMELS-US, most in autumn, in the low flows and in the events of winter and spring. In such landscapes surface depressions fill before they spill, and much of the catchment does not contribute to streamflow until they are full \citep{shook2011}. Direct runoff therefore first fills depressions with capacities that follow a Pareto distribution. The exact spill is calculated by applying the aggregation \eqref{eq:pareto} to depression capacities, each of which spills to the flow path on its own. The stored water seeps to the groundwater store and evaporates at a fraction of the remaining demand, a fraction that falls as the depressions empty. A vanishing mean capacity recovers the model without depressions.

\paragraph{Exterior inflow (operation c).} The spring-fed chalk and limestone streams of CAMELS-GB receive groundwater from beyond their topographic divide, and some, such as Costa Beck, carry more water than falls on their catchment. \CaRT{} cannot supply this water because the exchange is restricted to losses. Law~\ref{law:dissip} admits an inflow across the boundary only down a head difference. For a flux that does not decrease as the head difference increases and an exterior of fixed head, the exchange is $j_{\rm ext}=L\,[H_{\rm ext}-g(R/X_3)]$, $g(r)=r^{3.5}$, with a conductance $L\ge0$, and the exchange of \CaRT{} is the case $H_{\rm ext}=0$, $X_2=-L$. A positive exterior head adds a constant inflow $J=LH_{\rm ext}$, so the store gains water when it is low and loses it when it is high, with a stable equilibrium in between. This adds to the exchange of GR4J the threshold at which the exchange changes sign, as in GR5J and GR6J \citep{pushpalatha2011}. In the product form $LH_{\rm ext}$ the inflow vanishes with the loss and did not supply the spring-fed catchments, so we learn $J$ as a separate parameter, with $J\to0$ recovering \CaRT. The inflow uses the connection between the quick store and the exterior that the loss already has, and so changes the law of that flux.

\paragraph{Groundwater-controlled connectivity (operation d).} Summer flow was overestimated in the high-baseflow catchments of CAMELS-GB, and neither the demand nor the uptake removed this excess. In chalk and limestone catchments the stream network itself expands and contracts with the water table, which rises above the beds of the headwater reaches in winter and falls below them in summer (winterbournes; \citealp{sear1999}). Connectivity therefore depends on the groundwater head as well as on the soil. Shifting the centre of the connection thresholds by $\lambda_W(W/W_{\rm ss}-1)$, with $W_{\rm ss}$ the steady storage of the groundwater store, gives $A=A_{\min}+(1-A_{\min})\sg([\theta-\theta_c+\lambda s+\lambda_W(W/W_{\rm ss}-1)]/s_\theta)$. The model without this shift is $\lambda_W=0$, and we require $\lambda_W\ge0$ because a higher head can only connect more of the network.

\paragraph{Melt energy term (operation d).} In the snow-dominated basins of CAMELS-US snowmelt was too sensitive to air temperature, the spring flow centroid moving more with spring warming than observed. In spring the melt energy rises with radiation faster than with air temperature, so a degree-day factor alone overstates the temperature sensitivity, and enhanced temperature-index models add a radiation term \citep{kustas1994,hock2003,pellicciotti2005}. The only energy variable among the inputs is the potential evaporation, so the melt of band $k$ becomes
\begin{equation*}
M_k=\min\{N_k,\ \mathrm{CFMAX}\,(T_k-T_T)_++r_p\,\mathrm{PET}\}\quad\text{on days with }T_k\ge T_T ,
\end{equation*}
which lowers the temperature sensitivity, with the degree-day law recovered in the limit $r_p\to0$.

\paragraph{Channel transmission loss (operation c).} Low flows in the dry basins of CAMELS-US were overestimated and came mostly from the tail of the quick store, and the model had no sink for this flow. Ephemeral and intermittent channels lose water to their beds, with a loss capacity that varies along the network \citep{shanafield2014}. Let the loss sites act in parallel on the flow, with bed-loss capacities that are exponentially distributed with mean $L_c$, each site taking the smaller of its capacity and the flow offered to it. The exact aggregate loss of a flow $Q$, $L_c(1-e^{-Q/L_c})$, is obtained by aggregation over thresholds (operation c) and takes nearly all of a small flow and a bounded part of a large one. It is a flux from the outlet to the exterior, a connection that the exchange of \CaRT{} already uses on the direct path. The limit $L_c\to0$ recovers the model without it.

\begin{table}[tbp]\centering\small
\caption{Refinements of \CaRT{} that make up \CaRTp. Op.: operation of Corollary~\ref{cor:refine}. Origin: how the law is obtained, either exact (exact aggregate of a local law under the stated assumption), asymptotic (exact only in a limit), postulated, physical or heuristic (Section~\ref{sec:refine}). Parameters are per catchment except where marked global (g).}\label{tab:refine}
{\footnotesize\setlength\tabcolsep{3pt}\begin{tabular}{@{}>{\raggedright\arraybackslash}p{2.8cm}l>{\raggedright\arraybackslash}p{4.2cm}>{\raggedright\arraybackslash}p{3.2cm}>{\raggedright\arraybackslash}p{2.4cm}>{\raggedright\arraybackslash}p{1.9cm}@{}}
\toprule
Refinement & Op. & Fine-scale assumption & Origin & Parameters & Nested limit \\
\midrule
Connected fraction of quick release & c, d & connection thresholds, shared depth, wetness converging to a shared value; wetness balance \eqref{eq:theta} & exact on the slow manifold; logistic shape, shift and balance postulated & $A_{\min}$, $C_\theta$, $\tau_\theta$, $\lambda$; $\theta_c$, $s_\theta$ (g) & $A_{\min}\to1$ \\
Release exponent & c & power-law depth thresholds & asymptotic & $k$ & $k=0$ \\
Viscosity & d & conductance $\propto1/\mu(T)$ & physical & none & none (no free parameter) \\
Frost & d & frozen depth raises thresholds & heuristic & $a_F$; $T_F$, $s_F$ (g) & $a_F=0$ \\
Near-stream saturated zone & a, b & small-capacity zone, shallow water table & postulated & $w$, $C_z$, $\tau_z$ & $w=0$ \\
Surface layer & a & evaporation at demand, linear drainage & postulated & $C_u$, $\tau_u$ & $\tau_u\to0$ \\
Pareto partition & c & Pareto point capacities, uniform rain & exact for filling & $b$ & none (replaces $\eta=s^2$) \\
Infiltration excess & c, d & exponential infiltration capacities scaled by a shared near-surface wetness, uniform rain & exact; wetness balance postulated, scale function heuristic & $P^*$, $g_F$ & $P^*\to\infty$ \\
Vegetation demand & d & patches switching at a common rate towards thermal thresholds & exact; switching law and threshold distribution postulated & $\kappa$; $T_v$, $s_v$, $\tau_v$ (g) & $\kappa=0$ \\
Groundwater uptake & b & riparian uptake from $W$ & heuristic & $\lambda_E$ & $\lambda_E=0$ \\
Connected bypass & d & connected units pass surface runoff, others re-infiltrate & exact for hillslope runoff; applied to all runoff & $\gamma$ & $A\equiv1$, $\gamma=0.1$ \\
Depression storage & a, b & Pareto depression capacities, fill and spill & exact while filling; losses postulated & $C_d$, $b_d$, $k_d$ & $C_d\to0$ \\
Exterior inflow & c & exchange driven by the head difference with an exterior of fixed head & physical & $J$ & $J\to0$ \\
Groundwater-controlled connectivity & d & thresholds shift with groundwater head & postulated & $\lambda_W$ & $\lambda_W=0$ \\
Melt energy term & d & melt energy rises with radiation faster than with temperature & heuristic; energy argument physical & $r_p$ & $r_p\to0$ \\
Channel transmission loss & c & exponential bed-loss capacities along the network & exact & $L_c$ & $L_c\to0$ \\
\bottomrule
\end{tabular}}
\end{table}

\paragraph{Summary of \CaRTp.} \CaRTp{} extends the network of \CaRT{} with operations (a) and (b) of Corollary~\ref{cor:refine}. Splitting volumes adds a surface layer above the root zone, a near-stream zone beside it and surface depressions on the path of direct runoff. New boundary fluxes connect stores to the atmosphere and the outlet through groundwater uptake, the evaporation of the depressions and the drainage of the near-stream zone. Operations (c) and (d) change the laws on this network, and most of these changes take their form from one of two constructions (Figure~\ref{fig:cartp}, Table~\ref{tab:refine}). Threshold aggregation (Law~\ref{law:closure}) gives one formula, \eqref{eq:aggthr}, for the saturation excess of the root zone, the infiltration excess, the release exponent, the fill-and-spill of the depressions, the transmission loss of the channel and the connected fraction of the hillslope. Already in \CaRT{}, this formula produces the quadratic groundwater law. Variables that carry no water modulate the fluxes (operation d). Hillslope wetness, near-surface wetness and vegetation activity follow the forcing with lags set by their own time constants, which is the memory form of Law~\ref{law:closure}. Of the memory forms of Table~\ref{tab:memory}, \CaRTp{} realises the water in transit of the unit hydrographs and the shared driving variables $\theta$, $\omega$ and $v$. However, the turning points of units that fill and spill, such as the root zone and the depressions, are not represented. The variables $\theta$, $\omega$ and $v$, together with the connected fraction, the groundwater head, the viscosity of water and frost, enter only as multipliers of four flux coefficients, those of the quick release, the bypass, the infiltration capacity and the evaporative demand. Law~\ref{law:dissip} fixes the directions, so that the exchange is a loss plus an inflow from a fixed exterior head, and the threshold shifts and the frost step act one way. The evaporation of the stores and the groundwater uptake are bounded by the demand.

The state of \CaRTp{} has fourteen variables, eleven of which are water stores: five snow bands, the surface layer, root zone, near-stream zone, depressions, quick store and groundwater store. The three conditional variables are hillslope wetness, near-surface wetness and vegetation activity, while the unit hydrographs hold additional water in transit. \CaRTp{} has 35 parameters per catchment and seven global constants, one value per dataset for the frost step, the connection thresholds and the vegetation activity (Supporting Information S1). The per-catchment parameters are regionalised by one static network that takes 21 (CAMELS-GB) or 27 (CAMELS-US, CAMELS-AUS) standardised catchment attributes as inputs. The network is trained jointly on the training catchments of a dataset and applied unchanged to the held-out catchments.

\subsection{The refinement loop and the role of AI agents}\label{sec:loop}

\begin{figure}[tbp]\centering
\includegraphics[width=\textwidth]{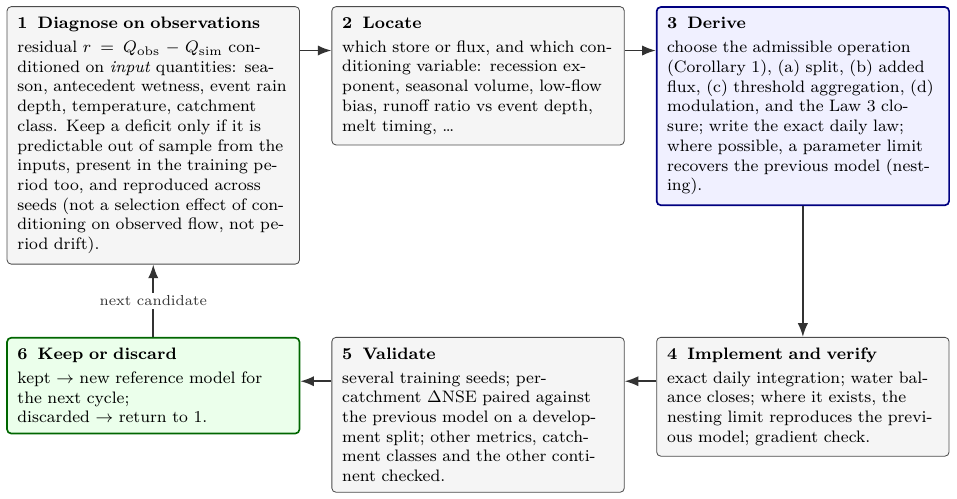}
\caption{The refinement loop.}\label{fig:loop}
\end{figure}

\begin{figure}[tbp]\centering
\includegraphics[width=\textwidth]{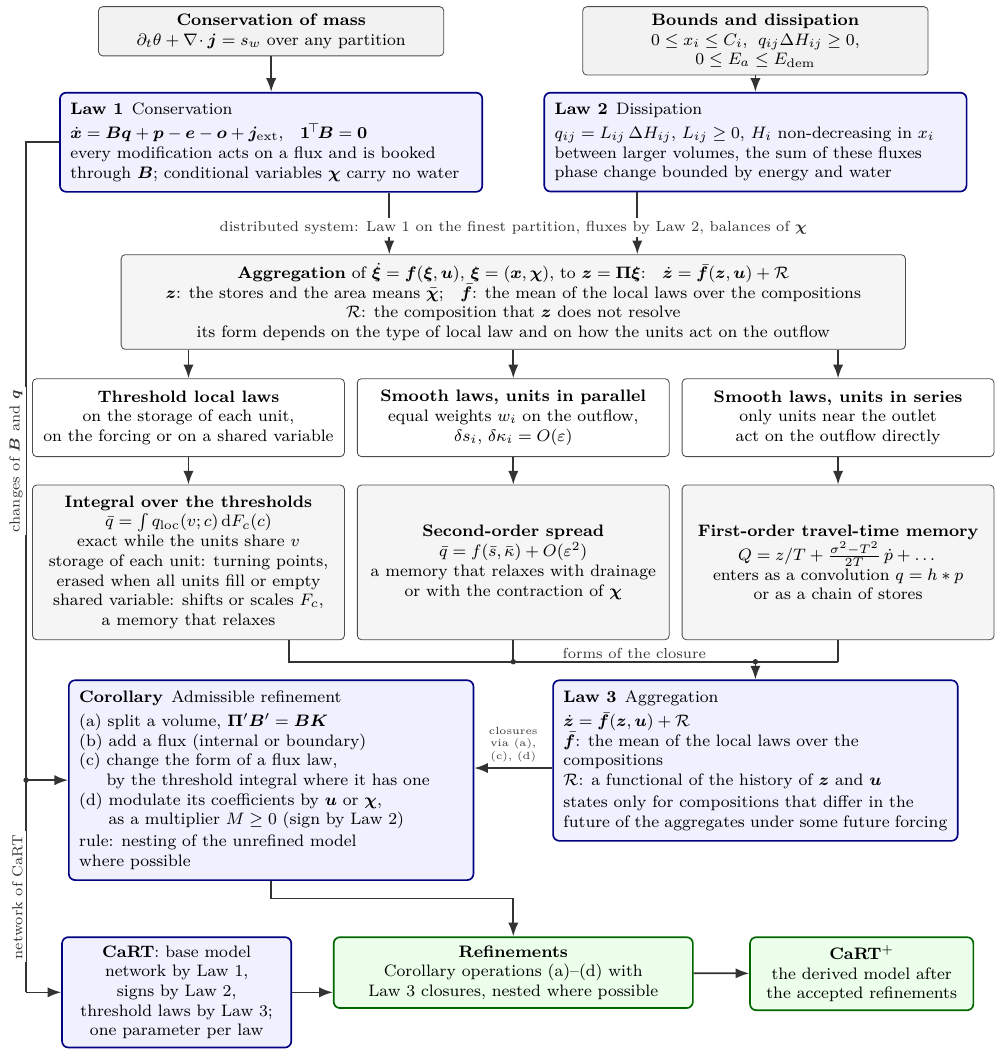}
\caption{How the laws guide refinement.}\label{fig:lawsflow}
\end{figure}

Because the laws restrict the candidates, we ran the refinement as an iterative loop with the help of AI agents. Figure~\ref{fig:loop} shows the loop, and Figure~\ref{fig:lawsflow} shows where the laws enter it. A deficit is diagnosed on the observations (Section~\ref{sec:diag}) and attributed to a store or flux and to the variable that conditions it. A refinement is derived with one admissible operation (Corollary~\ref{cor:refine}) and, where the deficit depends on history, with a closure of Law~\ref{law:closure}, nesting the previous model where a limit exists. The refinement is then trained with several seeds and compared catchment by catchment on development splits with the model it refines. On CAMELS-US, training runs from October 1980 to September 1995 and development from October 1995 to September 2010, and the corresponding periods on CAMELS-GB are 1989--1999 and 2000--2010, all before the test period 2011--2019 of Section~\ref{sec:results}. The refinement is kept if it improves the model overall, or in some classes of catchments without degrading it overall, and the diagnosis is repeated.

The loop was executed by AI agents (Claude Code from Anthropic and Codex from OpenAI) under the author's instructions and the constraints of Laws~\ref{law:cons}--\ref{law:closure}. The agents mainly shortened each cycle, so that a diagnosed deficit became a verified, trained candidate within hours and more admissible hypotheses could be tried. The author guided the search throughout and often redirected the agents on the basis of an understanding of the processes.
 
\section{Results on CAMELS-GB, CAMELS-US and CAMELS-AUS}\label{sec:results}
\subsection{Data, protocol and benchmarks}\label{sec:setup}

\paragraph{Training protocol.} All models on the three continents are trained on 1989-01-01 to 2008-12-31 with a spin-up period (details in Appendix~\ref{app:numerics}), and checkpoints or epochs are selected on the median NSE over 2009--2010. The models are evaluated on the held-out period 2011-01-01 to 2019-12-31, which was not used to fit the parameters or to select checkpoints or epochs. One structural choice, the snow store of CAMELS-GB, was first adopted after results for this period had been seen and was confirmed afterwards on the development split. The other structural choices of Section~\ref{sec:refine} were made on the development splits, and the seed-averaging analysis of Section~\ref{sec:acc} uses the test period after the structure was fixed. On each continent 10\,\% of the catchments are held out from training and selection altogether and predicted from their attributes alone (\emph{ungauged}), while the remaining catchments are \emph{gauged}. All regional models are trained with three seeds. Each metric is computed for each seed and catchment and averaged over the seeds, after which we report the median of this seed-mean metric over catchments. Differences between models are tested using the paired per-catchment difference, with 95\,\% bootstrap intervals for the median paired difference (resampling catchments only, with seeds held fixed). Comparisons of single models set one trained structure against another, and the metrics of the three-seed ensemble (mean of the simulations) are reported alongside.

\paragraph{CAMELS-GB.} Daily catchment-mean precipitation, air temperature, potential evapotranspiration with interception correction and specific discharge are available for the 671 catchments of CAMELS-GB \citep{coxon2020camelsgb}, with discharge after September 2015 taken from the extended series. The data also include 21 static attributes (area, elevation, soil texture, porosity, conductivity and depth, land cover, reservoir capacity, climate indices). Five band elevations are used by the conceptual models in their snow store and supplied to the LSTMs as attributes. The ungauged set is 67 catchments of CAMELS-GB, most of which have short discharge records before 2009 and 64 of which have at least one year of test data.

\paragraph{CAMELS-US.} We use the 671 CAMELS basins \citep{newman2015,addor2017camels} with the 27 static attributes of \citet{kratzert2019hess} and the five band elevations, which are used as described for CAMELS-GB. The forcing is Daymet version~4 \citep{thornton2021daymet4}, averaged over the CAMELS basin polygons for 1980--2019, with potential evaporation from the Priestley--Taylor equation \citep{priestley1972}. Discharge is taken from the CAMELS record to 2014 and extended to 2019 from the USGS National Water Information System. Version~4 is used because in the version distributed with CAMELS most of a day's rain is recorded on the following day (Appendix~\ref{app:numerics}). The ungauged set is 68 basins (10\,\%) drawn at random within strata of snow fraction and aridity.

\paragraph{CAMELS-AUS.} To test the transfer to a continent that played no part in development, we add the 561 catchments of CAMELS-AUS version~2 \citep{fowler2025camelsaus} that have at least one year of discharge in 1989--2008. Precipitation and air temperature are those of the Australian Gridded Climate Data and potential evapotranspiration is the Morton wet-environment estimate of SILO, all as distributed with the dataset, for 1980--2019. The 27 static attributes cover the same climate, topographic, land-cover, soil and geological properties as the CAMELS-US set, with the seasonal amplitude of net primary productivity in place of that of leaf area. Band elevations are not provided with the dataset and are drawn from a normal hypsometry centred on the mean elevation of each catchment and truncated at its lowest and highest elevation. The ungauged set is 56 catchments (10\,\%) drawn at random within strata of aridity and elevation.

\paragraph{Benchmarks and metrics.} We use four models as benchmarks.

\begin{itemize}\setlength\itemsep{0pt}
\item \emph{HBV}: the HBV-light structure \citep{bergstrom1995,seibert2012} with its snow routine on the same five elevation bands, 15 parameters per catchment, implemented as differentiable kernels and regionalised exactly as \CaRT{}, with static parameters and a single component.
\item \emph{dHBV}: the differentiable HBV of \citet{feng2022}, re-implemented on our data. It has sixteen parallel HBV components with daily $\beta$ and $\beta_{\rm ET}$, whose parameters come from an LSTM that takes the forcing and attributes as inputs. The original minimises a combination of root-mean-square errors and uses Hargreaves potential evaporation. Here the model is trained with the NSE loss and receives the same potential evaporation as the other models.
\item \emph{LSTM}: a stateful LSTM (256 units) that takes the daily precipitation, mean air temperature and potential evaporation and the same static attributes as inputs, run continuously over the record and trained by truncated back-propagation through time on all three continents (Appendix~\ref{app:numerics}).
\item \emph{EA-LSTM}: the entity-aware LSTM of \citet{kratzert2019hess}, in which the static attributes set the input gate, with the inputs of the stateful LSTM and otherwise the configuration of that paper, except for the length of the input sequences (Appendix~\ref{app:numerics}).
\end{itemize}

\CaRT, HBV and \CaRTp{} are trained for 5000 iterations with the same optimiser (Appendix~\ref{app:numerics}), while dHBV and the two LSTMs use their own schedules, and every model keeps the checkpoint or epoch with the highest median NSE over the selection period. The metrics include NSE, KGE \citep{gupta2009}, NSE of log flow, the flow-duration-curve biases FHV and FLV \citep{yilmaz2008}, the volume bias of the low-flow segment (the lowest 30\,\% of the flow-duration curve) and the number of catchments with NSE $<0$. We also report medians over classes of catchments (by baseflow index and snow fraction on CAMELS-GB and by snow fraction and aridity on CAMELS-US).

\subsection{Accuracy on the held-out period}\label{sec:acc}

\begin{table}[tbp]\centering\footnotesize
\caption{CAMELS-GB, held-out period 2011--2019, medians over the 602 gauged catchments (64 ungauged in the columns so marked). Single model: median NSE of each seed, and medians of the per-catchment metrics averaged over the three seeds; 3-seed ensemble: metrics of the mean simulation.}\label{tab:u_gb}
\resizebox{\textwidth}{!}{%
\begin{tabular}{lccccccccc|cccc}
\toprule
 & \multicolumn{9}{c|}{single model} & \multicolumn{4}{c}{3-seed ensemble} \\
Model & NSE (seeds 0/1/2) & NSE & NSE$_{\log}$ & KGE & FHV (\%) & FLV (\%) & low-flow volume (\%) & $n$(NSE$<$0) & NSE ungauged & NSE & NSE$_{\log}$ & KGE & NSE ungauged \\
\midrule
HBV & 0.825 / 0.825 / 0.825 & 0.825 & 0.799 & 0.823 & -9.6 & -86.9 & 17.6 & 7 & 0.735 & 0.827 & 0.808 & 0.824 & 0.739 \\
dHBV & 0.834 / 0.833 / 0.834 & 0.832 & 0.834 & 0.827 & -10.9 & -52.6 & 10.3 & 14 & 0.739 & 0.837 & 0.844 & 0.829 & 0.745 \\
\CaRT & 0.845 / 0.846 / 0.846 & 0.845 & 0.857 & 0.840 & -11.1 & -14.0 & 3.9 & 5 & 0.700 & 0.846 & 0.860 & 0.841 & 0.707 \\
\CaRTp & 0.864 / 0.864 / 0.862 & 0.863 & 0.883 & 0.856 & -10.3 & -16.2 & 7.5 & 0 & 0.755 & 0.866 & 0.893 & 0.857 & 0.766 \\
LSTM & 0.871 / 0.870 / 0.867 & 0.869 & 0.900 & 0.871 & -7.6 & -27.6 & 6.7 & 1 & 0.723 & 0.879 & 0.921 & 0.880 & 0.737 \\
EA-LSTM & 0.863 / 0.861 / 0.865 & 0.862 & 0.892 & 0.856 & -9.5 & -18.6 & 10.7 & 1 & 0.714 & 0.871 & 0.906 & 0.859 & 0.756 \\
\bottomrule
\end{tabular}
}
\end{table}

\begin{table}[tbp]\centering\footnotesize
\caption{CAMELS-US, held-out period 2011--2019, as Table~\ref{tab:u_gb}, for 593 gauged basins with at least one year of test data and 68 ungauged.}\label{tab:u_us}
\resizebox{\textwidth}{!}{%
\begin{tabular}{lccccccccc|cccc}
\toprule
 & \multicolumn{9}{c|}{single model} & \multicolumn{4}{c}{3-seed ensemble} \\
Model & NSE (seeds 0/1/2) & NSE & NSE$_{\log}$ & KGE & FHV (\%) & FLV (\%) & low-flow volume (\%) & $n$(NSE$<$0) & NSE ungauged & NSE & NSE$_{\log}$ & KGE & NSE ungauged \\
\midrule
HBV & 0.690 / 0.690 / 0.688 & 0.690 & 0.447 & 0.658 & -11.6 & -88.3 & 73.2 & 23 & 0.632 & 0.694 & 0.458 & 0.660 & 0.646 \\
dHBV & 0.715 / 0.715 / 0.715 & 0.713 & 0.591 & 0.680 & -11.0 & 5.8 & 59.7 & 17 & 0.677 & 0.719 & 0.605 & 0.689 & 0.681 \\
\CaRT & 0.723 / 0.724 / 0.725 & 0.723 & 0.658 & 0.726 & -9.0 & 18.4 & 50.2 & 24 & 0.635 & 0.725 & 0.661 & 0.729 & 0.652 \\
\CaRTp & 0.760 / 0.768 / 0.765 & 0.765 & 0.744 & 0.784 & -6.9 & 6.7 & 29.9 & 21 & 0.674 & 0.774 & 0.767 & 0.789 & 0.689 \\
LSTM & 0.761 / 0.753 / 0.752 & 0.751 & 0.773 & 0.773 & -5.7 & -44.1 & 17.0 & 17 & 0.719 & 0.781 & 0.817 & 0.792 & 0.754 \\
EA-LSTM & 0.751 / 0.772 / 0.752 & 0.754 & 0.747 & 0.767 & -8.2 & -29.5 & 44.7 & 13 & 0.722 & 0.773 & 0.779 & 0.772 & 0.749 \\
\bottomrule
\end{tabular}
}
\end{table}

\begin{table}[tbp]\centering\footnotesize
\caption{Paired per-catchment differences on the held-out period (median and 95\,\% bootstrap interval). Single model: metrics averaged over the three seeds in each catchment; 3-seed ensemble: NSE of the mean simulation. Top: CAMELS-GB; bottom: CAMELS-US.}\label{tab:u_paired}
\resizebox{\textwidth}{!}{%
\begin{tabular}{lcccc|cc}
\toprule
 & \multicolumn{4}{c|}{single model} & \multicolumn{2}{c}{3-seed ensemble} \\
Pair & $\Delta$NSE & $\Delta$NSE$_{\log}$ & $\Delta$KGE & $\Delta$NSE ungauged & $\Delta$NSE & $\Delta$NSE ungauged \\
\midrule
\CaRTp $-$ LSTM & $-0.004$ [-0.006, -0.003] & $-0.013$ [-0.015, -0.010] & $-0.014$ [-0.018, -0.012] & $-0.002$ [-0.010, +0.010] & $-0.011$ [-0.012, -0.009] & $-0.004$ [-0.018, +0.008] \\
\CaRTp $-$ EA-LSTM & $+0.001$ [-0.001, +0.002] & $-0.006$ [-0.009, -0.004] & $-0.002$ [-0.005, +0.002] & $+0.004$ [-0.011, +0.020] & $-0.003$ [-0.004, -0.002] & $-0.004$ [-0.019, +0.012] \\
\CaRTp $-$ \CaRT & $+0.015$ [+0.013, +0.016] & $+0.023$ [+0.018, +0.027] & $+0.013$ [+0.011, +0.015] & $+0.017$ [+0.003, +0.024] & $+0.016$ [+0.014, +0.017] & $+0.017$ [+0.006, +0.027] \\
\CaRTp $-$ dHBV & $+0.023$ [+0.022, +0.026] & $+0.038$ [+0.033, +0.045] & $+0.029$ [+0.025, +0.033] & $+0.006$ [-0.002, +0.022] & $+0.022$ [+0.021, +0.025] & $+0.015$ [-0.001, +0.031] \\
\CaRTp $-$ HBV & $+0.031$ [+0.028, +0.033] & $+0.078$ [+0.071, +0.083] & $+0.026$ [+0.023, +0.031] & $+0.021$ [+0.007, +0.029] & $+0.032$ [+0.029, +0.034] & $+0.022$ [+0.014, +0.031] \\
\CaRT $-$ HBV & $+0.015$ [+0.013, +0.017] & $+0.052$ [+0.046, +0.059] & $+0.011$ [+0.007, +0.013] & $+0.013$ [-0.005, +0.027] & $+0.015$ [+0.013, +0.017] & $+0.014$ [-0.000, +0.021] \\
\CaRT $-$ dHBV & $+0.010$ [+0.007, +0.012] & $+0.018$ [+0.012, +0.022] & $+0.014$ [+0.009, +0.018] & $+0.004$ [-0.007, +0.013] & $+0.007$ [+0.005, +0.009] & $+0.000$ [-0.009, +0.008] \\
dHBV $-$ LSTM & $-0.028$ [-0.031, -0.025] & $-0.054$ [-0.060, -0.049] & $-0.042$ [-0.047, -0.038] & $-0.011$ [-0.025, +0.009] & $-0.033$ [-0.036, -0.030] & $-0.014$ [-0.031, -0.001] \\
LSTM $-$ \CaRT & $+0.019$ [+0.017, +0.021] & $+0.036$ [+0.032, +0.039] & $+0.026$ [+0.023, +0.033] & $+0.016$ [-0.004, +0.034] & $+0.026$ [+0.025, +0.029] & $+0.019$ [-0.001, +0.040] \\
EA-LSTM $-$ LSTM & $-0.005$ [-0.006, -0.003] & $-0.006$ [-0.008, -0.005] & $-0.015$ [-0.017, -0.012] & $-0.010$ [-0.028, +0.003] & $-0.007$ [-0.008, -0.005] & $-0.009$ [-0.020, +0.003] \\
\bottomrule
\end{tabular}
}\\[4pt]
\resizebox{\textwidth}{!}{%
\begin{tabular}{lcccc|cc}
\toprule
 & \multicolumn{4}{c|}{single model} & \multicolumn{2}{c}{3-seed ensemble} \\
Pair & $\Delta$NSE & $\Delta$NSE$_{\log}$ & $\Delta$KGE & $\Delta$NSE ungauged & $\Delta$NSE & $\Delta$NSE ungauged \\
\midrule
\CaRTp $-$ LSTM & $+0.008$ [+0.004, +0.011] & $-0.017$ [-0.025, -0.013] & $+0.007$ [-0.002, +0.014] & $-0.003$ [-0.034, +0.020] & $-0.012$ [-0.017, -0.009] & $-0.016$ [-0.073, -0.005] \\
\CaRTp $-$ EA-LSTM & $+0.001$ [-0.004, +0.006] & $-0.001$ [-0.006, +0.007] & $+0.011$ [+0.002, +0.016] & $-0.031$ [-0.055, -0.010] & $-0.008$ [-0.013, -0.003] & $-0.033$ [-0.060, -0.017] \\
\CaRTp $-$ \CaRT & $+0.031$ [+0.026, +0.036] & $+0.084$ [+0.069, +0.093] & $+0.038$ [+0.032, +0.045] & $+0.034$ [+0.015, +0.049] & $+0.034$ [+0.030, +0.039] & $+0.029$ [+0.016, +0.047] \\
\CaRTp $-$ dHBV & $+0.039$ [+0.034, +0.044] & $+0.112$ [+0.099, +0.130] & $+0.070$ [+0.057, +0.080] & $+0.010$ [-0.013, +0.023] & $+0.037$ [+0.033, +0.041] & $+0.017$ [+0.004, +0.032] \\
\CaRTp $-$ HBV & $+0.056$ [+0.050, +0.063] & $+0.300$ [+0.271, +0.320] & $+0.085$ [+0.074, +0.101] & $+0.053$ [+0.033, +0.076] & $+0.060$ [+0.053, +0.067] & $+0.061$ [+0.033, +0.073] \\
\CaRT $-$ HBV & $+0.023$ [+0.017, +0.029] & $+0.187$ [+0.170, +0.206] & $+0.040$ [+0.026, +0.053] & $+0.029$ [+0.016, +0.054] & $+0.021$ [+0.016, +0.029] & $+0.026$ [+0.014, +0.051] \\
\CaRT $-$ dHBV & $+0.008$ [+0.002, +0.015] & $+0.023$ [+0.009, +0.033] & $+0.025$ [+0.018, +0.033] & $-0.028$ [-0.056, +0.009] & $+0.004$ [-0.002, +0.010] & $-0.025$ [-0.060, +0.013] \\
dHBV $-$ LSTM & $-0.027$ [-0.034, -0.023] & $-0.140$ [-0.160, -0.119] & $-0.062$ [-0.072, -0.052] & $-0.018$ [-0.053, +0.008] & $-0.047$ [-0.054, -0.041] & $-0.046$ [-0.074, -0.019] \\
LSTM $-$ \CaRT & $+0.022$ [+0.017, +0.027] & $+0.093$ [+0.077, +0.106] & $+0.037$ [+0.027, +0.042] & $+0.044$ [+0.016, +0.079] & $+0.047$ [+0.041, +0.052] & $+0.062$ [+0.034, +0.099] \\
EA-LSTM $-$ LSTM & $+0.007$ [+0.004, +0.011] & $-0.013$ [-0.019, -0.008] & $-0.002$ [-0.008, +0.005] & $+0.016$ [-0.003, +0.036] & $-0.004$ [-0.008, -0.000] & $+0.012$ [-0.011, +0.027] \\
\bottomrule
\end{tabular}
}
\end{table}

\begin{figure}[tbp]\centering
\includegraphics[width=\textwidth]{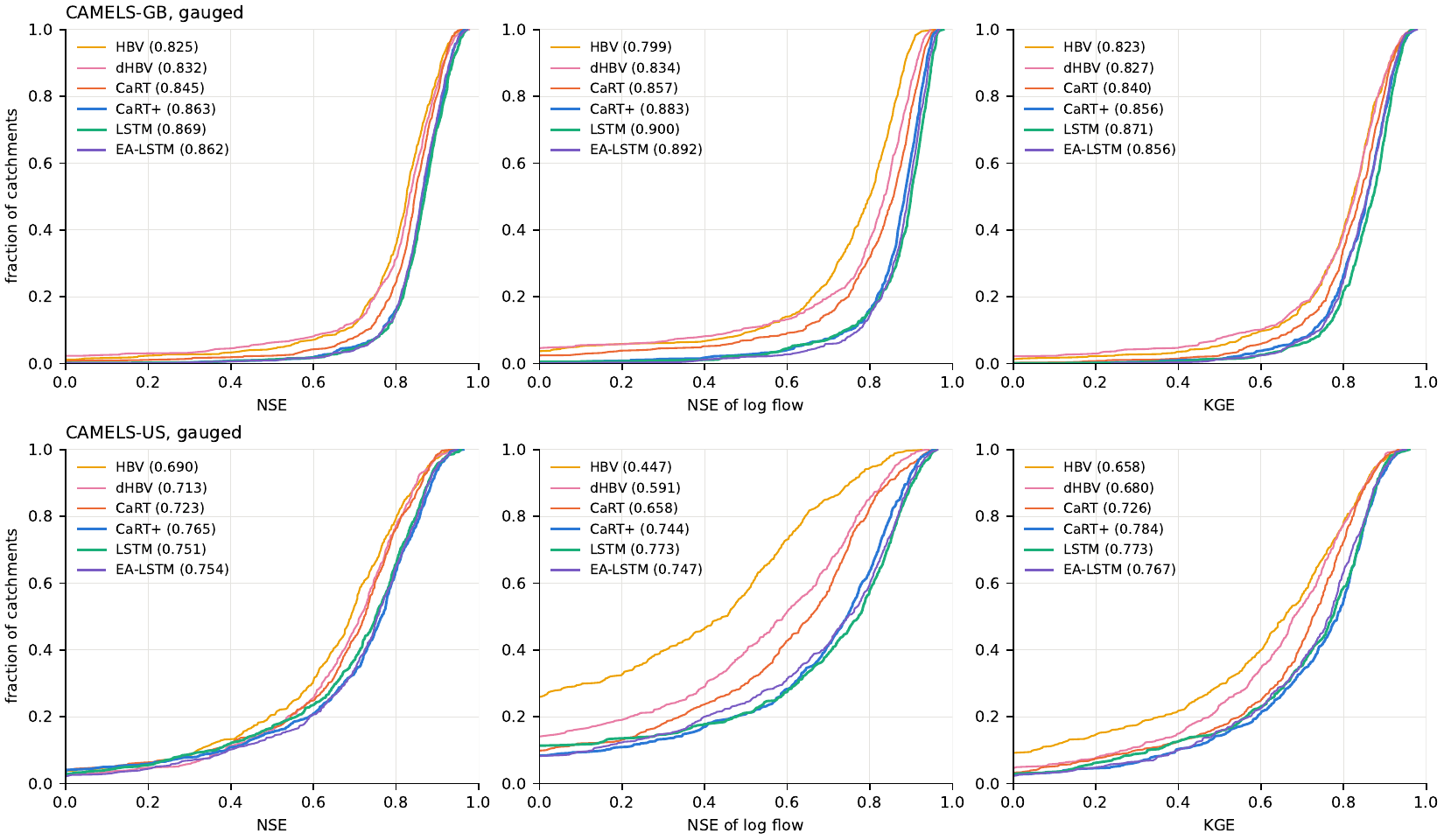}
\caption{Scores of the models in the held-out period for the gauged catchments, averaged over three seeds.}\label{fig:cdf}
\end{figure}

\paragraph{Overall.} We first compare single models, averaging each catchment metric over the three seeds, to isolate the effect of structure, with the three-seed ensembles reported alongside (Tables~\ref{tab:u_gb}--\ref{tab:u_paired}, Figure~\ref{fig:cdf}). As a single model, \CaRTp{} matches the two LSTMs on both continents. It is behind the stateful LSTM by 0.004 [0.003, 0.006] in paired NSE on CAMELS-GB and ahead of it by 0.008 [0.004, 0.011] on CAMELS-US, and within 0.001 of the EA-LSTM on both. It improves on every process-based benchmark (\CaRT, dHBV and HBV) by 0.015--0.031 on CAMELS-GB and 0.031--0.056 on CAMELS-US, while the stateful LSTM keeps a lead in the NSE of log flow (0.013 and 0.017). On CAMELS-GB \CaRTp{} has no gauged catchment with a negative NSE and reproduces the shape of the low-flow segment better than the LSTMs (FLV $-16\,\%$ against $-28$ and $-19\,\%$). On CAMELS-US its weaknesses are the lower tail of the NSE distribution and the dry season, with NSE below zero in 21 gauged basins (17 and 13 for the LSTMs). \CaRTp{} overestimates low-flow volume by 30\,\% (17 and 45\,\% for the stateful LSTM and the EA-LSTM).

\paragraph{Seed averaging.} Averaging three seeds changes the order, with paired NSE gains of 0.009 (CAMELS-GB) and 0.027 (CAMELS-US) in the median catchment for the stateful LSTM, against 0.007 and 0.016 for the EA-LSTM and 0.002 and 0.006 for \CaRTp. The ensembles of the stateful LSTM and the EA-LSTM lead the \CaRTp{} ensemble by 0.011 and 0.003 on CAMELS-GB and by 0.012 and 0.008 on CAMELS-US (Table~\ref{tab:u_paired}). In each catchment the gain of an ensemble in NSE equals the time-mean spread of its members divided by the variance of the observed flow \citep{krogh1995}, and it vanishes when the members agree. The spread of the \CaRTp{} seeds is about a quarter of that of the stateful LSTM. A smaller share of this spread falls on the days above the observed 95\,\% flow quantile (32 and 53\,\%, against 57 and 69\,\% for the stateful LSTM; Table~\ref{tab:u_spread}).

\begin{figure}[tbp]\centering
\includegraphics[width=\textwidth]{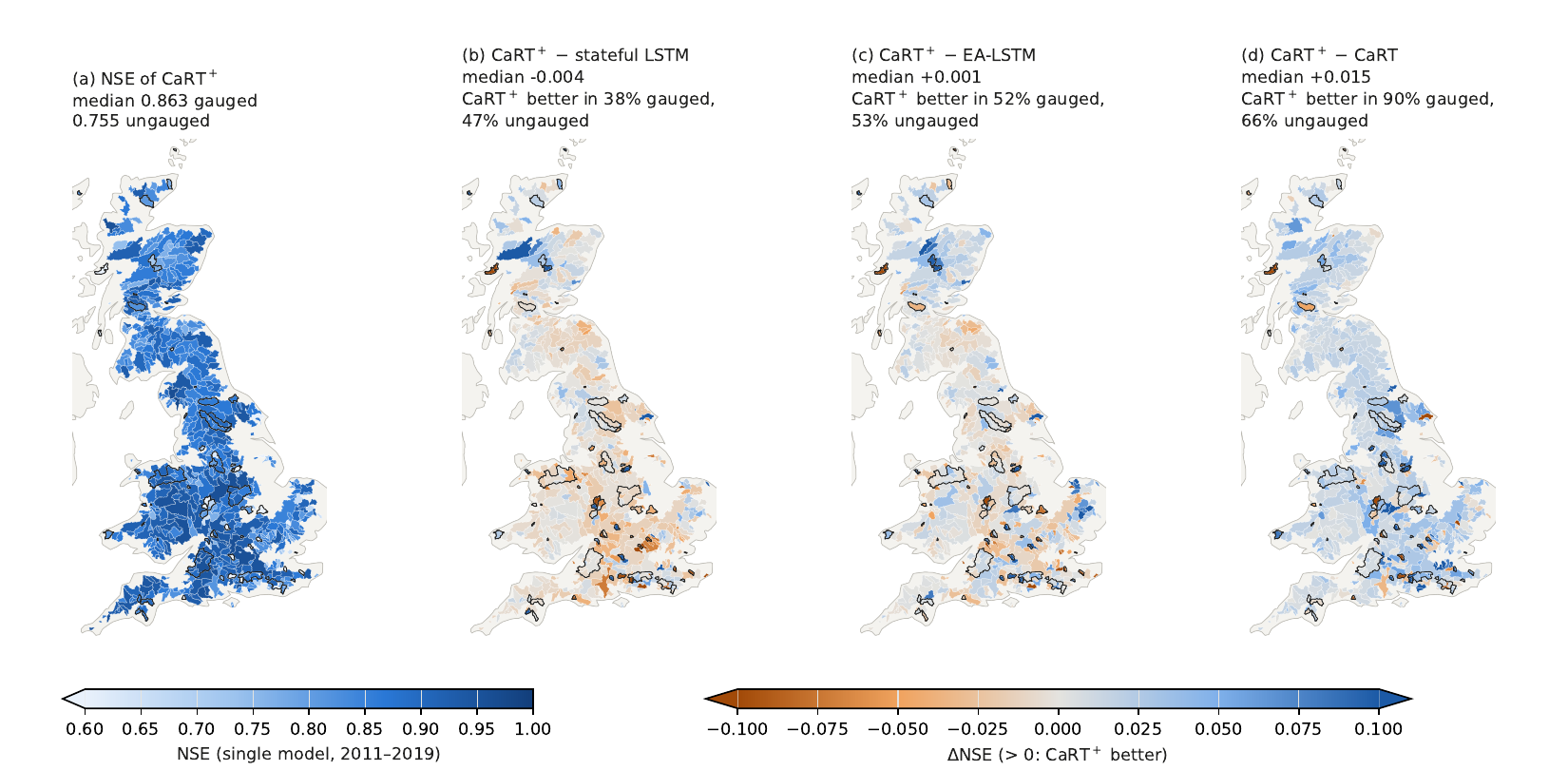}
\caption{NSE of \CaRTp{} on CAMELS-GB in the held-out period and its difference from the other models, with ungauged catchments outlined in black and white areas without test data.}\label{fig:map_gb}
\end{figure}

\begin{figure}[tbp]\centering
\includegraphics[width=\textwidth]{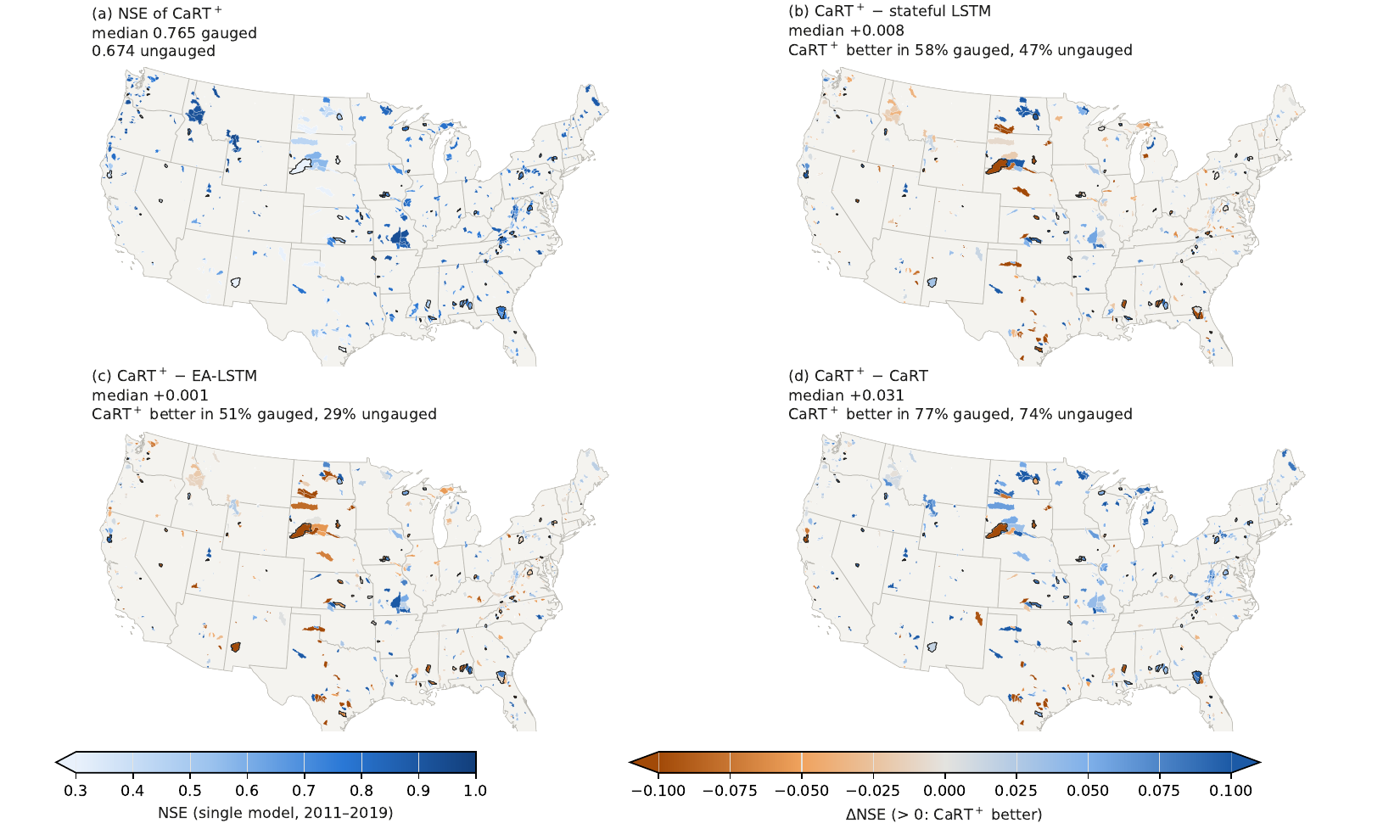}
\caption{NSE of \CaRTp{} on CAMELS-US in the held-out period and its difference from the other models, with ungauged catchments outlined in black and white areas without test data.}\label{fig:map_us}
\end{figure}

\paragraph{By class and in space.} On CAMELS-GB the class medians of the single \CaRTp{} and the stateful LSTM differ by at most 0.003 in three of the four classes (Table~\ref{tab:u_groups}). In the groundwater-fed catchments the median of the LSTM is higher by 0.008, although the refinements raised the NSE of \CaRTp{} most there. On CAMELS-US the class medians of \CaRTp{} exceed those of the stateful LSTM in every class, by 0.008 in the humid to 0.032 in the arid basins, and fall below those of the EA-LSTM in the snow-dominated and arid basins. On the maps (Figures~\ref{fig:map_gb} and \ref{fig:map_us}) the differences are small almost everywhere on CAMELS-GB. On CAMELS-US they are clustered, with the LSTMs ahead in the semi-arid northern Great Plains and Texas and \CaRTp{} in the upper Midwest and the east. The held-out catchments are discussed in Section~\ref{sec:ung}.

\begin{table}[tbp]\centering\small
\caption{Median single-model NSE / NSE$_{\log}$ (per-catchment mean over three seeds) over classes of gauged catchments, held-out period, with the NSE of the three-seed ensemble in parentheses. Top: CAMELS-GB; bottom: CAMELS-US. The classes follow the main hydrological gradient of each sample: the baseflow index on CAMELS-GB, where the permeable chalk and limestone catchments differ most from the rest, and snow fraction and aridity on CAMELS-US.}\label{tab:u_groups}
\resizebox{\textwidth}{!}{%
\begin{tabular}{lcccc}
\toprule
Model & BFI $<0.5$ & $0.5\le$ BFI $<0.65$ & BFI $\ge0.65$ & $f_{\rm snow}\ge0.04$ \\
 & $n=172$ & $n=273$ & $n=157$ & $n=82$ \\
\midrule
HBV & 0.820 / 0.815 (0.821) & 0.839 / 0.814 (0.839) & 0.802 / 0.751 (0.805) & 0.805 / 0.774 (0.806) \\
dHBV & 0.821 / 0.841 (0.825) & 0.849 / 0.849 (0.853) & 0.830 / 0.784 (0.837) & 0.818 / 0.818 (0.821) \\
\CaRT & 0.830 / 0.855 (0.832) & 0.859 / 0.866 (0.860) & 0.856 / 0.838 (0.859) & 0.831 / 0.848 (0.832) \\
\CaRTp & 0.847 / 0.877 (0.849) & 0.874 / 0.890 (0.876) & 0.879 / 0.882 (0.882) & 0.847 / 0.863 (0.849) \\
LSTM & 0.845 / 0.888 (0.856) & 0.877 / 0.913 (0.887) & 0.887 / 0.891 (0.900) & 0.845 / 0.877 (0.862) \\
EA-LSTM & 0.845 / 0.882 (0.851) & 0.872 / 0.905 (0.878) & 0.875 / 0.882 (0.885) & 0.838 / 0.869 (0.846) \\
\bottomrule
\end{tabular}
}\\[4pt]
\resizebox{\textwidth}{!}{%
\begin{tabular}{lcccccc}
\toprule
Model & $f_{\rm snow}<0.1$ & $0.1\le f_{\rm snow}<0.3$ & $f_{\rm snow}\ge0.3$ & PET/P $\le0.7$ & $0.7<$ PET/P $\le1.1$ & PET/P $>1.1$ \\
 & $n=299$ & $n=188$ & $n=106$ & $n=154$ & $n=264$ & $n=175$ \\
\midrule
HBV & 0.661 / 0.315 (0.666) & 0.677 / 0.484 (0.680) & 0.804 / 0.528 (0.813) & 0.773 / 0.628 (0.774) & 0.677 / 0.461 (0.681) & 0.511 / -0.388 (0.516) \\
dHBV & 0.683 / 0.474 (0.691) & 0.718 / 0.687 (0.728) & 0.795 / 0.658 (0.811) & 0.786 / 0.753 (0.791) & 0.709 / 0.573 (0.717) & 0.561 / 0.363 (0.596) \\
\CaRT & 0.723 / 0.578 (0.725) & 0.695 / 0.691 (0.696) & 0.773 / 0.743 (0.778) & 0.787 / 0.786 (0.792) & 0.727 / 0.650 (0.728) & 0.546 / 0.354 (0.553) \\
\CaRTp & 0.738 / 0.685 (0.746) & 0.758 / 0.791 (0.770) & 0.836 / 0.803 (0.842) & 0.817 / 0.840 (0.821) & 0.771 / 0.735 (0.777) & 0.621 / 0.418 (0.643) \\
LSTM & 0.718 / 0.705 (0.749) & 0.750 / 0.802 (0.778) & 0.825 / 0.846 (0.855) & 0.809 / 0.865 (0.835) & 0.754 / 0.759 (0.783) & 0.589 / 0.362 (0.642) \\
EA-LSTM & 0.729 / 0.654 (0.749) & 0.752 / 0.802 (0.767) & 0.848 / 0.856 (0.862) & 0.813 / 0.864 (0.830) & 0.753 / 0.720 (0.772) & 0.630 / 0.392 (0.660) \\
\bottomrule
\end{tabular}
}
\end{table}

\begin{table}[tbp]\centering\small
\caption{Gain from averaging three seeds on the held-out period: median over gauged catchments of the ensemble NSE minus the mean NSE of the members, which equals the spread of the members relative to the observed variance. Also shown is the share of that spread on the days above the observed 95\,\% and below the observed 30\,\% flow quantile.}\label{tab:u_spread}
\begin{tabular}{llccc}
\toprule
Dataset & model & ensemble gain & spread on days $\ge Q_{95}$ & spread on days $\le Q_{30}$ \\
\midrule
CAMELS-GB & \CaRTp & $+0.0022$ & 32\,\% & 6\,\% \\
 & stateful LSTM & $+0.0093$ & 57\,\% & 2\,\% \\
 & EA-LSTM & $+0.0066$ & 48\,\% & 3\,\% \\
\midrule
CAMELS-US & \CaRTp & $+0.0060$ & 53\,\% & 3\,\% \\
 & stateful LSTM & $+0.0269$ & 69\,\% & 1\,\% \\
 & EA-LSTM & $+0.0163$ & 62\,\% & 1\,\% \\
\bottomrule
\end{tabular}
 \end{table}

\subsection{What each refinement contributes}\label{sec:contrib}

\begin{figure}[tbp]\centering
\includegraphics[width=\textwidth]{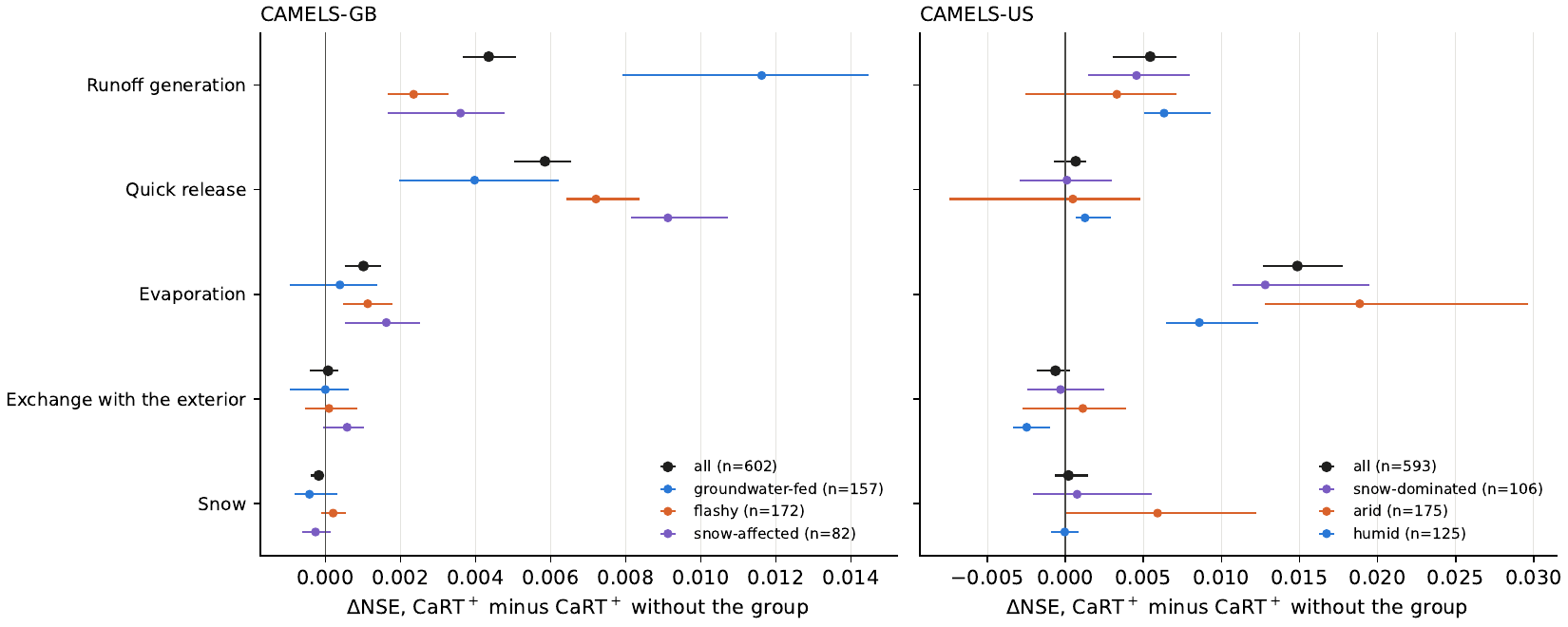}
\caption{What each group of refinements adds to \CaRTp{} in the held-out period, with medians and 95\,\% bootstrap intervals.}\label{fig:lofo}
\end{figure}

The refinements of \CaRTp{} form five groups by the process they act on:
\begin{itemize}\setlength\itemsep{0pt}
\item Runoff generation: surface layer, Pareto partition, infiltration excess, depression storage and near-stream zone.
\item Quick release: connected release with its exponent, viscosity and frost, connected bypass and groundwater-controlled connectivity.
\item Evaporation: vegetation demand and groundwater uptake.
\item Exchange with the exterior: exterior inflow and channel transmission loss.
\item Snow: melt energy term.
\end{itemize}
These groups also reflect dependencies among their constituent refinements. To test a group's effect on the model performance, the model was retrained with one group removed and the others retained. The NSE of each catchment is averaged over the seeds before the paired differences are calculated (Table~\ref{tab:lofo} and, by class, Figure~\ref{fig:lofo}). The classes are groundwater-fed (BFI $\ge0.65$), flashy (BFI $<0.5$) and snow-affected ($f_{\rm snow}\ge0.04$) catchments on CAMELS-GB, and snow-dominated ($f_{\rm snow}\ge0.3$), arid (PET/P $>1.1$) and humid (PET/P $\le0.7$, $f_{\rm snow}<0.3$) basins on CAMELS-US.

On CAMELS-GB \CaRTp{} gains 0.015 [0.013, 0.016] over \CaRT, with quick release contributing most at 0.006 [0.005, 0.007] overall, 0.009 in the snow-affected and 0.007 in the flashy catchments. Runoff generation follows at 0.004 [0.004, 0.005], with 0.012 in the groundwater-fed catchments. Evaporation adds 0.0010 [0.0005, 0.0015], while exchange with the exterior and the melt energy term leave the median unchanged. The exchange group improves the prediction at Costa Beck (27038), where NSE is 0.33 with this group and $-4.2$ without it.

On CAMELS-US, where the gain is 0.031 [0.026, 0.036], evaporation contributes most at 0.015 [0.013, 0.018], with 0.019 in the arid and 0.013 in the snow-dominated basins. Runoff generation follows at 0.005 [0.003, 0.007]. The contributions of quick release once the other groups are present, 0.0007 [$-0.0007$, 0.0013], and of the melt energy term, 0.0002 [$-0.0007$, 0.0014], include zero. Exchange with the exterior lowers the held-out NSE of the humid basins by 0.0025 [0.0010, 0.0033], but the group was retained on the basis of the development-split comparisons.

Three of the five groups of refinements (runoff generation, quick release and evaporation) raise the median NSE on at least one of the continents. The exchange and snow groups leave the medians unchanged and act in particular catchments. The melt energy term adds 0.006 [0.000, 0.012] in the 175 arid basins of CAMELS-US. The five median contributions sum to 0.011 on CAMELS-GB and 0.021 on CAMELS-US, less than the gains over \CaRT. Each contribution is measured with the other four groups present, so where two groups improve the same catchments, each is credited only with the part that the others do not replace.

\begin{table}[tbp]\centering\small
\caption{Contribution of each group of refinements to \CaRTp{} on the held-out period, their sum and the gain of \CaRTp{} over \CaRT. Group contributions are calculated as \CaRTp{} minus \CaRTp{} without the group: paired $\Delta$NSE of the per-catchment mean over three seeds, reported as the median and 95\,\% bootstrap interval.}\label{tab:lofo}
\resizebox{\linewidth}{!}{%
\begin{tabular}{p{0.5\textwidth}cc}
\toprule
Group removed from \CaRTp & CAMELS-GB & CAMELS-US \\
\midrule
Runoff generation (surface layer, Pareto partition, infiltration excess, depression storage, near-stream zone) & $+0.0043$ [+0.0037, +0.0051] & $+0.0054$ [+0.0030, +0.0071] \\
Quick release (connected release with exponent, viscosity and frost, connected bypass, groundwater-controlled connectivity) & $+0.0058$ [+0.0050, +0.0065] & $+0.0007$ [-0.0007, +0.0013] \\
Evaporation (vegetation demand, groundwater uptake) & $+0.0010$ [+0.0005, +0.0015] & $+0.0149$ [+0.0126, +0.0178] \\
Exchange with the exterior (exterior inflow, channel transmission loss) & $+0.0001$ [-0.0004, +0.0003] & $-0.0006$ [-0.0019, +0.0003] \\
Snow (melt energy term) & $-0.0002$ [-0.0004, -0.0000] & $+0.0002$ [-0.0007, +0.0014] \\
\midrule
Sum of the five & $+0.0111$ & $+0.0205$ \\
\CaRTp{} minus \CaRT{} & $+0.0148$ [+0.0133, +0.0161] & $+0.0308$ [+0.0256, +0.0357] \\
\bottomrule
\end{tabular}
}
\end{table}

\subsection{Hydrographs}\label{sec:hydro}

\begin{figure}[tbp]\centering
\includegraphics[width=\textwidth]{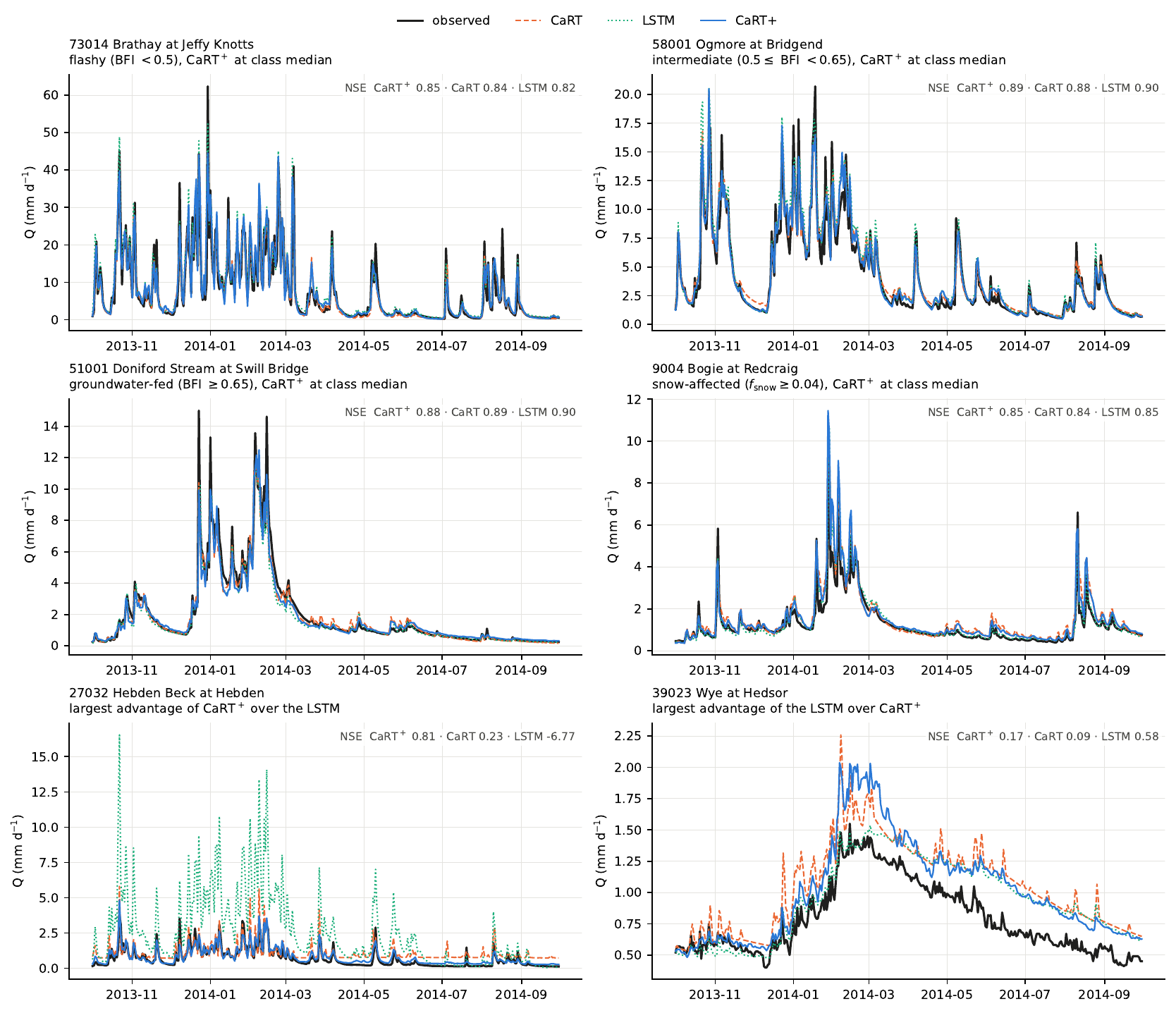}
\caption{Observed and simulated daily flow over one year in six catchments of CAMELS-GB, with the NSE of the whole held-out period in each title.}\label{fig:hydro_gb}
\end{figure}

\begin{figure}[tbp]\centering
\includegraphics[width=\textwidth]{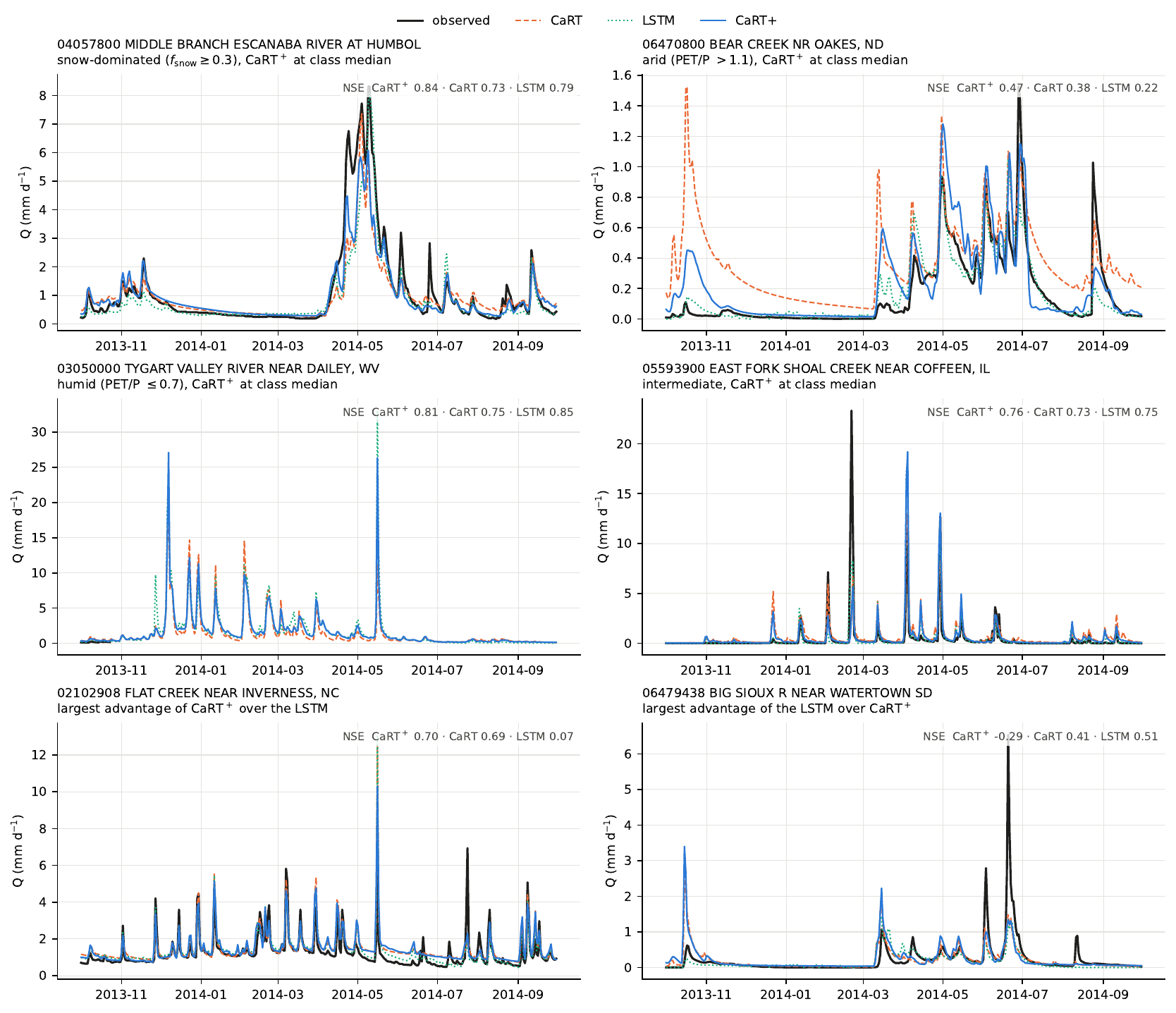}
\caption{Observed and simulated daily flow over one year in six catchments of CAMELS-US, with the NSE of the whole held-out period in each title.}\label{fig:hydro_us}
\end{figure}

Figures~\ref{fig:hydro_gb} and \ref{fig:hydro_us} show one water year in six gauged catchments per continent for seed~0 of each model, the seed that was also used to select the catchments. Four are the catchments at the median NSE of \CaRTp{} in each of four classes (those of Figure~\ref{fig:lofo} and an intermediate class). The other two are the catchments with the largest advantage of \CaRTp{} over the stateful LSTM and of the stateful LSTM over \CaRTp. In the class-median catchments the models follow the observed flow closely and differ mainly in the peaks, and only in the arid Bear Creek does no model exceed 0.5. On CAMELS-GB the largest advantage of \CaRTp{} is at Hebden Beck, where the LSTM produces spurious winter peaks (NSE $-6.8$). The largest advantage of the LSTM is at the Wye at Hedsor, a chalk stream where both conceptual models overestimate the late-winter rise. Its flow may be controlled by an aquifer head that is not among the inputs. On CAMELS-US the two extreme basins differ in kind. At Flat Creek all three LSTM seeds fail over the held-out period (NSE 0.07, 0.05 and 0.30), while the three \CaRTp{} seeds agree (0.66--0.70). At the Big Sioux River the result depends on the seed. The displayed \CaRTp{} seed and one other overestimate the autumn and spring events and miss the June peaks, a pattern also seen in one LSTM seed.

\subsection{Held-out catchments}\label{sec:ung}

\begin{figure}[tbp]\centering
\includegraphics[width=\textwidth]{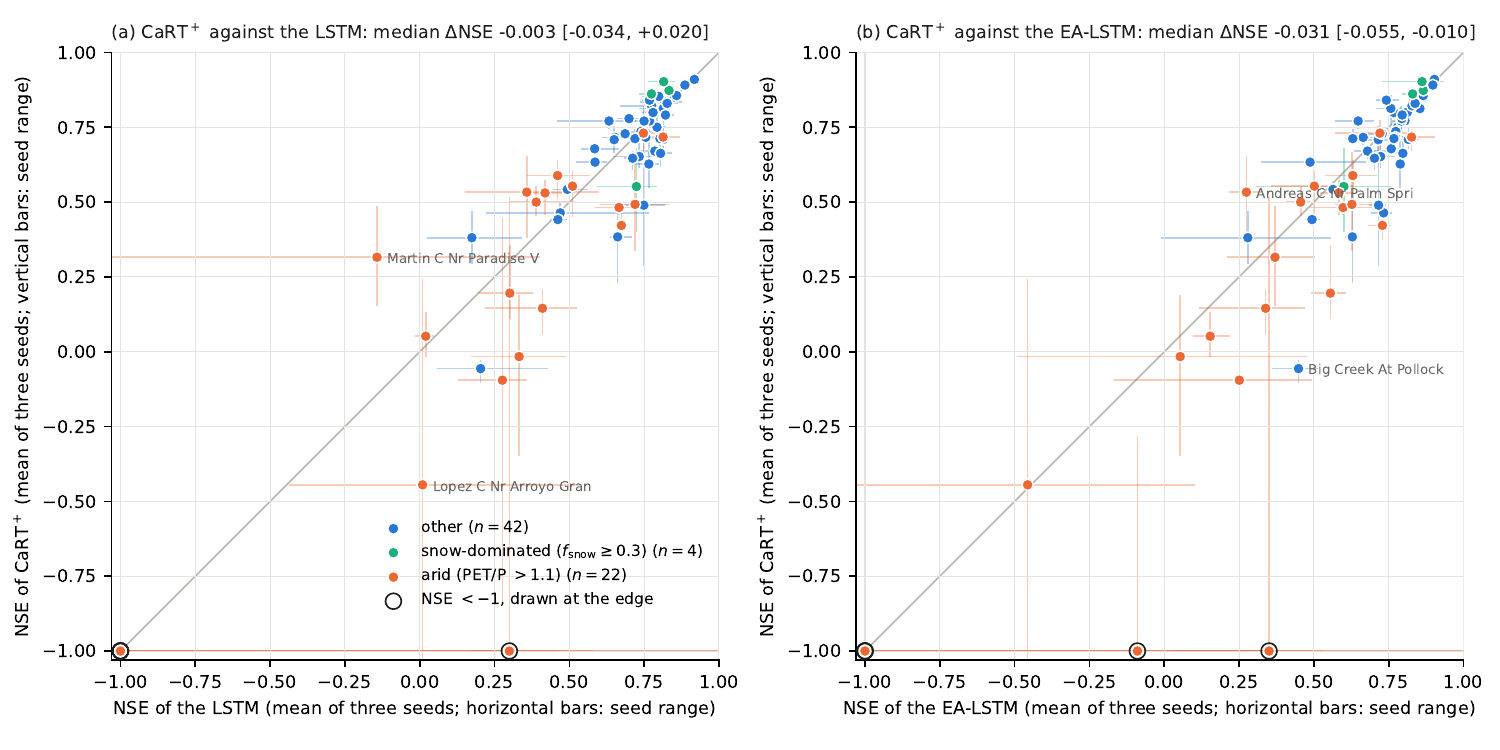}
\caption{NSE of \CaRTp{} against the two LSTMs in the 68 held-out basins of CAMELS-US, with five basins below $-1$ placed at the edge.}\label{fig:ung_us}
\end{figure}

The catchments held out from training test the map from attributes to parameters. On the 64 held-out catchments of CAMELS-GB with test data, the median NSE is 0.755 for the single \CaRTp{}, against 0.723 and 0.714 for the stateful LSTM and the EA-LSTM. However, the paired differences, $-0.002$ [$-0.010$, $+0.010$] and $+0.004$ [$-0.011$, $+0.020$], include zero, as do those of the ensembles (Table~\ref{tab:u_paired}). On CAMELS-US the paired difference with the stateful LSTM also includes zero (0.674 against 0.719; $-0.003$ [$-0.034$, $+0.020$]), and \CaRTp{} is behind the EA-LSTM (0.722; $-0.031$ [$-0.055$, $-0.010$]). Both LSTM ensembles are ahead of the \CaRTp{} ensemble (0.754 and 0.749 against 0.689). Figure~\ref{fig:ung_us} compares the 68 held-out basins of CAMELS-US one by one and Table~\ref{tab:ung_us} summarises them by class. The deficit is largest in the 22 arid basins (PET/P $>1.1$), where \CaRTp{} is behind the EA-LSTM by 0.112 [0.053, 0.326] and consistent with no difference from the stateful LSTM ($-0.145$ [$-0.319$, $+0.043$]). In the other 46 basins it is within 0.002 of the stateful LSTM ($+0.002$ [$-0.007$, $+0.025$]) and 0.013 [0.006, 0.031] behind the EA-LSTM. In the arid basins the three seeds of each model spread widely, often over more than 0.5 in NSE, because flow is zero on most days and a few peaks dominate the score.

Five held-out basins have an NSE below $-1$ for at least one model. They are Blackwater Creek, FL ($-35$ for \CaRTp, $-17$ and $-14$ for the two LSTMs), Castle Creek, SD ($-16$, $-1.7$, $-9.0$), Vernon Creek, UT ($-14$, $-22$, $-3.3$), Beaver Creek, OR ($-1.8$, $-1.2$, $-0.1$) and Firesteel Creek, SD ($-1.7$, $+0.3$, $+0.4$). The nine held-out basins in which \CaRTp{} has a negative NSE share a low runoff ratio in the training period (median 0.10, against 0.37 for the other held-out basins). Their runoff ratio is also lower than that of the gauged basins nearest to them in attribute space (median 0.18), and they lie farther from any gauged basin in attribute space (2.0 against 1.3 standard deviations). The attributes supplied to the network do not reveal that most of their rain leaves without reaching the gauge, and a model that conserves water must then send the surplus to the outlet. Decomposing NSE into correlation, amplitude ratio and bias shows two forms of the resulting failure. In basins with near-zero flow on most days, simulated peaks are several times too large (amplitude ratios of 6.2 at Blackwater Creek and 4.3 at Vernon Creek, against 4.4 and 4.9 for the LSTM, which fails there too). In the karst basin of Castle Creek, in the Black Hills, the simulated volume is 2.7 times the observed one (1.5 times for the LSTM), because the loss to the aquifer is not predicted. This tendency is also visible among the gauged basins. In the 68 gauged basins with a runoff ratio below 0.12 in the training period, the runoff ratio of their nearest neighbours in attribute space is 0.09 against 0.06 observed, and \CaRTp{} overestimates their volume by 13\,\% against 3\,\% elsewhere. %

There is a second cause in all held-out basins. For each basin the attribute network returns one set of parameters, and the model simulates the single hydrograph of that set. Both models are trained on the squared error, whose minimum for given attributes and forcing lies at the conditional mean of the flow, and the mean of several hydrographs is smoother than any of them. The LSTM outputs the flow itself and can return this mean for basins that cannot be distinguished by their attributes. However, \CaRTp{} returns a hydrograph of its own family, and since its response is nonlinear the mean of several of its hydrographs lies outside that family. The average is reached only by averaging the flows of several parameter sets. We therefore perturb the standardised attributes of each basin with Gaussian noise of standard deviation $\sigma$, run \CaRTp{} once for each of 32 draws with the trained network of each seed, and average the flows (Table~\ref{tab:hedge}). Every member conserves water, and the average is the outflow of 32 equal units in parallel, which is the aggregate of Section~\ref{sec:laws3} with the uncertainty of the map as its heterogeneity. Training perturbs the attributes in the same way with $\sigma=0.1$. If the deficit comes from this single parameter set, four things follow. The held-out score rises with $\sigma$ well beyond 0.1, and the gain is largest where the deficit is. At larger $\sigma$ the gauged score falls, because there the flow has pinned the parameters. The failing basins, whose failure comes from a missing loss, stay where they are. On CAMELS-US all four hold, the third from $\sigma=0.5$ and the fourth in most failing basins. The held-out median rises from 0.674 to 0.696 at $\sigma=0.3$ and 0.704 at $\sigma=0.5$ and falls at larger $\sigma$. At $\sigma=0.3$, 79\,\% of the basins improve (paired $+0.019$ [$+0.012$, $+0.027$]), and the paired gain in the median arid basin is 0.08 at $\sigma=0.3$ and 0.15 at $\sigma=0.5$, against 0.01 in the other 46. It comes through the correlation and the bias, which move from 0.76 to 0.79 and from 0.21 to 0.18 standard deviations in the median arid basin at $\sigma=0.3$, while the amplitude ratio stays near 0.9. The average removes peaks that the members place on different days. At $\sigma=0.3$ the paired differences with both LSTMs include zero ($+0.012$ [$-0.021$, $+0.030$] and $-0.007$ [$-0.020$, $+0.002$]). On the gauged basins the same average stays above the baseline up to $\sigma=0.3$, where it gains 0.006, but loses 0.017 at $\sigma=0.5$ and 0.046 at $\sigma=0.7$. Six of the nine failing basins stay below zero at every $\sigma$. On CAMELS-GB, where the single \CaRTp{} already matches the LSTMs, the held-out median stays within 0.01 of 0.755 up to $\sigma=0.7$. In the 22 held-out groundwater-fed catchments, in which all three models score 0.43, the median rises to 0.59 at $\sigma=0.5$, while in the other 42 it falls by 0.02. The gauged median falls from 0.863 to 0.830 at $\sigma=0.5$. The deficit on held-out basins is therefore a property of the map from attributes to parameters, which is too certain where the attributes constrain the response weakly. For prediction, $\sigma$ would be set on pseudo-ungauged basins within the training set.

\begin{table}[tbp]\centering\small
\caption{Held-out basins of CAMELS-US by class (test period, single models: per-basin NSE averaged over three seeds): median NSE and paired differences with 95\,\% bootstrap intervals.}\label{tab:ung_us}
\resizebox{\textwidth}{!}{%
\begin{tabular}{lcccccc}
\toprule
 & & \multicolumn{3}{c}{median NSE} & \multicolumn{2}{c}{paired $\Delta$NSE, \CaRTp{} minus} \\
Class & $n$ & \CaRTp & LSTM & EA-LSTM & LSTM & EA-LSTM \\
\midrule
all held-out basins & 68 & 0.674 & 0.719 & 0.722 & $-0.003$ [-0.034, +0.020] & $-0.031$ [-0.055, -0.010] \\
non-arid & 46 & 0.743 & 0.768 & 0.773 & $+0.002$ [-0.007, +0.025] & $-0.013$ [-0.031, -0.006] \\
arid (PET/P $>1.1$) & 22 & 0.256 & 0.345 & 0.360 & $-0.145$ [-0.319, +0.043] & $-0.112$ [-0.326, -0.053] \\
snow-dominated, not arid & 4 & 0.867 & 0.795 & 0.846 & $+0.064$ [-0.172, +0.087] & $+0.020$ [-0.048, +0.041] \\
\bottomrule
\end{tabular}
}
\end{table}

\begin{table}[tbp]\centering\small
\caption{Parameter-space ensemble of \CaRTp{} on the held-out period: median NSE over the held-out and the gauged catchments when the standardised attributes of every catchment are perturbed with Gaussian noise of standard deviation $\sigma$ and the flows of 32 draws are averaged (per-catchment NSE averaged over three seeds; $\sigma=0$ is the single model of the other tables).}\label{tab:hedge}
\begin{tabular}{llcccc}
\toprule
 & & \multicolumn{3}{c}{held-out catchments} & gauged \\
Dataset & $\sigma$ & all & arid (US) / groundwater-fed (GB) & other & all \\
\midrule
CAMELS-GB & 0 (plain) & 0.755 & 0.428 & 0.810 & 0.863 \\
 & 0.1 & 0.764 & 0.435 & 0.811 & 0.862 \\
 & 0.2 & 0.764 & 0.449 & 0.810 & 0.859 \\
 & 0.3 & 0.749 & 0.500 & 0.799 & 0.853 \\
 & 0.5 & 0.765 & 0.593 & 0.790 & 0.830 \\
 & 0.7 & 0.759 & 0.511 & 0.786 & 0.811 \\
 & 1.0 & 0.742 & 0.409 & 0.774 & 0.789 \\
\midrule
CAMELS-US & 0 (plain) & 0.674 & 0.256 & 0.743 & 0.765 \\
 & 0.1 & 0.673 & 0.263 & 0.749 & 0.768 \\
 & 0.2 & 0.682 & 0.287 & 0.756 & 0.772 \\
 & 0.3 & 0.696 & 0.310 & 0.749 & 0.771 \\
 & 0.5 & 0.704 & 0.343 & 0.747 & 0.748 \\
 & 0.7 & 0.695 & 0.418 & 0.732 & 0.719 \\
 & 1.0 & 0.658 & 0.474 & 0.692 & 0.672 \\
\bottomrule
\end{tabular}
 \end{table}

\subsection{Transfer to CAMELS-AUS}\label{sec:aus}

CAMELS-AUS tests whether the structure holds beyond the two continents on which it was built. No refinement was derived or selected on Australian data, and the catchments were added after the structure was fixed. They are drier than those of the other two datasets, with PET/P above 2 in 92 of the 505 gauged catchments. In 332 of the 505 gauged catchments the flow falls below 0.01\,mm\,d$^{-1}$ on at least 5\,\% of the test days, and for most of them the driest test years are 2018 and 2019. All models are trained with the configuration of CAMELS-US. In the driest catchments the infiltration capacity of \CaRTp{} underflowed in single precision and made the gradients undefined, so the infiltration capacity is bounded below by $10^{-4}$\,mm\,d$^{-1}$.

\begin{figure}[tbp]\centering
\includegraphics[width=\textwidth]{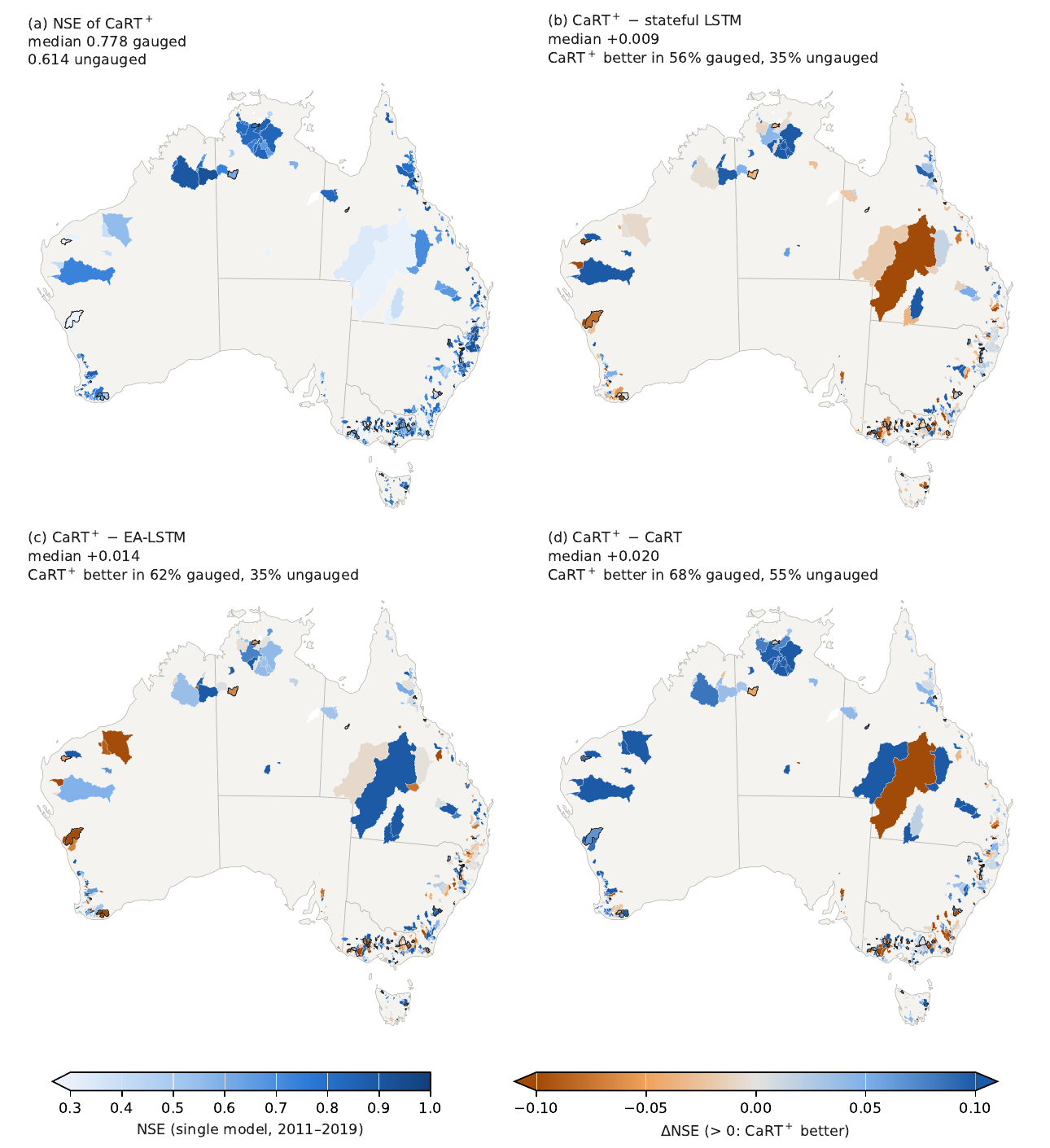}
\caption{NSE of \CaRTp{} on CAMELS-AUS in the held-out period and its difference from the other models, with ungauged catchments outlined in black and white areas without test data.}\label{fig:map_aus}
\end{figure}

On the gauged catchments the structure keeps its accuracy (Table~\ref{tab:aus} and Figure~\ref{fig:map_aus}), and as a single model \CaRTp{} has the highest median NSE there, ahead of the stateful LSTM by 0.009 [0.003, 0.015] in the paired median (Table~\ref{tab:aus}b). However, the LSTM ensemble keeps a lead of 0.010 [0.002, 0.014] over the \CaRTp{} ensemble, as on the other continents. In the paired median the refinements derived for Britain and the United States raise NSE over \CaRT{} by 0.020, against 0.015 and 0.031 on those continents (Table~\ref{tab:u_paired}). They raise NSE of log flow by 0.14 against 0.023 and 0.084, so their largest effect here is at low flow.

The models differ most in dry years. In the test years in which a catchment received less than its lower quartile of annual precipitation (mostly 2018 and 2019), the other conceptual models deliver one and a half to two and a half times the observed flow in the median catchment. \CaRTp{} delivers 17\,\% too much (59\,\% for \CaRT), and its dry-year NSE is higher than that of \CaRT{} by 0.125 [0.094, 0.160]. Only the stateful LSTM stays closer to the observed volume, and it leads \CaRTp{} in these years by 0.036 [0.019, 0.051] in NSE. The same holds for the cessation of flow. In the 332 gauged catchments with no flow on at least 5\,\% of the test days, the other conceptual models underestimate the share of days without flow by 24 to 43 percentage points, \CaRTp{} by 5 and the stateful LSTM by 4. In the 92 arid catchments (PET/P $>2$) \CaRTp{} has the highest KGE of all models over the whole test period, $+0.044$ [$+0.003$, $+0.078$] against the stateful LSTM.

On the held-out catchments \CaRTp{} is behind, as on CAMELS-US but by more, trailing the stateful LSTM by 0.037 [0.009, 0.081] and dHBV by 0.037 [0.001, 0.068]. Averaging the simulated flows over attribute draws as in Section~\ref{sec:ung} closes the deficit against the stateful LSTM. At $\sigma=0.5$ the held-out median rises from 0.614 to 0.708 and the paired difference with the stateful LSTM includes zero ($+0.017$ [$-0.013$, $+0.064$]), while the gauged median falls from 0.778 to 0.760.

\begin{table}[tbp]\centering\small
\caption{CAMELS-AUS, 505 gauged and 56 held-out catchments: medians over catchments of the single-model metrics (per-catchment mean over three seeds), test period 2011--2019 unless stated. Dry years: test years in which the catchment received less than its lower quartile of annual precipitation 1980--2019 (502 catchments, mostly 2018 and 2019). No flow: share of test days with flow below 0.01\,mm\,d$^{-1}$, simulated minus observed in percentage points, over the 332 gauged catchments where it is at least 5\,\% in the observations. Arid: the 92 gauged catchments with PET/P $>2$. $\beta$: ratio of simulated to observed mean flow. The rows with $\sigma$ average the flows of 32 attribute draws of standard deviation $\sigma$ (Section~\ref{sec:ung}). (a) Medians. (b) Paired per-catchment differences, median and 95\,\% bootstrap interval below it.}\label{tab:aus}
(a)\\[2pt]
\resizebox{\textwidth}{!}{%
\begin{tabular}{lcccccccccccc}
\toprule
 & \multicolumn{4}{c}{gauged, 2011--2019} & \multicolumn{3}{c}{gauged, dry years} & no flow & arid & held-out & \multicolumn{2}{c}{3-seed ensemble} \\
Model & NSE & NSE$_{\log}$ & KGE & $\beta$ & NSE & KGE & $\beta$ & (pp) & KGE & NSE & NSE & NSE held-out \\
\midrule
HBV & 0.707 & -0.176 & 0.466 & 1.45 & 0.025 & -0.615 & 2.49 & $-43.5$ & 0.096 & 0.609 & 0.708 & 0.626 \\
dHBV & 0.748 & 0.438 & 0.641 & 1.16 & 0.401 & 0.198 & 1.51 & $-34.7$ & 0.558 & 0.677 & 0.759 & 0.697 \\
\CaRT & 0.750 & 0.505 & 0.645 & 1.16 & 0.307 & 0.094 & 1.59 & $-24.1$ & 0.387 & 0.664 & 0.756 & 0.669 \\
\CaRTp & 0.778 & 0.687 & 0.703 & 1.07 & 0.452 & 0.390 & 1.17 & $-5.0$ & 0.623 & 0.614 & 0.784 & 0.640 \\
LSTM & 0.762 & 0.672 & 0.698 & 0.99 & 0.520 & 0.447 & 1.07 & $-4.2$ & 0.585 & 0.692 & 0.788 & 0.725 \\
EA-LSTM & 0.753 & 0.610 & 0.680 & 1.08 & 0.483 & 0.312 & 1.34 & $-22.8$ & 0.565 & 0.681 & 0.771 & 0.715 \\
\midrule
\CaRTp, $\sigma=0.3$ & 0.780 & 0.684 & 0.701 & 1.07 & 0.473 & 0.371 & 1.22 & $-6.9$ & 0.627 & 0.672 & 0.785 & 0.705 \\
\CaRTp, $\sigma=0.5$ & 0.760 & 0.673 & 0.676 & 1.07 & 0.450 & 0.334 & 1.26 & $-9.5$ & 0.631 & 0.708 & 0.766 & 0.724 \\
\bottomrule
\end{tabular}
}\\[8pt]
(b)\\[2pt]
\resizebox{\textwidth}{!}{%
\begin{tabular}{lcccccc}
\toprule
 & \multicolumn{2}{c}{gauged, 2011--2019} & gauged, dry years & arid & held-out & 3-seed ensemble \\
Pair & NSE & NSE$_{\log}$ & NSE & KGE & NSE & NSE \\
\midrule
\CaRTp $-$ LSTM & $+0.009$ & $+0.010$ & $-0.036$ & $+0.044$ & $-0.037$ & $-0.010$ \\
 & {\footnotesize [+0.003, +0.015]} & {\footnotesize [+0.000, +0.029]} & {\footnotesize [-0.051, -0.019]} & {\footnotesize [+0.003, +0.078]} & {\footnotesize [-0.081, -0.009]} & {\footnotesize [-0.014, -0.002]} \\[2pt]
\CaRTp $-$ EA-LSTM & $+0.014$ & $+0.056$ & $+0.004$ & $+0.046$ & $-0.054$ & $+0.001$ \\
 & {\footnotesize [+0.007, +0.022]} & {\footnotesize [+0.027, +0.086]} & {\footnotesize [-0.011, +0.016]} & {\footnotesize [+0.017, +0.071]} & {\footnotesize [-0.089, -0.009]} & {\footnotesize [-0.003, +0.008]} \\[2pt]
\CaRTp $-$ dHBV & $+0.016$ & $+0.200$ & $+0.038$ & $+0.075$ & $-0.037$ & $+0.011$ \\
 & {\footnotesize [+0.010, +0.020]} & {\footnotesize [+0.140, +0.244]} & {\footnotesize [+0.023, +0.059]} & {\footnotesize [+0.046, +0.103]} & {\footnotesize [-0.068, -0.001]} & {\footnotesize [+0.005, +0.017]} \\[2pt]
\CaRTp $-$ \CaRT & $+0.020$ & $+0.140$ & $+0.125$ & $+0.135$ & $+0.018$ & $+0.024$ \\
 & {\footnotesize [+0.016, +0.025]} & {\footnotesize [+0.098, +0.179]} & {\footnotesize [+0.094, +0.160]} & {\footnotesize [+0.068, +0.238]} & {\footnotesize [-0.028, +0.036]} & {\footnotesize [+0.019, +0.029]} \\[2pt]
\CaRTp $-$ HBV & $+0.052$ & $+0.802$ & $+0.329$ & $+0.507$ & $+0.035$ & $+0.055$ \\
 & {\footnotesize [+0.045, +0.061]} & {\footnotesize [+0.656, +0.918]} & {\footnotesize [+0.271, +0.430]} & {\footnotesize [+0.308, +0.861]} & {\footnotesize [+0.017, +0.064]} & {\footnotesize [+0.049, +0.065]} \\[2pt]
\midrule
\CaRTp, $\sigma=0.3$ $-$ LSTM & $+0.008$ & $+0.015$ & $-0.028$ & $+0.040$ & $-0.008$ & $-0.011$ \\
 & {\footnotesize [+0.000, +0.015]} & {\footnotesize [+0.002, +0.031]} & {\footnotesize [-0.043, -0.018]} & {\footnotesize [-0.005, +0.065]} & {\footnotesize [-0.030, +0.032]} & {\footnotesize [-0.017, -0.006]} \\[2pt]
\CaRTp, $\sigma=0.5$ $-$ LSTM & $+0.001$ & $+0.010$ & $-0.042$ & $+0.038$ & $+0.017$ & $-0.017$ \\
 & {\footnotesize [-0.008, +0.008]} & {\footnotesize [-0.005, +0.024]} & {\footnotesize [-0.059, -0.028]} & {\footnotesize [-0.005, +0.060]} & {\footnotesize [-0.013, +0.064]} & {\footnotesize [-0.022, -0.011]} \\[2pt]
\bottomrule
\end{tabular}
}
\end{table}

\section{Discussion}\label{sec:discussion}

\subsection{Can known processes describe the catchment response sufficiently?}\label{sec:disc_theory}

To the accuracy of a single trained neural network, they can. As a single model \CaRTp{} matches a stateful LSTM and an EA-LSTM trained on the same inputs on CAMELS-GB and CAMELS-US (paired differences in median NSE between $-0.004$ and $+0.008$). On the gauged Australian catchments, whose data played no part in deriving the structure, it is ahead of both by 0.009 and 0.014. This suggests that the structure of \CaRTp{} can extract enough information from the inputs to reach the accuracy of a single LSTM. All the assumptions on which the model is based were proposed in earlier work, including the probability-distributed store \citep{moore1985}, infiltration excess \citep{horton1933}, the variable source area \citep{dunne1970}, interception \citep{rutter1971}, depression storage \citep{shook2011}, the exchange of the GR models \citep{perrin2003} and the memory left by unresolved heterogeneity \citep{mori1965,harman2009}. Known processes, assembled under the laws, therefore suffice to reach that accuracy. What is needed is the aggregation of the fine-scale assumptions into catchment-scale flux relations. The remaining gaps are in the three-seed ensembles, in the low flows and in held-out catchments, largest in the arid ones, where the map from attributes to parameters limits the prediction (Section~\ref{sec:ung}).

\subsection{Do universal hydrological laws exist?}\label{sec:disc_laws}

The results support the existence of laws of form, since one structure built under the three laws performs well for 1342 catchments on two continents. The same structure carries over to 561 Australian catchments that played no part in its derivation and lie in a climate drier than either. It keeps the flow of their dry years closer to the observations than the other conceptual models. A structure built under the laws thus describes daily catchment dynamics as a dissipative network of stores whose fluxes follow from aggregated local laws and carry a memory of the unresolved composition. The best-known candidate for a catchment-scale law, the relation between the long-term evaporative ratio and the aridity index \citep{budyko1974,fu1981,zhang2001}, is a law of response, a regularity in what catchments do on average. The laws here constrain the dynamical system that produces the response, and the two kinds are complementary.

The constitutive families are also shared across the two continents, and because most refinements nest a simpler model, a catchment that does not need a family can approach its simpler form. What differs between catchments is the parameters, which one network maps from the catchment attributes, and this is where the uniqueness of place resides \citep{beven2000}.

To the question of \citet{dooge1986}, whether there are laws at the catchment scale, our answer from this sample is that there are laws constraining what a catchment model can be and defining its admissible topologies. The laws leave open the regularities in the map from catchment attributes to parameters, which data-driven methods working under the laws are well placed to find.

\subsection{What the laws bring together}\label{sec:disc_unify}

Several lines of work that have run separately can be read through the three laws, of which the conceptual models in use are the most relevant. The probability-distributed store \citep{moore1985} and the partition of GR4J \citep{perrin2003} are both aggregates of a local threshold law over a distribution of thresholds (Section~\ref{sec:laws2}), and the two models differ only in the distribution they assume. The release exponent of the quick store follows in the same way. Depth thresholds that follow a power law turn the local law of GR4J into a release proportional to $R^{5+k}$, so the exponent is a property of the threshold distribution. Law~\ref{law:dissip} also constrains the structure of these models, for instance in the exchange term of GR4J, which has a branch in which the store gains water as it fills. Under Law~\ref{law:dissip} that branch is inadmissible, and under the monotonicity of Section~\ref{sec:laws1} the admissible exchange is a loss together with an inflow driven by an exterior head (Section~\ref{sec:derive}). The storage--discharge relation of \citet{kirchner2009}, in which a catchment behaves as a first-order dynamical system, has its place under the same law. The relation holds when the downstream head is fixed and both the head and the conductance depend on one store alone. Appendix~\ref{app:realise} adds the cases in which the eliminated composition leaves no memory, so that one state suffices.

Where the units lie in series, the memory of the composition takes the form that the theory of travel times describes. For linear units it is the convolution of the input with the travel-time distribution of the network, so the travel-time distribution and the lag of the hydrograph come from one derivation. The same memory can be kept as a chain of stores \citep{nash1957}, exactly where the kernel is that of a chain, while GR4J uses travel-time kernels of finite length with one parameter. Units that fill and spill carry a memory of another kind, the one behind hillslope connectivity and the hysteresis seen between events. Their deficits follow the turning points of the accumulated forcing, and their superposition is a Prandtl--Ishlinskii operator \citep{okane2007,brokate1996}. This memory has no time scale, and an event that fills or empties every unit erases it.

These memories are what the closure relations of the representative elementary watershed were missing. Its balance equations \citep{reggiani1998,reggiani1999} left the closure relations open, and the search for them stopped at the finding that they are nonlinear, hysteretic and scale dependent \citep{beven2006}. Law~\ref{law:closure} starts from the result of Appendix~\ref{app:moments} that the exact elimination of the unresolved composition leaves the lumped stores with a memory of the forcing. The closure then becomes a construction that gives, for a stated fine-scale assumption, the states that carry the memory and the way they enter the fluxes. Table~\ref{tab:memory} lists the outcome for the local structures considered here.

Modular frameworks combine interchangeable stores and fluxes \citep{clark2008,fenicia2011,knoben2019} without a rule for which combinations are admissible. Corollary~\ref{cor:refine} supplies that rule as four operations, with each refinement nesting the model it refines where a natural limit exists. The three laws thus give a concise and unified account of these theories, each of which was developed on its own.

\subsection{Constructions the laws led to}\label{sec:disc_constr}

With the laws, processes that earlier models treated empirically or did not consider can also be built into a model, as we show for two terms of \CaRTp. In earlier models the connected fraction of a catchment is a function of the root-zone storage, and it splits the rain into runoff and infiltration \citep{moore1985,gao2019}. Our diagnosis showed that the error in the quick flow depended on how wet the catchment had been in the preceding days, information that the stores of \CaRT{} do not carry. By Law~\ref{law:closure}, an unresolved wetness of this kind enters the model as a conditional variable with a balance of its own, which we obtained by eliminating the wetness of the individual hillslope units (Appendix~\ref{app:memory}). In \CaRTp{} the connected fraction depends on this wetness and multiplies the release of the quick store. The vegetation demand is the second example, for which indices of foliar phenology already use a running mean of a temperature response \citep{jolly2005}. In the rain-fed basins of CAMELS-US the spring flow had a phase error that followed the greening of the vegetation. In \CaRTp{} the form of this lag follows from an assumption about vegetation patches. If the patches become active and dormant at a common rate, each when the temperature crosses its own threshold, the area-averaged activity is exactly an exponential moving average of a logistic function of temperature. The threshold, its width and the time constant are then constants shared by all catchments of a dataset.

\subsection{Limits}\label{sec:disc_limits}

The results show that a closure of the stated form suffices for daily streamflow. However, the physical reality of the individual processes needs further evidence. Additional observations, such as soil moisture, groundwater levels and evaporation, are needed to constrain the internal states and fluxes. The learned parameters must also be checked against independent estimates of the properties they stand for, such as the soil water capacity, the depression storage and the conductivity of the quick-flow pathways. The static attributes alone do not distinguish catchments with different types of response, and the resulting prediction errors in ungauged catchments can be reduced by perturbing the attributes and averaging the simulated flows. The closed law is an average over the states that the forcing of the training period visits, so it depends on that climate. Under a lasting change of climate or land use the distribution of the composition moves and the closure with it. This is one possible source of the changed response of catchments after long droughts. The Australian catchments, drier than any used in development, test how far the learned closure carries to another climate. \CaRT{} and \CaRTp{} are daily lumped models. But the laws are not tied to the daily step, since Laws~\ref{law:cons} and \ref{law:closure} do not depend on the time step and Law~\ref{law:dissip} holds at shorter steps as long as inertia remains negligible. Where inertia matters, as may happen for channel flow at hourly steps, a flux needs a momentum state, and the dissipation law becomes an energy balance that includes the kinetic energy.

\section{Conclusions}\label{sec:conclusions}
We asked whether a low-dimensional system built under physical laws can predict streamflow as accurately as a neural network trained on the same inputs. For single models in gauged catchments the answer is yes on 671 British and 671 US catchments (602 and 593 gauged), and it holds on 561 Australian catchments (505 gauged) whose data played no part in building the structure. Three laws, conservation, dissipation and aggregation, restrict a lumped model to a compartmental system on a flux network driven by potential differences. In this system the flux laws follow from the aggregation of local laws and the unresolved composition enters as memory. Their corollary gives the four operations by which such a model may be refined.

A parsimonious model consistent with the laws, \CaRT{} with 11 parameters per catchment, was refined in an iterative loop. The loop diagnoses deficits from the observations and derives each refinement from a stated fine-scale assumption with one of the four operations. In most cases the refinement nests the previous model, and it is tested against the observations over several seeds. AI agents helped execute the loop under the laws. The result, \CaRTp, has fourteen states and 35 parameters per catchment, with one structure for both continents. As a single model it matches a stateful LSTM and an EA-LSTM trained on the same inputs, with median NSE 0.863 against 0.869 and 0.862 on CAMELS-GB and 0.765 against 0.751 and 0.754 on CAMELS-US. It is ahead of \CaRT, HBV and the differentiable HBV by a larger margin. However, three-seed LSTM ensembles remain ahead of the \CaRTp{} ensemble by 0.003--0.012. Transferred to CAMELS-AUS with the same structure and settings, \CaRTp{} is the most accurate single model on the gauged catchments (median NSE 0.778) and, among the conceptual models, the closest to the observed volume in dry years. Removing the groups of refinements one at a time shows that runoff generation, quick release and evaporation carry most of the gain.

In response to \citet{dooge1986}, who asked whether there are hydrological laws at the catchment scale, this work supports their existence as laws of form. The three laws say what a catchment model must be and how it may be changed, and in this sense they define the admissible topologies of a catchment model. One structure built under the three laws and their corollary describes 1342 catchments on two continents and carries over unchanged to 561 catchments on a third, with the same constitutive families in every catchment. But the parameters remained each catchment's own and had to be learned, so the uniqueness of place resides in the parameters, which one network maps from the catchment attributes. 

The work thus provides evidence that catchment dynamics have a low-dimensional structure. Models with this structure are constructed systematically under physical constraints and match state-of-the-art neural networks in accuracy as single models. These models are dynamical systems of the class \eqref{eq:class} with finite states, whose equations are written in closed form. The laws and their corollary are a basis for two lines of work. Better process models can be sought within them by the four operations, tested against streamflow and against the internal states and fluxes that streamflow alone does not constrain. In parallel, the catchment can be studied with the analytical methods of dynamical systems, opening the questions of stability, memory and equifinality to direct mathematical analysis.
 
\section*{Open Research}
CAMELS-GB is available from the UK Centre for Ecology \& Hydrology \citep{coxon2020camelsgb}; CAMELS-US \citep{newman2015,addor2017camels} from the National Center for Atmospheric Research. The CAMELS-US forcing used here is Daymet version 4 \citep{thornton2021daymet4}, averaged over the CAMELS basin polygons, with USGS daily discharge extended to 2019 from the National Water Information System. CAMELS-AUS version~2 \citep{fowler2025camelsaus} is available from Zenodo (\url{https://doi.org/10.5281/zenodo.12575680}, release 2.03), and its forcing and attributes are used as distributed. Model code (\CaRT, \CaRTp, the dHBV, HBV and LSTM implementations and the fused daily kernels with their adjoints) and the model input files, with a list of the data sources and links, are available at \url{https://github.com/xiaxilin/cartplus}.

\section*{Acknowledgements}
This work was supported by the Science and Technology Facilities Council (STFC) STORMS grant (ST/Y00406X/1) and a seed fund from the Birmingham Institute for Sustainability and Climate Action (BISCA). AI agents (Claude Code from Anthropic and Codex from OpenAI) assisted the refinement loop as described in Section~\ref{sec:loop}. Claude (Anthropic) and ChatGPT (OpenAI) were used to assist with manuscript drafting, language editing, and improving clarity of presentation. All AI-assisted content was critically reviewed and revised by the author, who takes full responsibility for the scientific content, accuracy, and conclusions of this work.

\appendix
\section{The composition as memory and the states that carry it}\label{app:moments}

The lumped system \eqref{eq:split} reads $\dot{\bm z}=\bm F_0(\bm z,\bm u)+\Delta\bm F(\bm z,\bm y,\bm u)$, and everything in this appendix concerns the one term $\Delta\bm F$. Appendix~\ref{app:lose} asks what the aggregation loses and why the loss acts as a memory. It considers how the term depends on the spatial moments at one instant, how it depends on the history once the composition is eliminated, and its conditional mean given the aggregates. The departure of $\Delta\bm F$ from this mean is the part of $\mathcal R$ due to the composition. Appendix~\ref{app:realise} asks under which conditions this memory is carried by a few states, exactly, approximately or without a time scale, and Appendices~\ref{app:prop1} and \ref{app:memory} give two examples.

The appendix combines exact statements with three expansions, each in its own small quantity. The decomposition \eqref{eq:ycomp}, the elimination \eqref{eq:elim}, the conditional mean, the realisation by modes \eqref{eq:embed}, exact lumping, the two reservoirs and the stop operator \eqref{eq:stop} hold without approximation. The expansion in the amplitude of the composition, of order $\varepsilon$ for units in parallel, yields \eqref{eq:taylor}, the moment equations \eqref{eq:vardyn}--\eqref{eq:covrho} and the kernel \eqref{eq:gle}, which is exact in two linear cases. The expansion in the ratio $\zeta$ of the slow to the fast rates gives the reduction by a spectral gap, \eqref{eq:fastmode}--\eqref{eq:slowred}. The expansion in frequency, for input that varies slowly over the travel times, gives \eqref{eq:ttexpand} and \eqref{eq:series}. The three conditions are distinct, since a small spread implies neither a spectral gap nor slowly varying input.

\subsection{From aggregation to memory}\label{app:lose}

\paragraph{State decomposition.} Let the fine state obey $\dot{\bm\xi}=\bm f(\bm\xi,\bm u)$ with $\bm f$ locally Lipschitz in $\bm\xi$ and the forcing $\bm u(t)$ a catchment mean, any fixed spatial pattern of the forcing being part of $\bm f$. Let $\bar{\bm\xi}(\bm z)$ be the reference state of Section~\ref{sec:laws2}, the distribution of minimum stored potential energy for given aggregates, with $\bm\Pi\bar{\bm\xi}(\bm z)=\bm z$. It is affine in $\bm z$ where the potentials are linear in the storages with fixed coefficients and no unit reaches a bound. Where the potentials are continuously differentiable in the storages and the conditional variables, with a derivative in the storage bounded away from zero, it is Lipschitz in $\bm z$ and continuously differentiable within each region in which the same units are at a bound, with kinks where a unit reaches or leaves a bound. Where $\bar{\bm\xi}$ is differentiable, differentiating $\bm\Pi\bar{\bm\xi}(\bm z)=\bm z$ yields $\bm\Pi\,\mathrm D\bar{\bm\xi}=\bm I$, with $\mathrm D\bar{\bm\xi}$ the Jacobian of $\bar{\bm\xi}$ with respect to $\bm z$, whose columns distribute a change of each aggregate over the units. Hence $\mathrm D\bar{\bm\xi}\,\bm\Pi$ is a projection. The composition $\bm y=\bm\xi-\bar{\bm\xi}(\bm z)$ satisfies $\bm\Pi\bm y=\bm 0$ and, since $\dot{\bm y}=\dot{\bm\xi}-\mathrm D\bar{\bm\xi}(\bm z)\,\dot{\bm z}$,
\begin{equation}\label{eq:ycomp}
\dot{\bm y}=\big(\bm I-\mathrm D\bar{\bm\xi}(\bm z)\,\bm\Pi\big)\,\bm f\big(\bar{\bm\xi}(\bm z)+\bm y,\bm u\big)\equiv\bm g(\bm z,\bm y,\bm u).
\end{equation}
Together with \eqref{eq:split}, this is the fine system in the coordinates $(\bm z,\bm y)$ where $\bar{\bm\xi}$ is differentiable along the trajectory. A trajectory of $\bm z$ can stay at a point where it is not, and in general the composition obeys the integral form $\bm y(t)=\bm y(0)+\bar{\bm\xi}(\bm z(0))-\bar{\bm\xi}(\bm z(t))+\int_0^t\bm f\big(\bar{\bm\xi}(\bm z)+\bm y,\bm u\big)\,\dd s$, which needs no derivative of $\bar{\bm\xi}$ and reduces to \eqref{eq:ycomp} where one exists.

\paragraph{Instantaneous error of aggregation, the derivation of \eqref{eq:taylor}.} Let $N$ units of equal area have states $s_i$ and properties $\kappa_i$ with means $\bar s$, $\bar\kappa$ and deviations $\delta s_i$, $\delta\kappa_i$ of order $\varepsilon$, and outflow $q_i=f(s_i,\kappa_i)$ with $f$ smooth. Taylor's theorem about $(\bar s,\bar\kappa)$ gives
\begin{equation}
q_i=f+f_s\,\delta s_i+f_\kappa\,\delta\kappa_i+\tfrac12f_{ss}\,\delta s_i^2+f_{s\kappa}\,\delta s_i\,\delta\kappa_i+\tfrac12f_{\kappa\kappa}\,\delta\kappa_i^2+O(\varepsilon^3),
\end{equation}
all derivatives evaluated at the means. In the average over $i$, the first-order terms vanish because the deviations have zero mean, and the second-order terms give the variances and the covariance of \eqref{eq:taylor}. For a power law $f=as^p$ and two units at $\bar s(1\pm\varepsilon)$ the average is $a\bar s^p[(1+\varepsilon)^p+(1-\varepsilon)^p]/2$, which for $p=5$ equals $a\bar s^5(1+10\varepsilon^2+5\varepsilon^4)$. The second-order term of \eqref{eq:taylor}, $\tfrac12f_{ss}\Var(s)=\tfrac12\cdot20a\bar s^3\cdot\varepsilon^2\bar s^2=10a\bar s^5\varepsilon^2$, is the leading correction to the flux at the mean storage.

\paragraph{Order of the correction and the outlet weights.} The average above weights every unit equally, as for units in parallel. Let $w_i$ be the share of the outflow of unit $i$ that leaves the aggregate and $a_i$ its area weight. The aggregate outflow is $\sum_iw_ia_if(s_i,\kappa_i)$, and its first-order term about the means is $f_s\sum_iw_ia_i\delta s_i+f_\kappa\sum_iw_ia_i\delta\kappa_i$. The deviations satisfy $\sum_ia_i\delta s_i=0$ and $\sum_ia_i\delta\kappa_i=0$, so both sums vanish for all compositions if and only if the $w_i$ are equal. Units in parallel have $w_i=1$ and meet the condition. Units in series have $w_i=0$ except at the outlet, where $w_n=1$, and the first-order term is $a_n(f_s\,\delta s_n+f_\kappa\,\delta\kappa_n)$. The second part of this term is set by the mean state and the fixed properties and enters $\bar{\bm f}$. The first part remains, so the correction for units in series is of first order in the composition and present for identical units, the origin of the correction in \eqref{eq:series}. In general, the first-order term is $\sum_i(\partial Q/\partial\xi_i)\,y_i$, with $Q$ the sum of the fluxes that leave the aggregate. It vanishes for all compositions if and only if $\partial Q/\partial x_i$ is the same for all storages of the aggregate and $a_{ki}^{-1}\partial Q/\partial\chi_{ki}$ is the same for all units of each kind $k$. Section~\ref{sec:laws2} takes these sensitivities as the weights $w_i$, of which the shares above are the special case for outflows that depend only on the state of their own unit. An area mean has equal weights, so for units with the same balance the deviations $\chi_{ki}-\bar\chi_k$ enter the balance of $\bar\chi_k$ at second order, as $V$ does in Appendix~\ref{app:memory}.

\paragraph{Elimination of the composition.} For given trajectories $\bm z(\cdot)$ and $\bm u(\cdot)$, \eqref{eq:ycomp}, or in general its integral form, is an equation in $\bm y$ with a locally Lipschitz right-hand side. It therefore has one solution for each initial composition as long as the fine state stays in a compact part of the domain of $\bm f$ and the forcing is piecewise continuous,
\begin{equation}\label{eq:elim}
\bm y(t)=\mathcal Y_t\big[\bm y(0);\,\bm z|_{[0,t]},\,\bm u|_{[0,t]}\big],\qquad
\dot{\bm z}(t)=\bm F_0(\bm z,\bm u)+\Delta\bm F\big(\bm z,\mathcal Y_t[\bm y(0);\bm z,\bm u],\bm u\big).
\end{equation}
The composition can therefore be kept as a state, through \eqref{eq:ycomp}, or replaced by a functional of the history of the aggregates and the forcing plus the contribution of its initial value, through \eqref{eq:elim}. Whether that contribution fades is a question of stability, taken up below, since $\dot{\bm y}=\bm 0$ would keep it indefinitely. The two descriptions are the same equations in different coordinates, which is the statement of Law~\ref{law:closure}. The threshold laws of this paper are of two kinds. Excess laws $(v-c)_+$ and their powers are continuous and keep \eqref{eq:ycomp} locally Lipschitz. For units that fill to a capacity and spill, the reference state fills the units whose capacity lies below a common level and holds the others at that level. This reference state thus stays within the capacities, and the fine system is a sweeping process with a unique solution \citep{brokate1996}. In both cases the composition is a functional of the history, and \eqref{eq:elim} holds.

\paragraph{Fading of the initial composition.} Whether the initial term fades depends on the full system. Suppose that each flux depends only on the storages of the two volumes it connects and on conditional variables held fixed, and that it has the monotonicity stated in Section~\ref{sec:laws1}, as on a K-orthogonal finest partition (Section~\ref{sec:laws1}) with conductances that are taken from the source and grow with its storage. The Jacobian of $\bm f$ is then a compartmental matrix, with non-negative off-diagonal entries and column sums equal to minus the sensitivity of the outflow to the exterior. The logarithmic norm of this matrix in the water-volume norm is non-positive, so two fine trajectories under the same forcing do not move apart in water volume. They converge where, in addition, the rates are bounded uniformly in time and each volume has a path to the exterior along which the outflow sensitivities are bounded away from zero. For a single store they converge where the cumulative sensitivity $\int\partial q/\partial x\,\dd t$ diverges \citep{jacquez1993}. However, a flux modulated by a third storage, such as the quick release whose connectivity rises with the groundwater store (Section~\ref{sec:derive}), adds off-diagonal entries of the opposite sign, and the fading of the full model is then a property of the fitted system. For $\bm\chi$ the drainage is replaced by a contracting balance of its own, the condition of Appendix~\ref{app:realise}, as in \eqref{eq:Vdecay}. The full fine state forgets its initial value where, in addition, the conditional variables act on the stores in a cascade, driven by some stores and modulating others that do not drive them. In \CaRTp{} the surface layer drives the wetness, and the wetness modulates the quick release. The initial composition is part of the initial state and is forgotten with it. The stability of \eqref{eq:ycomp} alone, for a prescribed trajectory of $\bm z$, is a separate question, settled here for the two heterogeneous reservoirs and the linear networks of Appendix~\ref{app:realise} and for the balance of the hillslope wetness in \eqref{eq:Vdecay}.

\paragraph{Dynamics of the spread.} Let $\dot s_i=I_i-f(s_i,\kappa_i)$ with inputs $I_i$ and time-invariant properties $\kappa_i$. We measure the spread of the storages by their variance $V_s$ and their covariances with the properties and the input. The deviation of unit $i$ obeys $\delta\dot s_i=\delta I_i-\delta q_i$, where $\delta q_i$ is the deviation of its outflow from the mean outflow, and the expansion of $f$ about the means gives $\delta q_i=f_s\,\delta s_i+f_\kappa\,\delta\kappa_i+O(\varepsilon^2)$. Inserting this into $\dot V_s=2\,\overline{\delta s_i\,\delta\dot s_i}$ for the variance $V_s=\overline{\delta s_i^2}$ and into $\tfrac{\dd}{\dd t}\Cov(s,\kappa)=\overline{\delta\dot s_i\,\delta\kappa_i}$ for the covariance with the properties gives
\begin{align}
\dot V_s&=2\Cov(s,I)-2f_s\,V_s-2f_\kappa\Cov(s,\kappa)+O(\varepsilon^3),\label{eq:vardyn}\\
\tfrac{\dd}{\dd t}\Cov(s,\kappa)&=\Cov(I,\kappa)-f_s\Cov(s,\kappa)-f_\kappa\Var(\kappa)+O(\varepsilon^3),\label{eq:covdyn}
\end{align}
where the terms of third order are the third moments of the states and properties, among them $f_{ss}\mu_3$ with $\mu_3$ the third central moment of $s$.

The variance has two sources and one sink. Uneven inputs change it directly through $\Cov(s,I)$, which is positive when the wetter units receive more and negative when the drier ones do, and uneven properties create it through the covariance. If $\kappa$ is a release rate ($f_\kappa>0$) and the input is uncorrelated with the properties, \eqref{eq:covdyn} drives $\Cov(s,\kappa)$ towards $-f_\kappa\Var(\kappa)/f_s<0$, with the slow units holding more, so that the term $-2f_\kappa\Cov(s,\kappa)$ in \eqref{eq:vardyn} is positive. Drainage damps the variance at the rate $2f_s$, twice the linearised rate $f_s$ of the store, because the wetter units drain faster. A single deviation decays at the rate $f_s$, and the variance, its square, decays twice as fast.

The covariance with the input is itself a moment of the state. Let the input have a fixed spatial pattern, $I_i=\rho_i\bar I$ with time-invariant shares $\rho_i$ of mean one and deviations $\delta\rho_i$ of order $\varepsilon$, so that $\Cov(s,I)=\bar I\Cov(s,\rho)$. The same steps give
\begin{equation}\label{eq:covrho}
\tfrac{\dd}{\dd t}\Cov(s,\rho)=\bar I\Var(\rho)-f_s\Cov(s,\rho)-f_\kappa\Cov(\kappa,\rho)+O(\varepsilon^3),
\end{equation}
and \eqref{eq:vardyn}, \eqref{eq:covdyn} and \eqref{eq:covrho} are closed in the three moments $V_s$, $\Cov(s,\kappa)$ and $\Cov(s,\rho)$ for a given mean and forcing.

Under steady forcing with a fixed distribution of properties and a fixed pattern of input, the moments settle where their rates vanish, at $\Cov(s,\kappa)=[\bar I\Cov(\rho,\kappa)-f_\kappa\Var(\kappa)]/f_s$, $\Cov(s,\rho)=[\bar I\Var(\rho)-f_\kappa\Cov(\kappa,\rho)]/f_s$ and $V_s=[\bar I\Cov(s,\rho)-f_\kappa\Cov(s,\kappa)]/f_s$ to second order. The steady mean satisfies $\bar I=f(\bar s,\bar\kappa)$ to leading order, and $\bar s$ increases with the input where $f_s>0$. The input, and with it the settled moments, are therefore functions of the mean. Under varying forcing the moments lag behind those values, and the lag is the part of the spread that the current mean does not contain.

With the coefficients frozen at the current mean, the three equations are linear in the moments. The two covariances do not depend on the variance, so the matrix of the system is triangular, with $-f_s$ on the diagonal for each covariance and $-2f_s$ for the variance, and for $f_s>0$ the lag decays with the rates $f_s$ and $2f_s$. For $f=as^p$ the rate $f_s$ equals $p$ times the inverse $f/s$ of the residence time $s/f$. The spread therefore remembers the inputs over a time of order $1/f_s$, a fraction $1/p$ of the residence time and the whole residence time for linear units. This is the statement in Section~\ref{sec:laws3} that the moments are not functions of the current mean, and it is why the closure can be written either with the moments as extra states or as a memory of the forcing. Since the aggregate flux depends on the moments through \eqref{eq:taylor} and the moments evolve by \eqref{eq:vardyn}, \eqref{eq:covdyn} and \eqref{eq:covrho}, the aggregate flux carries a memory of the forcing. This is the concrete form of \eqref{eq:elim} for units in parallel to second order in the spread.

\paragraph{Conditional mean and the closure term.} The closed law \eqref{eq:fbar} is the conditional mean of the right-hand side of \eqref{eq:elim} for given aggregates and forcing, taken over the distribution of the composition that the forced system visits. We take this to be the occupation distribution of a bounded trajectory under stationary forcing, the time fraction that the trajectory spends at each composition, and assume that it exists and is unique. This is the assumption behind the phrase ``weighted by how often it occurs''. Boundedness is a further assumption, met where the net loss of the total storage has a positive lower bound at large storage. The part of $\mathcal R$ due to the composition is the departure of $\Delta\bm F\big(\bm z,\mathcal Y_t[\bm y(0);\bm z,\bm u],\bm u\big)$ from the conditional mean $\mathrm E[\Delta\bm F\mid\bm z,\bm u]$ and has zero conditional mean by construction. Forcing whose pattern over the units changes from event to event adds to \eqref{eq:ycomp} a term that the aggregates and their history do not determine. The fine system then also depends on this pattern and on the fluxes that people control, and the elimination \eqref{eq:elim} holds for their given history. Where the local laws are linear with identical rates this term cancels in the sum. Where they are nonlinear or heterogeneous it is the part of $\mathcal R$ of Section~\ref{sec:laws2} that no lumped structure recovers.

\subsection{Realisation and reduction of the memory}\label{app:realise}

\paragraph{Memory kernel and the two exact cases.} Linearising \eqref{eq:ycomp} about $\bm y=\bm 0$ gives $\dot{\bm y}=\bm A(\bm z,\bm u)\bm y+\bm b(\bm z,\bm u)+O(|\bm y|^2)$ with $\bm A=\partial_{\bm y}\bm g|_{\bm 0}=(\bm I-\mathrm D\bar{\bm\xi}\,\bm\Pi)\,\partial_{\bm\xi}\bm f|_{\bar{\bm\xi}}$ and $\bm b=\bm g(\bm z,\bm 0,\bm u)$ the rate at which heterogeneous units drift apart from the reference state, and $\Delta\bm F=\bm C(\bm z,\bm u)\bm y+O(|\bm y|^2)$ with $\bm C=\bm\Pi\,\partial_{\bm\xi}\bm f|_{\bar{\bm\xi}}$. The variation of constants with the state-transition matrix $\Phi(t,s)$ of $\bm A$, the solution of $\partial_t\Phi=\bm A\Phi$ with $\Phi(s,s)=\bm I$ that carries a composition from time $s$ to time $t$, then gives
\begin{equation}\label{eq:gle}
\dot{\bm z}(t)=\bm F_0(\bm z,\bm u)+\int_0^t\bm K(t,s)\,\dd s+\bm C(t)\Phi(t,0)\bm y(0)+O(|\bm y|^2),
\end{equation}
with $\bm K(t,s)=\bm C(t)\Phi(t,s)\bm b(\bm z(s),\bm u(s))$. The first term is the law of the reference state, the second a memory of the history of $\bm z$ and $\bm u$, and the third the contribution of the initial composition. This decomposition has the structure of the generalised Langevin equation \citep{mori1965}. The remainder, which collects the history of the quadratic terms, is bounded on a finite window by a multiple of $\sup|\bm y|^2$. However, a bound uniform in time needs the stability discussed below.

Equation~\eqref{eq:gle} is exact when $\bm g$ is affine and $\Delta\bm F$ linear in $\bm y$. The three terms have the roles of the Markov, memory and noise terms of the Mori--Zwanzig decomposition for the projection onto the reference state $\bm y=\bm 0$, whose Markov term is $\bm F_0$. Where the coefficients depend on the state, the orthogonal dynamics of that decomposition, which act on observables, differ from $\Phi$. The conditional-expectation projection moves $\mathrm E[\Delta\bm F\mid\bm z,\bm u]$ into the Markov term, which then equals the closed law $\bar{\bm f}$ of \eqref{eq:fbar}. The memory and noise terms of this projection together equal the part of $\mathcal R$ due to the composition, with zero conditional mean \citep{zwanzig1973,chorin2002}. Equation~\eqref{eq:gle} is exact in two cases of this paper, in which the composition enters linearly. In the first, the two heterogeneous reservoirs of Appendix~\ref{app:prop1} with $\varepsilon\ne0$ have $\bm y=h$, $\bm A=-r_0m(W)$ and $\bm b=r_0m(W)W$. The kernel acts on the history of $W$ with weights that decay over the residence time, and the initial imbalance is forgotten once the cumulative drainage $\int r_0m(W)\,\dd t$ diverges. In the second, a linear network of stores has constant $\bm A$, so $\Phi$ depends on $t-s$ only. The kernel acting on the history of $\bm z$ and $\bm u$ is then a sum of exponentials where $\bm A$ is diagonalisable, which become damped oscillations where the rates are complex, as the modal decomposition below shows.

\paragraph{Markov embedding.} Equation~\eqref{eq:elim} is a functional of the history because the composition was eliminated from \eqref{eq:ycomp}. Representing the functional in \eqref{eq:elim} by states means choosing functions of the history which themselves obey ordinary differential equations driven by $\bm z$ and $\bm u$, so that the model integrates them instead of storing the history. A history functional is carried exactly by finitely many states when it is the output of a finite-dimensional system, and approximately otherwise. The composition is the output of such a system, since it obeys the finite set of equations \eqref{eq:ycomp} driven by $\bm z$ and $\bm u$, and retaining all of it represents its memory exactly. Retaining all of it can, however, return most of the fine state to the model. The question is therefore which parts of the composition must be kept to reproduce its effect on the aggregates. This is the Markov embedding of Section~\ref{sec:laws3} and Appendix~\ref{app:memory}. Each case below names the state and its equation and explains why it reproduces the functional.

\emph{Exact state realisation and modal decomposition.} The composition records how the water of given aggregates is distributed over the units. Two states with the same storage can hold more water upstream in one case and downstream in the other, and they then release different outflows. Linearised as in \eqref{eq:gle}, the composition evolves according to $\dot{\bm y}=\bm A\bm y+\bm b(\bm z,\bm u)$. Its contribution to the aggregate rates of change is $\Delta\bm F=\bm C\bm y$. For constant $\bm A$, a decomposition into modes identifies the patterns of the composition, the rates at which they relax and their effects on the aggregates. Where $\bm A$ is diagonalisable, as it is when its rates are distinct, $\bm A=-\bm V\bm\Lambda\bm V^{-1}$ with the relaxation rates $\lambda_j$ on the diagonal of $\bm\Lambda$, ordered by their real parts and with complex ones in conjugate pairs. The columns of $\bm V$ are the modes. In the modal coordinates $\bm a=\bm V^{-1}\bm y$, the amplitudes of the modes, with $\bm\beta=\bm V^{-1}\bm b$ and $\bm c_j$ the $j$-th column of $\bm C\bm V$, the composition separates into scalar equations,
\begin{equation}\label{eq:embed}
\dot a_j=\beta_j(\bm z,\bm u)-\lambda_j a_j
\quad\Longleftrightarrow\quad
a_j(t)=e^{-\lambda_jt}a_j(0)+\int_0^te^{-\lambda_j(t-s)}\beta_j(s)\,\dd s,\qquad \Delta\bm F=\sum_j\bm c_ja_j .
\end{equation}
The model integrates the left-hand form and thereby carries the right-hand form. One state $a_j$ thus holds the history of its drive $\beta_j$, weighted by $e^{-\lambda_j(t-s)}$, and adds $\bm c_ja_j$ to the rates of the aggregates. Where $\operatorname{Re}\lambda_j>0$, the drive keeps exciting the mode, the relaxation weakens the older contributions, and the initial composition fades at the same rate. A real rate needs one real state. A complex pair $\mu\pm i\nu$ needs two, the real and imaginary parts of $a_j$, whose amplitude decays as $e^{-\mu\tau}$ while their phase turns by $-\nu\tau$, and the conjugate terms in $\Delta\bm F$ add to a real quantity. Keeping all modes reproduces the linearised \eqref{eq:elim} exactly, and \eqref{eq:elim} itself where $\bm g$ is affine and $\Delta\bm F$ linear, as for the linear networks of \eqref{eq:gle}. A scalar composition needs no eigenvectors and admits a rate that depends on the state, as in the two reservoirs, where $a_1=h$, $\lambda_1=r_0m(W)$ and $\beta_1=r_0m(W)W$. The rate then changes the weights of the history at no cost to the state equation.

Where rates repeat, $\bm A$ may lack enough independent eigenvectors to separate the composition into independent modes. Some states then remain coupled: for example, $\dot a_2=-\lambda a_2$ and $\dot a_1=-\lambda a_1+a_2$ give $a_2(t)=a_2(0)e^{-\lambda t}$ and $a_1(t)=[a_1(0)+t\,a_2(0)]e^{-\lambda t}$. The factor $t$ arises because one state drives another with the same relaxation rate. Longer chains of such coupling produce higher powers of time, so the memory kernel can contain terms $\tau^k e^{-\lambda\tau}$. The state representation remains exact for the linear system and can be computed by integrating the coupled equations directly, without constructing the Jordan form.

\emph{Decay of the modes.} Under Law~\ref{law:dissip} and the conditions stated below, the modes of a linear network relax, so that the memory of its composition fades. For a linear network lumped into one aggregate, $\bm A$ acts on the compositions, the vectors with $\bm\Pi\bm y=\bm 0$, as $\bm M=(\bm I-\mathrm D\bar{\bm\xi}\,\bm\Pi)\bm N$ does. Here $\bm N=\partial_{\bm\xi}\bm f$ is the network matrix and $\bm\psi=\mathrm D\bar{\bm\xi}$ holds the shares in which the reference state distributes a change of the aggregate, proportional to $\dd x_i/\dd H_i$ and summing to one. Every unit is taken to lie between its bounds at the reference state, so that each $\psi_i$ is positive. On a K-orthogonal finest partition with the monotonicity stated in Section~\ref{sec:laws1}, the off-diagonal entries of $\bm N$ are non-negative, and so is $-\bm\Pi\bm N$, the increase of the outflow to the exterior per unit storage of each unit. The off-diagonal entries of $\bm M=\bm N-\bm\psi\,\bm\Pi\bm N$ are then non-negative, and its columns sum to zero, since $\bm\Pi\bm\psi=1$. By the Gershgorin theorem applied to its columns, each eigenvalue of $\bm M$ lies in a disc centred at a non-positive diagonal entry with a radius equal to that entry's magnitude, so it has a non-positive real part and reaches the imaginary axis only at zero. The zero column sums give $\bm M$ a zero eigenvalue, which belongs to the aggregate, since $\bm M$ leaves the aggregate unchanged. The term $-\bm\psi\,\bm\Pi\bm N$ links every unit with an outflow to every unit, since each $\psi_i$ is positive. Where each unit has a path to the exterior, all units are then connected with one another and $\bm M$ is irreducible, so its zero eigenvalue is simple by the Perron--Frobenius theorem \citep{jacquez1993}. The relaxation rates $\lambda_j$, the negatives of the other eigenvalues, therefore have positive real parts. For linear exchange in both directions, with constant conductances and linear relations between storage and potential, $\bm A$, the restriction of $\bm M$ to the compositions, is self-adjoint in the inner product weighted by $\dd H_i/\dd x_i$, and the rates are real. Decay leaves the number of states unchanged, and an exact representation keeps one state for each component of the composition. Fewer states need further structure in the rates $\lambda_j$, the drives $\beta_j$ or the effects $\bm c_j$ of the modes in \eqref{eq:embed}, which the reductions below use.

\paragraph{Reduction by a spectral gap.} Representing the memory by fewer states than modes needs an assumption, outside the exact cases noted below. We assume a spectral gap, $\operatorname{Re}\lambda_{q+1}\gg\Omega_s$, with the first $q$ rates slow and the others fast and with a complex pair kept or dropped as a whole. Here $\Omega_s$ is the largest of the moduli $|\lambda_j|$ of the kept modes, the rate at which the forcing changes and the rates of the slow closed system. The gap has a physical basis in the separation of hydrological time scales. Water redistributes within an event over hours, soils and hillslopes dry over days to weeks, and groundwater drains over months to years. Recession analysis uses this separation to tell quick flow from base flow \citep{tallaksen1995}. At a daily step, a mode that relaxes within hours is close to equilibrium with the daily forcing and enters the effective drift whatever its origin. Integrating the history form of \eqref{eq:embed} by parts gives
\begin{equation}\label{eq:fastmode}
a_j(t)=\frac{\beta_j(t)}{\lambda_j}-\frac{1}{\lambda_j}\int_0^te^{-\lambda_j(t-s)}\dot\beta_j(s)\,\dd s+e^{-\lambda_jt}\Big(a_j(0)-\frac{\beta_j(0)}{\lambda_j}\Big),
\end{equation}
so that a fast mode equals $\beta_j(\bm z,\bm u)/\lambda_j$ up to the history term and the initial term. The history term is bounded by the largest $|\dot\beta_j|$ over the history divided by $|\lambda_j|\operatorname{Re}\lambda_j$, and the initial term decays as $e^{-t\operatorname{Re}\lambda_j}$. We neglect the history term and the decaying initial term. The fast modes are then approximated by $\beta_j(\bm z,\bm u)/\lambda_j$, a function of the present state, and contribute to the effective drift, while the slow modes are kept as states,
\begin{equation}\label{eq:reduced}
\dot{\bm z}=\bm F_0(\bm z,\bm u)+\sum_{j>q}\frac{\bm c_j\beta_j(\bm z,\bm u)}{\lambda_j}+\sum_{j\le q}\bm c_ja_j,\qquad \dot a_j=\beta_j(\bm z,\bm u)-\lambda_ja_j,\quad j\le q .
\end{equation}
The drift changes with the aggregates on the slow time scale, $|\dot\beta_j|\sim\Omega_s|\beta_j|$. The relative error of each fast mode, on the scale of its typical size, is therefore of the order of $\Omega_s/\operatorname{Re}\lambda_{q+1}$. The modulus of the fast rate cancels, so the decay rate alone decides whether a mode is fast.

The error of the reduced system \eqref{eq:reduced} remains of this order over long times where its slow dynamics damp perturbations at a uniform exponential rate, which we assume together with the gap and with couplings $\bm c_j$ that stay bounded as the fast rates grow. The damping has a physical basis in dissipation, since a store that drains or evaporates more when it holds more water forgets a perturbation of its storage at the rate $\partial L/\partial S$ at which the losses $L$ grow with the storage $S$. For a lumped store this holds at the reference state of Section~\ref{sec:laws2}, where the losses are those of units at one potential, and a flux modulated by a third store can break it (Appendix~\ref{app:lose}). A snowpack, whose melt follows the energy supply, and a store below its spill threshold lose no more water when they hold more, and keep a perturbation until they melt out or spill, as do the units of the rate-independent case.

For a nonlinear $\bm g$ the same reduction rests on a slow manifold, since the elimination \eqref{eq:elim} needs only a solution of \eqref{eq:ycomp} and the memory exists where the linearisation about $\bm y=\bm 0$ fails. Where the composition still splits into a slow and a fast part, the reduction by a spectral gap carries over to a composition of large amplitude. Collect the aggregates and the slow part of the composition in $\bm\sigma$, the fast part in $\bm\phi$, and let $\zeta$ be the ratio of the slow to the fast rates, $\Omega_s/\operatorname{Re}\lambda_{q+1}$ in the linear case. The system then reads
\begin{equation}\label{eq:fastslow}
\dot{\bm\sigma}=\bm F(\bm\sigma,\bm\phi,\bm u),\qquad \zeta\,\dot{\bm\phi}=\bm G(\bm\sigma,\bm\phi,\bm u),
\end{equation}
with the forcing varying on the slow time scale. Daily forcing changes in steps, so the reduction holds within each day, and the fast part relaxes again after each step. Over a boundary layer as long as the fast time scale after each step the error is of order one, and its contribution to the daily fluxes is of order $\zeta$. Setting $\zeta=0$ defines the critical manifold $\bm G=\bm 0$, solved as $\bm\phi=\bm h_0(\bm\sigma,\bm u)$, on which the fast part is in equilibrium with the slow one. The manifold is normally attracting where the eigenvalues of $\partial\bm G/\partial\bm\phi$ on it have real parts below a negative bound, so that a fast part displaced from it returns. Fenichel's theorem \citep{fenichel1979} then provides an invariant slow manifold $\bm\phi=\bm h_0+O(\zeta)$ that attracts nearby trajectories, and the slow dynamics reduce to
\begin{equation}\label{eq:slowred}
\dot{\bm\sigma}=\bm F\big(\bm\sigma,\bm h_0(\bm\sigma,\bm u),\bm u\big)+O(\zeta).
\end{equation}
For linear modes $\bm h_0$ is $\beta_j/\lambda_j$ for each fast mode and \eqref{eq:slowred} is \eqref{eq:reduced}. The conditions concern the Jacobian of the fast part on the manifold and therefore hold for a composition that the linearisation about $\bm y=\bm 0$ does not describe.

The spectral gap is the premise we adopt for every construction of Section~\ref{sec:laws3} that keeps fewer modes than the composition has, among them the single-mode variables of Appendix~\ref{app:memory}. Together with the uniform damping and, for a nonlinear $\bm g$, the conditions of Fenichel's theorem, it suffices for the reduction.

\paragraph{Reduction without a spectral gap.} The reduction is exact in two cases. In the first, a mode can be dropped because the aggregates do not see it ($\bm c_j=0$) or because nothing drives it and it starts at zero ($\beta_j=0$ and $a_j(0)=0$). An undriven mode that starts away from zero fades at the rate $\operatorname{Re}\lambda_j$, and the error of dropping it decays at that rate. The same holds for a nonlinear $\bm g$ where the reference state is invariant, so that the forcing creates no differences between the units, and where dissipation removes the differences that exist at a rate at least proportional to their size. The composition then decays under any forcing and at any amplitude, and the aggregates follow $\bm F_0$ once it has faded. 

In the second, the kept states are exactly lumpable in the sense of Section~\ref{sec:laws2}, as the total storage of parallel linear stores with equal rates. Splitting the volume (operation a of Corollary~\ref{cor:refine}) is also exact at any spread, since it keeps the composition itself as states, one per unit, as for the snow bands, and so reduces nothing. Outside these cases a finite set of states approximates the memory. Where the rates form a continuum, as in heterogeneous linear stores that give power-law recessions \citep{harman2009}, the kernel $\sum_j\bm c_j\beta_je^{-\lambda_j\tau}$ becomes an integral over the rate distribution. Where the spread is small, for linear and nonlinear $\bm g$ alike, the moments of the composition are such states. For units in parallel the state is $\big(\bar s,V_s,\Cov(s,\kappa)\big)$, with $\Cov(s,\rho)$ added where the input has a spatial pattern. The moments evolve by \eqref{eq:vardyn}, \eqref{eq:covdyn} and \eqref{eq:covrho}, and the aggregate flux follows from \eqref{eq:taylor}. The third moments enter at the next order and are closed by a hypothesis, such as $\mu_3=0$ for a symmetric spread. Where nothing is known beyond a fading memory, a general approximation remains. Let $\bm v=(\bm z,\bm u)$ be the inputs of the history functional. States $\bm r$ that relax linearly under them and a polynomial output $\bm P$,
\begin{equation}\label{eq:fadingapprox}
\dot{\bm r}=-\bm\Lambda_r\bm r+\bm B_r\bm v,\qquad \Delta\bm F\approx\bm P(\bm r),
\end{equation}
approximate any time-invariant history functional with fading memory uniformly over inputs bounded by $M_1$ whose rates of change are bounded by $M_2$ \citep{boyd1985}. For daily forcing, which changes in steps, the discrete-time form of this result needs only the bound $M_1$ \citep{grigoryeva2018}. The states $\bm r$ have in general no physical meaning.

\paragraph{Travel-time memory of units in series.} Along a chain of stores with similar rates there is no gap. For a linear network the memory is then reduced by expanding its transfer function in frequency, a function that takes the place of the matrix $\bm A$. Let the units form a finite, stable and time-invariant network of linear stores, so that the outflow is a convolution of the input with a fixed kernel. They are connected in any way that conserves water and eventually drains all of it to the outlet, so that the travel-time distribution has a finite mean $T$ and variance $\sigma^2$. They start empty, which removes the initial term of \eqref{eq:gle}, and receive the input $p(t)$ in fixed proportions. The outflow is $Q=h*p$ with $h$ the travel-time distribution of the input, and the total storage obeys $\dot z=p-Q$. In the Laplace domain, with hats for transforms, $\hat Q=H\hat p$ and $\hat z=(1-H)\hat p/s$, so the departure of the outflow from the quasi-steady closure $z/T$ is $\hat Q-\hat z/T=[H-(1-H)/(Ts)]\,\hat p$. With $H(s)=1-Ts+\tfrac12(\sigma^2+T^2)s^2+O(s^3)$, the storage per unit input is $(1-H)/s=T-\tfrac12(\sigma^2+T^2)s+O(s^2)$, and the factor of $\hat p$ is
\begin{equation}\label{eq:ttexpand}
H-\frac{1-H}{Ts}=\frac{\sigma^2-T^2}{2T}\,s+O(s^2).
\end{equation}
Multiplication by $s$ in \eqref{eq:ttexpand} is differentiation in time, which yields, to first order in frequency,
\begin{equation}\label{eq:series}
Q=\frac{z}{T}+\frac{\sigma^2-T^2}{2T}\,\dot p+O(\ddot p).
\end{equation}
The first term is the quasi-steady closure and the second the correction. The approximation holds where the forcing varies slowly over the whole travel-time distribution, which a long tail makes much longer than $T$, and the next term carries the third moment of $h$ and $\ddot p$. For stores in series with rates $k_i$ and the input at the head, the times spent in the stores are independent and exponential, so their means and variances add, $T=\sum_i1/k_i$ and $\sigma^2=\sum_i1/k_i^2$, so $\sigma<T$ for two stores or more. For $n$ identical stores $\sigma^2=T^2/n$, and relative to the quasi-steady flux $z/T\approx p$ the correction is $-\frac{n-1}{2n}\,T\dot p/p$. For stores in parallel the Cauchy--Schwarz inequality gives $\sigma\ge T$, and the coefficient is of second order in the spread of the rates, as in \eqref{eq:taylor}. For two stores in series with $k_f\gg k_s$ the coefficient $(\sigma^2-T^2)/(2T)\approx-1/k_f$ is a fraction of order $k_s/k_f$ of $T$, the order of the error of the reduction by a spectral gap. A chain with one slow store therefore reduces to that store either way. A chain of similar stores keeps its stores, which hold the water in transit, or is closed in terms of $z$ with the first-order correction of \eqref{eq:series}, which adds no state but holds only for input that varies slowly compared with $T$, which daily rain does not for the quick flow. The travel-time distribution acts on the input $p$ and includes the delay of the resolved store itself, whereas the composition kernel of \eqref{eq:gle} acts on the history of $z$ and $p$. For units in parallel under shared forcing the reference state drifts only through the heterogeneity, $\bm b=O(\varepsilon)$, and the composition kernel vanishes at $\varepsilon=0$. But for units in series that state is not invariant even for identical units, since the input enters at the head and the outflow leaves at the end, so $\bm b\ne\bm 0$ and the composition kernel is of first order.

\paragraph{Compositions with thresholds.} A threshold law can make $\bm g$ non-differentiable at the threshold, and two cases differ. Where the units share the variable that drives a threshold law, or lateral exchange brings them to share it (Appendix~\ref{app:memory}), the aggregation \eqref{eq:aggthr} is exact in that variable and as smooth as the local law and the threshold distribution allow, continuously differentiable for an excess law and a continuous distribution. The memory then sits in the driving variable, a single relaxing mode that remains after the reduction by a spectral gap, and its balance is a closure of its own (Appendix~\ref{app:memory}). Where each unit crosses its threshold with its own state, the set of units above their thresholds defines a regime, within which every unit follows a smooth law. The composition then forms a switching system, whose equations change from one regime to the next as units cross their thresholds, and it relaxes at the rates of the regime it is in. Within a regime that has a spectral gap and lasts longer than the fast relaxation, it is reduced by that gap. A regime without drainage does not relax, as for units that store between two bounds, and neither a gap nor an expansion in frequency applies there.

\emph{Rate-independent memory.} For these units the reduction follows from the bounds. Let the units be in parallel under uniform forcing, lose water at a rate that does not depend on their state and have no outflow of their own between empty and full, such as units that fill to a capacity $c$ and spill. Units in series do not share the forcing, because the spill of one unit forces the next. Their stops are coupled through the spill, the deficits are no longer a superposition, and the turning points of the accumulated forcing describe them only approximately. \CaRTp{} lets each depression spill to the flow path on its own, the parallel case (Section~\ref{sec:derive}). With $U(t)=\int_0^t(p-e)\,\dd r$ the accumulated net forcing, common to all units, the deficit $d\in[0,c]$ of each unit follows the forcing one to one and stops at either bound,
\begin{equation}\label{eq:stop}
\dot d=-\dot U\quad\text{for }0<d<c,\qquad d=\mathcal S_c[U],
\end{equation}
the stop operator. A store of such units has the deficits $\mathcal S_c[U]$ weighted by its capacity distribution, and this superposition is a Prandtl--Ishlinskii operator \citep{brokate1996}. Between the bounds the rate of change does not depend on $d$, so $\bm A=\bm 0$, $\Phi=\bm I$ and no mode relaxes, and there are no modes for a spectral gap to separate. Since the differential form $\dd d=-\dd U$ contains no $\dd t$, the memory ignores how fast the path of the accumulated forcing is travelled, $\mathcal S_c[U\circ\varphi]=\mathcal S_c[U]\circ\varphi$ for any increasing change of time $\varphi$. In contrast, a kernel weights the history by elapsed time. Under uniform forcing every deficit between the bounds moves by the same amount, so the shape of the deficit distribution changes only where units stop at a bound, and there $\bm g$ is not differentiable. The memory is set by the bounds. Starting full, drying by $E$ and then wetting by $P_w<E$ leaves the deficits
\begin{equation}\label{eq:onereversal}
d(c)=\max\big\{\min(c,E)-P_w,\,0\big\},
\end{equation}
so that every unit with a capacity larger than $E$ has the same deficit $E-P_w$ and, where the capacities span $E$, the deficit distribution takes a shape that a shift or scaling of the capacity distribution does not give in general. The two numbers $E$ and $P_w$ are the state, and after repeated reversals the state is the current value of $U$ together with the list of its turning points in the order they occurred. Each reversal adds to the end of the list the value of $U$ at which the forcing turns. When $U$ later falls below a minimum in the list, or rises above a maximum in it, that entry and all entries after it are deleted, because every unit whose deficit depended on them has been emptied or filled again. Drying to $-30$\,mm, wetting to $-20$\,mm and drying to $-26$\,mm produces the list $(-30,-20,-26)$, and drying on below $-30$\,mm deletes all three. Along the list that remains the minima rise and the maxima fall, and a complete filling or emptying deletes the whole list. Keeping the last one or two levels is a truncated embedding with an error that depends on the discarded entries. Where evaporation depends on the storage, the state follows the paths of accumulated rain and accumulated demand separately, and no longer the single net forcing $U$. The memory still ignores how fast these paths are travelled, but rain and demand now act in an order that $U$ does not record, so the turning points of $U$ only approximate the state.

\paragraph{What the model uses.} \CaRTp{} uses part of these results. The conditional variables are area means $\bar{\bm\chi}$ of local conditional variables, to which the split \eqref{eq:split} and the elimination \eqref{eq:elim} apply as to the stores (Appendix~\ref{app:memory}, operation d). One state carries the hillslope wetness $\theta$ (Appendix~\ref{app:memory}). It does so exactly on the synchronised state $\bm y=\bm 0$, which attracts the units under the shared forcing by \eqref{eq:Vdecay}, and otherwise under a spectral gap, \eqref{eq:fastslow} and \eqref{eq:slowred}. The scalar modes of \eqref{eq:embed} with a constant rate yield the exponential moving averages $\omega$ and $v$. The threshold aggregation \eqref{eq:aggthr} with a shared driving variable leads to the connected fraction $A(\theta)$ and the infiltration capacities scaled by $\omega$. The travel-time memory of \eqref{eq:ttexpand} is kept as water in transit in the unit hydrographs. The depressions keep no turning points, because their aggregation \eqref{eq:pareto} assumes the deficits reached by filling from empty, a state without turning points, and is exact while they fill (Section~\ref{sec:laws2}). The fading of the initial composition underlies the spin-up (Appendix~\ref{app:numerics}). The outlet weights and the moment equations \eqref{eq:vardyn}--\eqref{eq:covrho} support the statements of Section~\ref{sec:laws3} on the order and the time scale of the memory and do not enter \CaRTp{} as states.

Table~\ref{tab:memory} lists, for each local structure, the states that carry the memory after these reductions. Appendix~\ref{app:memory} builds one such state, the hillslope wetness $\theta$, from a diagnosed deficit to its daily update.

\section{Two heterogeneous reservoirs: derivation and discrete form}\label{app:prop1}

\paragraph{Derivation.} The two units obey $\dot s_1=G-r_0(1+\varepsilon)m(W)s_1$ and $\dot s_2=G-r_0(1-\varepsilon)m(W)s_2$. With $W=(s_1+s_2)/2$ and $D=(s_1-s_2)/2$,
\begin{equation}
\dot W=G-r_0m(W)\,[W+\varepsilon D],\qquad \dot D=-r_0m(W)\,[D+\varepsilon W].
\end{equation}
Setting $h=-D/\varepsilon$, which requires $\varepsilon\ne0$, gives \eqref{eq:hstate} exactly, for any $m$ and any $G(t)$. At $\varepsilon=0$ the total obeys $\dot W=G-r_0m(W)W$ on its own. The composition vanishes initially when $D(0)=0$, i.e.\ $h(0)=0$. The equation for $h$ is linear with the time-varying rate $r_0m(W(t))$, and with $M(t)=\int_0^tm(W(v))\,\dd v$ its solution is
\begin{equation}\label{eq:hkernel}
h(t)=h(0)\,e^{-r_0M(t)}+\int_0^t r_0m(W(s))\,e^{-r_0[M(t)-M(s)]}\,W(s)\,\dd s,
\end{equation}
a weighted average of the past values of $W$ with weights that decay over the residence time of the store, $1/(r_0m)$. Substituting \eqref{eq:hkernel} into the equation for $W$ gives the single-store law plus $\varepsilon^2r_0m(W)h$, the memory integral of \eqref{eq:hstate}. An initial composition $h(0)\ne0$ adds the transient $h(0)e^{-r_0M(t)}$, which is forgotten after a few residence times.

\paragraph{Discrete time.} Integrating the equation for $h$ over one day with $m$ and $W$ frozen, as the stores themselves are integrated, gives $h_{t+1}=\alpha_th_t+(1-\alpha_t)W_t$ with $\alpha_t=e^{-r_0m(W_t)}$, hence $h_t=\big(\prod_{u<t}\alpha_u\big)h_0+\sum_{s<t}(1-\alpha_s)\big(\prod_{s<u<t}\alpha_u\big)W_s$. Only when $\alpha$ is constant (linear units, $m\equiv1$) does this reduce to
\begin{equation}
h_t=\alpha^th_0+(1-\alpha)\sum_{s<t}\alpha^{t-1-s}W_s,
\end{equation}
a geometric kernel, i.e.\ an exponential moving average of $W$ with time constant $-1/\ln\alpha=1/r_0$, and otherwise the kernel is state dependent. This elimination applies to any linear daily update of the composition $c_{t+1}=\alpha c_t-\gamma W_t$, resulting in $c_t=\alpha^tc_0-\gamma\sum_{s<t}\alpha^{t-1-s}W_s$, a form that includes the exponential moving averages used in Section~\ref{sec:refine} (Appendix~\ref{app:memory}).

\section{Constructing a memory variable: the hillslope wetness}\label{app:memory}

The hillslope wetness $\theta$, the near-surface wetness $\omega$ and the vegetation activity $v$, three of the fourteen states of \CaRTp, carry no water and follow the forcing with a lag. They are conditional variables of operation (d) of Corollary~\ref{cor:refine}, area means $\bar{\bm\chi}$ of local conditional variables (Section~\ref{sec:laws2}), and they carry the memory form of Law~\ref{law:closure}. The construction of $\theta$ uses every element of that law, an aggregation over thresholds, a memory of the forcing and a multiplier. This appendix follows it from the hypothesis on the hillslope units to the daily update, while the other two variables follow the same template with simpler choices. In Table~\ref{tab:memory}, $\theta$ is the shared driving variable of a threshold aggregate.

\paragraph{Step 1: the hypothesis.} The deficit is the one of Section~\ref{sec:derive}, an event-scale error of the quick flow that depended on antecedent wetness (Section~\ref{sec:diag}). The quick store $R$ is one volume, and the stores alone do not resolve which parts of the hillslope are connected to the channel. Field studies describe hillslope units that release water only once their wetness exceeds a local threshold, with thresholds that vary from place to place. We therefore take units $i$ with area weights $a_i$ that share the quick-flow depth $R$ and release at the GR4J rate $a_fR^5$ only when their own wetness $\theta_i$ exceeds the unit's threshold $\theta_{c,i}$, with a fraction $A_{\min}$ next to the channel always connected (Supporting Information S2). The wetness $\theta_i$ is a local conditional variable of Section~\ref{sec:laws2}. For its balance we give each unit the form of a production store with a linear loss, the form of \eqref{eq:theta}, under the input $P_n$ and demand $E_n$ shared by the units. We also let neighbouring units exchange wetness,
\begin{equation}\label{eq:unitbal}
C_\theta\dot\theta_i=f(\theta_i,P_n,E_n)+\sum_jk_{ij}(\theta_j-\theta_i),\qquad f(\theta,P_n,E_n)=P_n(1-\theta^2)-E_n\theta(2-\theta)-C_\theta\theta/\tau_\theta ,
\end{equation}
with $k_{ij}\ge0$ and $a_ik_{ij}=a_jk_{ji}$, so that the exchange conserves the area-weighted mean. The right-hand side divided by $C_\theta$ is the component of $\bm\gamma$ of Section~\ref{sec:laws2} for unit $i$. The wetness stays in $[0,1]$, since at $\theta_i=0$ both terms are non-negative and at $\theta_i=1$ their sum is negative. The input $P_n$ and the demand $E_n$ are those that reach the root zone after the surface layer and the infiltration excess, the forcing shared with the near-stream zone (Supporting Information S1). The input enters with the efficiency $1-\theta_i^2$, so that the wetness saturates at one, and the demand removes $\theta_i(2-\theta_i)$, so that the wetness falls to zero under drying. These are the forms of the GR4J partition $\eta=s^2$ and evaporation $\varepsilon=2s-s^2$ of Section~\ref{sec:model}. The loss $C_\theta\theta_i/\tau_\theta$ is the drainage that disconnects the hillslope between events. The balance is postulated by analogy with the production store and is the closure hypothesis. The construction adds the scale $C_\theta$ and the loss time $\tau_\theta$ as two parameters per catchment and $\theta_c$ and $s_\theta$ as two global constants.

\paragraph{Step 2: aggregation.} Let $\theta=\sum_ia_i\theta_i$ be the mean wetness, the area mean $\bar\chi_k$ of Section~\ref{sec:laws2}, and $y_i=\theta_i-\theta$ the deviations, with $\sum_ia_iy_i=0$ and variance $V=\sum_ia_iy_i^2$. Since $f$ is quadratic in $\theta$, summing \eqref{eq:unitbal} with the weights $a_i$ yields the mean exactly, and subtracting gives the deviations,
\begin{equation}\label{eq:meandev}
C_\theta\dot\theta=f(\theta,P_n,E_n)+(E_n-P_n)V,\qquad
C_\theta\dot y_i=-C_\theta\lambda(\theta)\,y_i-\sum_jk_{ij}(y_i-y_j)+(E_n-P_n)(y_i^2-V),
\end{equation}
with $\lambda(\theta)=1/\tau_\theta+[2P_n\theta+2E_n(1-\theta)]/C_\theta$. This is the split \eqref{eq:chisplit} for $\theta$, with $C_\theta F_0=f(\theta,P_n,E_n)$ and $C_\theta\Delta F=(E_n-P_n)V$. The exchange drops out of the mean and acts on the deviations as a weighted graph Laplacian $\bm L$. The eigenvalue zero of $\bm L$ belongs to the uniform vector, which $\sum_ia_iy_i=0$ excludes, and the smallest positive eigenvalue $\mu_1$ exists when every unit is linked to the rest. On the set $y=0$, to which step 3 reduces the units, every unit has the mean wetness $\theta$, and summing the release of the units gives
\begin{equation}\label{eq:appA}
q_r=a_fR^5\Big[A_{\min}+(1-A_{\min})\sum_ia'_i\,\mathbf 1[\theta>\theta_{c,i}]\Big]=a_f\,A(\theta)\,R^5,\qquad A=A_{\min}+(1-A_{\min})F_\theta(\theta),
\end{equation}
with $a'_i$ the weights within the part that is not always connected and $F_\theta$ the area distribution of the thresholds. This is the integral of Law~\ref{law:closure} for a local law with a threshold, Eq.~\eqref{eq:aggthr} with the wetness as the shared variable. The aggregate is exact on that set and has no shape parameters beyond those of $F_\theta$. We take a logistic $F_\theta$ with centre $\theta_c$ and width $s_\theta$, and let a wetter root zone $s$ lower the centre, $F_\theta=\sg((\theta-\theta_c+\lambda s)/s_\theta)$ with $\lambda\ge0$. The groundwater-controlled connectivity of Section~\ref{sec:derive} later adds a shift by the groundwater storage. The connected fraction enters the release as a multiplier of its coefficient, $m=\ln A\in[\ln A_{\min},0]$, which is operation (d). Because $0<A\le1$, the multiplier lowers the conductance of the release and never reverses the flux (Law~\ref{law:dissip}), and $q_r$ still vanishes at $R=0$.

\paragraph{Step 3: reduction to one state.} Under the shared forcing and the same balance in every unit the deviations have no source. This is the case of Appendix~\ref{app:realise} in which the reference state is invariant and dissipation removes the differences at a rate at least proportional to their size. We measure the differences by their area-weighted variance $V$. With $f(\theta+y_i)-f(\theta)=f'(\tilde\theta_i)\,y_i$ and $f'\le-C_\theta/\tau_\theta$ on $[0,1]$,
\begin{equation}\label{eq:Vdecay}
\tfrac12C_\theta\dot V=\sum_ia_iy_i\big[f(\theta+y_i)-f(\theta)\big]-\sum_{i<j}a_ik_{ij}(y_i-y_j)^2\le-\big(C_\theta/\tau_\theta+\mu_1\big)V ,
\end{equation}
so $V$ decays at least at the rate $2(1/\tau_\theta+\mu_1/C_\theta)$ under any forcing sequence, and at $2/\tau_\theta$ even without exchange. On $y=0$ every unit sits at $\theta$, the aggregate of step 2 follows, and $\theta$ is the only state.

However, where the forcing has a spatial pattern or the units differ in their balance, the deviations have a source and the reduction rests on a spectral gap. On the manifold $y=0$ the Jacobian of the deviations is $-\lambda(\theta)\bm I-\bm L/C_\theta$. Its eigenvalues $-\lambda(\theta)-\mu_k/C_\theta$, $k\ge1$, lie below the eigenvalue $-\lambda(\theta)$ of the mean (step 4) by the gap $\mu_1/C_\theta$, the rate at which lateral exchange evens out a difference in wetness. The manifold is therefore normally attracting, the condition of the fast--slow form \eqref{eq:fastslow}, with $\zeta=\Omega_s/(\lambda+\mu_1/C_\theta)$ and $\Omega_s$ the larger of $\lambda$ and the rate at which the forcing changes. For a weak pattern, with the deviations relaxing within the day, the slow manifold lies approximately at deviations set by $\theta$ and the day's forcing, of the order of their source divided by $\lambda+\mu_1/C_\theta$. These deviations shift the thresholds and add terms of second order in the pattern to the mean balance, both functions of $\theta$ and the day's forcing, so a single state still suffices, the slow reduction \eqref{eq:slowred} with $q=1$. \CaRTp{} keeps the balance under the shared forcing, \eqref{eq:theta}, and a fixed threshold distribution, and leaves these effects to its fitted parameters as an empirical closure. The gap requires lateral exchange faster than the relaxation of $\theta$, which takes days to weeks (step 4).

\paragraph{Step 4: the memory of $\theta$.} Under the shared forcing the slow manifold is $y=0$, where $V=0$ and the mean wetness $\theta\in[0,1]$ obeys the balance \eqref{eq:theta}, the first part of \eqref{eq:meandev}. This single state carries the history of the forcing on which the connected fraction depends, the Markov embedding of Appendix~\ref{app:realise}. Let $\bar\theta$ be the equilibrium of the balance under the mean forcing $(\bar P_n,\bar E_n)$. To first order in the anomalies $\delta\theta$, $\delta P_n$ and $\delta E_n$,
\begin{equation}\label{eq:applin}
\dot{\delta\theta}=-\frac{\delta\theta}{\tau_{\rm eff}}+\frac{(1-\bar\theta^2)\,\delta P_n-\bar\theta(2-\bar\theta)\,\delta E_n}{C_\theta},\qquad
\frac1{\tau_{\rm eff}}=\frac1{\tau_\theta}+\frac{2\bar\theta\bar P_n+2(1-\bar\theta)\bar E_n}{C_\theta}.
\end{equation}
This is the state form of \eqref{eq:embed} with a single mode $\delta\theta$ of rate $\lambda=1/\tau_{\rm eff}$ and with the weighted net-input anomaly over $C_\theta$ as its drift. The anomaly $\delta\theta$ is therefore $\tau_{\rm eff}$ times an exponential moving average of the drift, with time constant $1/\lambda$, plus a decaying initial anomaly. With the learned parameters, at the equilibrium under the mean forcing, $\tau_{\rm eff}$ has a median of 4.5\,d on CAMELS-GB and 19\,d on CAMELS-US, with the central 80\,\% of the catchments between 2.7 and 55\,d and between 3.9 and 78\,d, respectively. Away from the mean the rate is $\lambda(\theta)$ of \eqref{eq:meandev}, which changes with $\theta$ and with the day's forcing. The state is still the single number $\theta$, with a kernel that depends on the path, as for the two reservoirs with $m(W)\ne1$ (Appendix~\ref{app:prop1}). In Appendix~\ref{app:realise} the rate is an eigenvalue of the fine-scale dynamics, whereas here it follows from the postulated balance, with $C_\theta$ and $\tau_\theta$ learned per catchment.

\paragraph{Step 5: daily update, nesting and validation.} The balance has exact one-day solutions, those of a Riccati equation on days with net input and a Bernoulli equation on days with net demand, checked against a fourth-order Runge--Kutta integration (Appendix~\ref{app:numerics}, Supporting Information S2). The wetness $\theta$ starts at $0.3$ and is spun up over four years. The limit $A_{\min}\to1$ sets $A\equiv1$ and recovers the release of \CaRT. The refinement was kept on the development splits of both continents (Section~\ref{sec:loop}). In the final model its group contributes 0.006 on CAMELS-GB, while on CAMELS-US it adds 0.0007 [$-0.0007$, 0.0013] once the other groups are present (Section~\ref{sec:contrib}).

\paragraph{Constructions of $\omega$ and $v$.} \emph{Near-surface wetness.} The near-surface wetness $\omega$ scales the infiltration capacities. A wetness shared by all points that scales every local capacity by the same factor keeps the capacities exponentially distributed, so the infiltration excess of Section~\ref{sec:derive} stays exact, and points with the same linear balance under the same input converge to one wetness by the argument of step 3. The model's $\omega$, an exponential moving average of the water leaving the surface layer with a time constant of 3 days, is the exact daily solution of that balance. But the scale function in $F=P^*e^{-g_F\omega/\bar\omega}$ is heuristic (Supporting Information S2).

\emph{Vegetation activity.} The vegetation activity $v$ is the active fraction of patches that switch between activity and dormancy at a common rate, depending on whether the air temperature exceeds their threshold. Each patch obeys a linear equation, so the area average relaxes exactly towards the threshold distribution at the day's temperature, and with a logistic distribution its daily solution is the exponential moving average of the model. Averaging the demand of active and dormant patches gives $E_{\rm dem}=\mathrm{PET}[1-\kappa(1-v)]$ (Supporting Information S2).

Both $\omega$ and $v$ are single modes of \eqref{eq:embed} with a constant rate, $\lambda=1/(3\,\mathrm d)$ for $\omega$ and $\lambda=1/\tau_v$ for $v$, which is the case $m\equiv1$ of Appendix~\ref{app:prop1}.

\section{Numerical implementation}\label{app:numerics}

The operators of a day are applied in sequence as in GR4J: snow, surface layer, infiltration excess, production and near-stream zone, depressions, groundwater, routing, quick store with exchange, and channel loss (Supporting Information S1.2). State- or forcing-dependent coefficients are evaluated once per day and held fixed (Section~\ref{sec:loss}), including the release multiplier and shape factor of the quick store, evaluated at the storage after the day's inflow, and the uptake rate of the groundwater store. Stores with a closed-form law use the exact one-day solution, while the surface layer, the filling of the depressions and the exchange use explicit steps from the storage at the start of the day. Integrating all processes simultaneously would give a slightly different model \citep{santos2018}. The energy bound of Law~\ref{law:dissip} is applied flux by flux. The evaporation of the surface layer, root zone, near-stream zone and depressions and the groundwater uptake draw on one demand in sequence. Sublimation and the melt term are computed from the potential evaporation of the day without being taken from that demand, so on snow-covered days the phase-change fluxes together can exceed it. The wetness $\theta$ and the near-stream zone use the closed-form daily solutions of \eqref{eq:theta} (Riccati on days with net input, Bernoulli on days with net demand, checked against a fourth-order Runge--Kutta integration to $10^{-14}$). The daily recursions are implemented as fused GPU kernels with adjoint recursions, verified against automatic differentiation of the unrolled loop (Supporting Information S1.7).

Training uses AdamW (learning rate $2\times10^{-3}$, one-cycle schedule, weight decay $10^{-4}$, gradient clipping at 1) for 5000 iterations for \CaRT, \CaRTp{} and the regionalised HBV on all three continents. The static network is trained with Gaussian noise of standard deviation 0.1 added to the standardised attributes. For spin-up, the first four years of the record (from December 1988 on CAMELS-GB and January 1980 on CAMELS-US and CAMELS-AUS) are run without gradient for the production store, near-stream zone, surface layer, depressions and wetness. The quick store uses the first two years of the record, also without gradient. The groundwater store starts at the steady state under the mean recharge of the training period, and the snow bands and unit hydrographs start empty (Supporting Information S1.6).

\paragraph{Benchmark configurations and forcing.} The EA-LSTM follows the configuration of \citet{kratzert2019hess}, with 256 units, dropout 0.4, the NSE$^*$ loss and its learning-rate schedule, except for input sequences of 365 days instead of 270. Its dynamic inputs are the daily precipitation, mean air temperature and potential evaporation that the conceptual models receive, in place of the five Maurer forcing variables of the original study, so that every model has the same information. Training runs for 30 epochs with batches of 256, each epoch a random draw of 1.75 million training samples (the size of an epoch in the original study), with an initial forget-gate bias of 5, and the epoch with the highest median NSE over the selection period is kept. The configuration of the stateful LSTM is based on our own hyperparameter tuning. It has 256 units, output dropout 0.3 and an initial forget-gate bias of 3, and it is trained with the NSE$^*$ loss by truncated back-propagation through time on 180-day chunks for groups of 151 catchments at a constant learning rate, for at most 300 epochs with early stopping on the selection period. dHBV is trained with Adam (learning rate $10^{-3}$) for 6000 iterations on batches of 100 two-year windows, the first year of which spins up the states. Its median NSE over the selection period is evaluated every 500 iterations, and the best checkpoint is kept. The LSTM of dHBV has 256 units and dropout 0.5. dHBV receives the static attributes without the band elevations, and its snow routine works on the catchment-mean temperature, as in the original. In evaluation its static parameters are the LSTM output on the last day of the training period. All three benchmarks are trained with Adam, starting from a learning rate of $10^{-3}$, and clip the gradient norm at 1.  \CaRT, \CaRTp{} and the regionalised HBV were trained with one fixed configuration throughout. We rebuilt the CAMELS-US forcing from Daymet version~4 rather than extending the version distributed with CAMELS because the two differ in their day convention. Version~4 corrects the time of observation of the precipitation gauges so that a daily value covers the calendar day \citep{thornton2021daymet4}, whereas in the CAMELS version most of a day's rain is recorded on the following day. A model trained on one version and run on the other loses about 0.2 in median NSE from this shift alone, with a loss of 0.006--0.009 once the shift is removed.

\section{Connection matrix of \CaRTp}\label{app:matrix}

The area-weighted storages are $\tilde N_k=a_kN_k$ (five snow bands), $U$ (surface layer), $\tilde S=(1-w)S$ (root zone), $\tilde Z=wZ$ (near-stream zone), $V$ (depressions), $H$ (water in transit in the unit hydrographs), $R$ (quick store) and $W$ (groundwater). The incidence matrix of \CaRTp{} is written transposed below, with one row per flux and one column per store, while the boundary node at the other end of a boundary flux is named in the last column and has no column of its own:
\begin{center}\small
\begin{tabular}{l*{8}{c}l}
\toprule
flux & $\tilde N_k$ & $U$ & $\tilde S$ & $\tilde Z$ & $V$ & $H$ & $R$ & $W$ & boundary \\
\midrule
snowfall & $+\bm I$ & & & & & & & & from atmosphere \\
sublimation $u_k$ & $-\bm I$ & & & & & & & & to atmosphere \\
rainfall & & $+1$ & & & & & & & from atmosphere \\
melt & $-\bm I$ & $+\bm 1^{\!\top}$ & & & & & & & \\
surface evaporation $e_u$ & & $-1$ & & & & & & & to atmosphere \\
direct runoff $i_e$, $d$, $d_z$ & & $-1$ & & & $+1$ & & & & \\
Pareto infiltration $p_s$ & & $-1$ & $+1$ & & & & & & \\
input to the near-stream zone & & $-1$ & & $+1$ & & & & & \\
root-zone evaporation $e_s$ & & & $-1$ & & & & & & to atmosphere \\
percolation $g$ & & & $-1$ & & & & & $+1$ & \\
near-stream evaporation $e_Z$ & & & & $-1$ & & & & & to atmosphere \\
near-stream drainage $q_z$ & & & & $-1$ & & & & & to outlet \\
depression evaporation $e_V$ & & & & & $-1$ & & & & to atmosphere \\
seepage & & & & & $-1$ & & & $+1$ & \\
spill & & & & & $-1$ & $+1$ & & & \\
UH1 output $h\!\to\!R$ & & & & & & $-1$ & $+1$ & & \\
UH2 output $h\!\to\!O$ & & & & & & $-1$ & & & to outlet \\
quick release $q_r$ & & & & & & & $-1$ & & to outlet \\
exchange $0.9j_{\rm ext}$ with the quick store & & & & & & & $\pm1$ & & exterior \\
exchange $0.1j_{\rm ext}$ on the direct path & & & & & & & & & exterior $\leftrightarrow$ outlet \\
baseflow $q_s$ & & & & & & & & $-1$ & to outlet \\
groundwater uptake $e_W$ & & & & & & & & $-1$ & to atmosphere \\
channel transmission loss & & & & & & & & & outlet $\to$ exterior \\
\bottomrule
\end{tabular}
\end{center}
Here $i_e$, $d$ and $d_z$ are infiltration excess and the saturation excess of the root and near-stream zones, all of which enter the depressions $V$. The exchange $j_{\rm ext}$ with the exterior (inflow $J$ and dissipative loss together) is applied with weight 0.9 to $R$ and 0.1 to the direct path to the outlet. The exchange on the direct path and the channel loss connect two boundary nodes and have no store entry. Merging $U$, $\tilde S$, $\tilde Z$ and $V$ maps every internal flux (row) onto a flux of \CaRT{} ($\bm\Pi'\bm B'=\bm B\bm K$). The near-stream drainage to the outlet, the evaporation of the depressions and the groundwater uptake are the boundary fluxes that \CaRT{} lacks. The exterior inflow and the channel loss use connections that \CaRT{} already has and change the laws of those fluxes. The wetness $\theta$, the near-surface wetness $\omega$ and the vegetation activity $v$ are conditional variables, which carry no water and have zero entries in $\bm B$. The groundwater head, a function of the store $W$, and the frost indicator, a function of the forcing, enter the fluxes in the same way. Wetness acts through the connected fraction $A$, which modulates both the quick release and the bypass to UH2, and the groundwater head also acts through $A$.

\clearpage
\setcounter{section}{0}\setcounter{equation}{0}\setcounter{table}{0}\setcounter{figure}{0}
\renewcommand{\thesection}{S\arabic{section}}\renewcommand{\theequation}{S\arabic{equation}}
\renewcommand{\thetable}{S\arabic{table}}\renewcommand{\thefigure}{S\arabic{figure}}
\renewcommand{\theHsection}{S\arabic{section}}\renewcommand{\theHequation}{S\arabic{equation}}
\renewcommand{\theHtable}{S\arabic{table}}\renewcommand{\theHfigure}{S\arabic{figure}}
\begin{center}{\Large\bfseries Supporting Information for: From hydrological laws to catchment models: A process-based model matches deep-learning accuracy on three continents\par}\end{center}
\bigskip
\noindent\textbf{Summary.} Section~\ref{sec:S1} gives the complete specification of \CaRT{} and \CaRTp{} including the numerical implementation, parameter ranges and initial values. It also documents initial states and spin-up, the static-attribute neural network and its training, and the water-balance bookkeeping. Section~\ref{sec:S2} derives each refinement of \CaRTp{} from its fine-scale assumption and states the nested limit and the origin of the derivation. References cited in this Supporting Information are listed in the reference list of the main text. %

\section{Complete specification of \CaRT{} and \CaRTp}\label{sec:S1}

The continuous laws of the main text are implemented as discrete daily operations. This section specifies the operations as they were trained and evaluated, including where they differ from the continuous laws (operator splitting, clamps, the bounded release multiplier).

\subsection{Forcing, states and conventions}\label{sec:S1conv}

\paragraph{Units and time step.} Depths are in mm, fluxes in mm\,d$^{-1}$, temperatures in \si{\celsius}, and the time step is one day ($\Delta t=1$). Fluxes and storages are catchment means except in the root zone, the near-stream zone and the snow bands. Root-zone and near-stream states and fluxes are per unit area of their zone (areas $1-w$ and $w$, respectively) and are weighted by these fractions when they leave the store. Snow bands are weighted by their area fractions $a_k$. We write $(x)_+=\max(x,0)$ and $\sg(x)=1/(1+e^{-x})$.

\paragraph{Values within a day.} A store written without a mark is its value at the start of the day. A subscript $m$ marks its value after the inflow of the day, primes mark further values within the day, and a superscript $+$ marks the value at the end of the day, which is the start value of the next day. Where a store is integrated exactly over the day, $z_0$ and $z_1$ are its scaled values at the start and the end of the day. A subscript $t$ marks the value of a series on day $t$, that is, the forcing of the day, a coefficient computed for the day or the end-of-day value of a state or conditional variable ($s_t=S^+/C$, $W_t=W^+$, $\theta_t$, $v_t$). The wetness index $\omega_t$ enters day $t$ and is computed from the input up to day $t-1$. How each daily update relates to the continuous law is stated at the end of Sections~\ref{sec:S1cart} and \ref{sec:S1cartp}.

\paragraph{Forcing.} Per catchment and day the forcing is precipitation $P$, daily mean air temperature $T$ and potential evapotranspiration $E$, all three distributed with the dataset. On CAMELS-GB $E$ is the dataset PET with interception correction. On CAMELS-US $T$ is the mean of the Daymet daily maximum and minimum, and $E$ is the Priestley--Taylor PET from the Daymet forcing. On CAMELS-AUS $P$ and $T$ are those of the Australian Gridded Climate Data, with $T$ the mean of the daily maximum and minimum, and $E$ is the Morton wet-environment PET of SILO. The record runs from 3 December 1988 (CAMELS-GB) or 1 January 1980 (CAMELS-US and CAMELS-AUS) to 31 December 2019. Each simulation starts on the first day of the record and runs continuously to its end. The days before the training period are simulated but do not enter the loss (Section~\ref{sec:S1spin}).

\paragraph{Elevation bands.} Both models use $K=5$ elevation bands with elevations $h_k$ and area fractions $a_k$. On CAMELS-GB, $a_k=1/5$ and $h_k$ is the elevation at the 10, 30, 50, 70 and 90\,\% points of the hypsometric curve, which is interpolated linearly through the minimum, 10\,\%, 50\,\%, 90\,\% and maximum elevations of the topographic attributes. On CAMELS-US, the 100\,m bands of the dataset with positive area are sorted by elevation. Band $i$ is assigned to group $\min\{\lfloor(\text{cumulative area up to and including }i-10^{-9})/0.2\rfloor,4\}$ (a band that ends exactly at 20, 40, 60 or 80\,\% of the area stays in the lower group). The group elevation $h_k$ is the area-weighted mean of its bands and $a_k$ is its area, which is renormalised so that $\sum_ka_k=1$, with $a_k=0$ for empty groups. Because a single 100\,m band often covers more than a fifth of a basin, the groups are generally not equal-area and some are empty. Of the 671 CAMELS-US basins, 446 have at least one empty group, and 93 have a single non-empty group (44 of them a single 100\,m band). Among basins with two or more groups, the median ratio of the largest to the smallest group area is 3.1. CAMELS-AUS distributes no band elevations, so its five bands have $a_k=1/5$ and $h_k$ at the 10, 30, 50, 70 and 90\,\% quantiles of a normal hypsometry truncated at the lowest and highest elevation of the catchment. The distribution has a standard deviation of 0.17 times the elevation range (the median ratio of the CAMELS-US bands) and a centre chosen so that the mean of the truncated distribution equals the mean elevation. The elevation offsets $\Delta h_k=(h_k-\sum_ja_jh_j)/1000$ are in km.

\paragraph{States.} Table~\ref{tab:states} lists the stores and the conditional variables, which carry no water.

\begin{table}[h]\centering\small
\caption{States of \CaRT{} and \CaRTp. Stores carry water and enter the water balance, whereas conditional variables carry none. The fourteen states of \CaRTp{} are the eleven stores other than $H$ and the three conditional variables. $H$ denotes the water in transit and is listed for the water balance.}\label{tab:states}
\begin{tabular}{@{}llll@{}}\toprule
Symbol & Meaning & Model & Unit\\\midrule
$N_k$, $k=1..5$ & snow water equivalent of band $k$ & both & mm (per band area)\\
$U$ & surface layer & \CaRTp & mm\\
$S$ & production store (root zone in \CaRTp) & both & mm (per area $1-w$)\\
$Z$ & near-stream saturated zone & \CaRTp & mm (per area $w$)\\
$V$ & depression storage & \CaRTp & mm\\
$H$ & water in transit in the unit hydrographs & both & mm\\
$R$ & quick store & both & mm\\
$W$ & groundwater store & both & mm\\\midrule
$v$ & vegetation activity & \CaRTp & --\\
$\omega$ & near-surface wetness index & \CaRTp & mm\,d$^{-1}$\\
$\theta=\Theta/C_\theta$ & mean hillslope wetness (connectivity state) & \CaRTp & --\\\bottomrule
\end{tabular}
\end{table}

\paragraph{Parameters.} Each per-catchment parameter $p$ with range $(p_{\rm lo},p_{\rm hi})$ is an output of the static-attribute network (Section~\ref{sec:S1net}),
\begin{equation}\label{eq:map}
p=p_{\rm lo}+(p_{\rm hi}-p_{\rm lo})\,\sg\big(\bm w_p^{\!\top}\bm h+b_p\big),
\end{equation}
where $\bm h$ is the last hidden layer of the network. Parameters marked ``log'' in Tables~\ref{tab:parCaRT}--\ref{tab:parCaRTp} are mapped in this way in $\ln p$. At initialisation $\bm w_p=\bm 0$ and $b_p=\mathrm{logit}[(p_0-p_{\rm lo})/(p_{\rm hi}-p_{\rm lo})]$, with $\ln p$ in place of $p$ for the log-mapped parameters, so that all catchments start at the initial value $p_0$. Because of the logistic function the range is open, and a bound is approached but never attained. The global constants (Table~\ref{tab:glob}) are unbounded scalars, with scale constants learned as logarithms.

\subsection{Order of operations within a day}\label{sec:S1order}

Each step is written as (inputs) $\to$ (outputs).

\paragraph{\CaRT.} (1) Snow bands: $(P,T,E)\to(P_L,u)$. (2) Production store: $(P_L,E)\to(D,G,e_s+I)$. (3) Unit hydrographs: $0.9D\to x_9$ (UH1) and $0.1D\to x_1$ (UH2). (4) Quick store with exchange: $(x_9,x_1)\to(Q_r,Q_d)$. (5) Groundwater store: $G\to q_s$. (6) Outlet: $(Q_r,Q_d,q_s)\to Q$.

\paragraph{\CaRTp.} (1) Vegetation: $T\to v_t$ and $(E,v_t)\to E_{\rm dem}$. (2) Snow bands: $(P,T,E)\to(P_L,u)$. (3) Surface layer: $(P_L,E_{\rm dem})\to(x,\ein,e_u)$. (4) Infiltration excess: $(x,\omega_t)\to(i_e,x')$. (5) Root zone (area $1-w$): $(x',\ein)\to(d_s,g,e_s,s_t)$. (6) Near-stream zone (area $w$), in parallel with (5): $(x',\ein)\to(d_z,q_z,e_z)$. (7) Catchment fluxes of the soil: $(d_s,d_z,i_e,e_s,e_z)\to(D,{\rm AET}_{sz})$. (8) Depressions: $(D,E_d)\to(D',e_d,g_d)$. (9) Groundwater: $(G,E_r)\to(e_W,q_s,W_t)$. (10) Hillslope wetness and connected fraction: $(x',\ein)\to\theta_t$ and $(\theta_t,s_t,W_t)\to A_t$. (11) Bypass and unit hydrographs: $(D',A_t)\to(x_9,x_1)$. (12) Quick store: $(x_9,x_1,J,A_t,T)\to(Q_r,Q_d)$. (13) Outlet and channel loss: $(Q_r,Q_d,q_s,q_z)\to Q$.

\paragraph{Daily scheme.} Stores with a closed-form law use the exact one-day solution, while the surface layer, the filling of the depressions and the exchange use explicit steps from the storage at the start of the day. Any coefficient that depends on the state or the forcing is evaluated once and held fixed over the day. Explicit steps evaluate a flux from the storage at the start of the day and hold it over the day, so on some days the flux can exceed the water present. For example, the share $0.1j$ of the exchange loss that the direct flow carries exceeds that flow on days without direct runoff. Such a flux is limited to the water available by $(\cdot)_+$ or $\min\{\cdot\}$. The relative storages $s$ and $s_d$ and the activity $v$ are also limited to $[0,1]$, which rounding could otherwise move slightly outside. These limits are the clamps referred to below, and the water balance uses the fluxes after them (Section~\ref{sec:S1bal}).

\subsection{\CaRT: flux laws and daily updates}\label{sec:S1cart}

\paragraph{Snow bands.} Band $k$ has temperature $T_k=T-\Gamma\,\Delta h_k$ and indicator $c_k=\mathbf 1[T_k<T_T]$. Each day, for each band,
\begin{equation}\label{eq:snow}
\begin{aligned}
N_{k,m}&= N_k+c_k\,\mathrm{SFCF}\,P, & u_k&=\min(\kappa_sE,\,N_{k,m}), & N_k'&= N_{k,m}-u_k,\\
M_k&=(1-c_k)\min\{\cf(T_k-T_T)+r_pE,\ N_k'\}, & N_k^+&= N_k'-M_k, &&
\end{aligned}
\end{equation}
and the catchment receives the liquid input $P_L=\sum_ka_k[(1-c_k)P+M_k]$ and loses the sublimation $u=\sum_ka_ku_k$. Here $\cf$ is the degree-day factor CFMAX. Rain is not corrected by SFCF, and $E$ is the PET of the forcing. In \CaRT{} $r_p=0$, while the melt term $r_pE$ is a refinement of \CaRTp{} (Section~\ref{sec:S1cartp}).

\paragraph{Production store.} With input $x=P_L$ and demand $e=E$: $P_n=(x-e)_+$, $E_n=(e-x)_+$, interception $I=\min(x,e)$, $s=S/C$ at the start of the day, $t_P=\tanh(P_n/C)$, $t_E=\tanh(E_n/C)$,
\begin{equation}\label{eq:prod}
p_s=\frac{C(1-s^2)\,t_P}{1+s\,t_P},\quad e_s=\frac{S(2-s)\,t_E}{1+(1-s)\,t_E},\quad S'=S-e_s+p_s,\quad g=S'\Big[1-\big(1+\kappa_q(S'/C)^4\big)^{-1/4}\Big],
\end{equation}
$S^+=S'-g$, direct runoff $D=P_n-p_s$, evaporation $e_s+I$, percolation $G=g$. Here $\kappa_q=4K_p/C=(4/9)^4\,r_K$ with $r_K=K_p/K_{p,\rm GR4J}$ the learned ratio and $K_{p,\rm GR4J}=C(4/9)^4/4$.

\paragraph{Unit hydrographs.} The first cumulative hydrograph is $\mathrm{SH}_1(t)=(t/X_4)^{2.5}$ for $0<t<X_4$ (0 before, 1 after), and the second is $\mathrm{SH}_2(t)=\frac12(t/X_4)^{2.5}$ for $0<t\le X_4$, $1-\frac12(2-t/X_4)^{2.5}$ for $X_4<t<2X_4$ (0 before, 1 after). The ordinates are $o^{(i)}_j=\mathrm{SH}_i(j)-\mathrm{SH}_i(j-1)$, $j=1..6$ for UH1 and $j=1..13$ for UH2, and the convolutions are
\begin{equation}\label{eq:uh}
x_9(t)=\sum_{j=1}^{6}o^{(1)}_j\,0.9D(t-j+1),\qquad x_1(t)=\sum_{j=1}^{13}o^{(2)}_j\,0.1D(t-j+1).
\end{equation}
Here $D=0$ before the first day, and the first ordinate acts on the day of input.

\paragraph{Quick store with exchange.} With $R$ at the start of the day,
\begin{equation}\label{eq:quickCaRT}
j=X_2\Big(\frac{R_+}{X_3}\Big)^{3.5},\quad R_m=\big(R+x_9+0.9j\big)_+,\quad Q_d=\big(x_1+0.1j\big)_+,\quad R^+=R_m\big[1+(R_m/X_3)^4\big]^{-1/4}.
\end{equation}
Here $Q_r=R_m-R^+$. The exchange is evaluated at the start-of-day storage and applied once, with 0.9 going to the store and 0.1 to the direct flow. The learned exchange parameter is $X_2\in(-20,0)$\,mm\,d$^{-1}$, so that the exchange is a loss (Law~2). The realised exchange $(R_m-R-x_9)+(Q_d-x_1)$, which differs from $j$ only when a clamp is active, enters the water balance.

\paragraph{Groundwater store.} Let $\bar g=\max(\overline{G},10^{-6})$ be the mean percolation over the training period of the same simulation, with the residence time $\tau$ at that recharge as the learned parameter:
\begin{equation}\label{eq:gw}
\begin{aligned}&a_g=\frac{1}{\bar g\,\tau^2},\qquad W_{\rm ss}=\sqrt{\bar g/a_g}=\bar g\,\tau,\\ &W_m=W+G,\qquad W^+=\frac{W_m}{1+a_gW_m},\qquad q_s=W_m-W^+.\end{aligned}
\end{equation}
\paragraph{Outlet.} $Q=Q_r+Q_d+q_s$.

\paragraph{Relation to the continuous laws.} Each release is the exact one-day solution of its continuous law, with the storage after the day's inflow as the initial value. The percolation $g$ solves $\dot S=-K_p(S/C)^5$ starting at $S'$. The quick release $Q_r$ solves $\dot R=-a_fR^5$ starting at $R_m$, where $a_f=1/(4X_3^4)$ is not a separate parameter, and the groundwater release $q_s$ solves $\dot W=-a_gW^2$ starting at $W_m$. The storage $W_{\rm ss}=\bar g\tau$ is the equilibrium of the continuous law $\dot W=\bar g-a_gW^2$ under the mean recharge $\bar g$.

\subsection{\CaRTp: flux laws and daily updates}\label{sec:S1cartp}

\CaRTp{} keeps the operations of Section~\ref{sec:S1cart} except where stated.

\paragraph{(1) Vegetation activity and demand.} With $\sigma_t=\sg((T_t-T_v)/s_v)$ and $\alpha_v=e^{-1/\tau_v}$,
\begin{equation}\label{eq:veg}
v_t=\bar\sigma+\sum_{j=0}^{2047}(1-\alpha_v)\,\alpha_v^{\,j}\,(\sigma_{t-j}-\bar\sigma),\qquad E_{{\rm dem},t}=E_t\,[1-\kappa(1-v_t)],
\end{equation}
with $\sigma_{t-j}=\bar\sigma$ before the first day, where $\bar\sigma$ is the mean of $\sigma$ over the first 365 days of the record, and with $v_t$ clamped to $[0,1]$. $E_{\rm dem}$ is the demand of all evaporative sinks below the snow bands, while sublimation and the melt term use the unmodified $E$.

\paragraph{(2) Snow bands with melt energy term.} Eq.~\eqref{eq:snow} applies with $r_p>0$, so that on days with $T_k\ge T_T$ the melt is $\min\{\cf(T_k-T_T)+r_pE,\,N_k'\}$, with a common $E$ across all bands.

\paragraph{(3) Surface layer.} With $U$ at the start of the day and $k_u=1-e^{-1/\tau_u}$:
\begin{equation}\label{eq:top}
\begin{aligned}&U_m=U+P_L,\qquad e_u=\min(E_{\rm dem},U_m),\qquad U'=U_m-e_u,\\ &o_u=(U'-C_u)_+,\qquad U''=U'-o_u,\qquad d_u=k_uU''.\end{aligned}
\end{equation}
After the update to $U^+=U''-d_u$, the layer passes the input $x=d_u+o_u$ and the remaining demand $\ein=E_{\rm dem}-e_u$ to the stores below, with the overflow passing through the infiltration excess to the root zone.

\paragraph{(4) Infiltration excess.} The near-surface wetness index is an exponential moving average, with time constant 3\,d, of the layer output up to the previous day, truncated at 36 days and started at $\bar\omega$:
\begin{equation}\label{eq:omega}
\omega_t=\alpha^t\bar\omega+\sum_{j=1}^{\min(t,36)}(1-\alpha)\,\alpha^{j-1}x_{t-j},\qquad\alpha=e^{-1/3},
\end{equation}
$t=0$ on the first day of the record, $\bar\omega=\max(\bar P,0.1\,\text{mm\,d}^{-1})$ with $\bar P$ the mean precipitation $P$ of the training period. The mean infiltration capacity and the excess are
\begin{equation}\label{eq:ie}
F_t=\max\{P^*\exp(-g_F\,\omega_t/\bar\omega),\,F_{\min}\},\qquad i_e=x-F_t\big(1-e^{-x/F_t}\big),\qquad x'= x-i_e .
\end{equation}
$i_e$ joins the direct runoff of the whole catchment, and $x'$ is the input of the stores below. On CAMELS-AUS the capacity is bounded below by $F_{\min}=10^{-4}$\,mm\,d$^{-1}$, and on the other datasets $F_{\min}=0$.

\paragraph{(5) Root zone.} The production store of Eq.~\eqref{eq:prod} uses the Pareto partition, with $P_n=x'$, $E_n=\ein$, $s=\min(\max(S/C,0),1)$, $\delta=\max(1-s,10^{-7})$, $c_{\max}=(b+1)C$:
\begin{equation}\label{Seq:pareto}
p_s=\min\Big\{\Big[C\Big(\delta-\big(\delta^{1/(b+1)}-P_n/c_{\max}\big)_+^{\,b+1}\Big)\Big]_+,\ P_n\Big\},\qquad e_s=\min\Big\{\frac{C(1-\delta^2)\,t_E}{1+\delta\,t_E},\ S\Big\}.
\end{equation}
Here $t_E=\tanh(E_n/C)$, and $e_s$ is the GR4J evaporation of Eq.~\eqref{eq:prod}. Then $S'=S-e_s+p_s$, and $g$ and $S^+$ follow Eq.~\eqref{eq:prod}, with the saturation excess $d_s=P_n-p_s$ and the end-of-day wetness $s_t=\min(\max(S^+/C,0),1)$, which enters the connected fraction.

\paragraph{(6) Near-stream zone.} Per unit area of the zone, $\dot Z=P_n(1-z^2)-E_nz(2-z)-Z/\tau_z$, where $z=Z/C_z$. This equation is driven by the same $(P_n,E_n)=(x',\ein)$ as the root zone and integrated exactly over the day for fixed forcing. With $z_0=Z/C_z$, $a=P_n/C_z$, $e=E_n/C_z$, $\beta=1/\tau_z$:

\emph{Day without demand} ($E_n=0$; Riccati): $\Delta=\sqrt{\beta^2+4a^2}$, $z_+=2a/(\beta+\Delta)$, $\rho_0=2a(z_0-z_+)/(2az_0+\beta+\Delta)$,
\begin{equation}\label{eq:ric}
z_1=z_++\frac{(z_0-z_+)(1-\rho_0)\,e^{-\Delta}}{1-\rho_0e^{-\Delta}},\qquad \int_0^1z\,\dd t=z_++\frac{(z_0-z_+)(1-e^{-\Delta})}{\Delta}\,\psi\Big(\frac{\rho_0(1-e^{-\Delta})}{1-\rho_0}\Big),
\end{equation}
\emph{Day with demand} ($E_n>0$, $P_n=0$; Bernoulli): $\alpha_z=2e+\beta$, $\xi=z_0(e/\alpha_z)(1-e^{-\alpha_z})$,
\begin{equation}\label{eq:bern}
z_1=\frac{z_0\,e^{-\alpha_z}}{1-\xi},\qquad \int_0^1z\,\dd t=z_0\,\frac{1-e^{-\alpha_z}}{\alpha_z}\,\varphi(\xi),
\end{equation}
with $\psi(y)=\ln(1+y)/y$, $\varphi(\xi)=-\ln(1-\xi)/\xi$. The drainage is $q_z=C_z\beta\int_0^1z\,\dd t$, and on days without demand the saturation excess is $d_z=P_n-C_z(z_1-z_0)-q_z$ and $e_z=0$. On days with demand $d_z=0$ and $e_z=C_z(z_0-z_1)-q_z$, with $Z^+=C_zz_1$ after either branch. ($1-e^{-y}$, $\psi$ and $\varphi$ are evaluated by their Taylor series for arguments below $10^{-3}$, with $\psi(0)=\varphi(0)=1$, and the degenerate cases of the two solutions are their continuous limits.)

\paragraph{(7) Catchment fluxes of the soil.} In \CaRTp{} the direct runoff of all sources is $D=(1-w)\,d_s+w\,d_z+i_e$, the percolation is $(1-w)\,g$, the evaporation is ${\rm AET}_{sz}=(1-w)\,e_s+w\,e_z$, and the zone drainage $w\,q_z$ goes to the outlet.

\paragraph{(8) Depression storage.} The direct runoff $D$ of all sources fills depressions with Pareto-distributed capacities (mean $C_d$, shape $b_d$, $c_{d,\max}=(b_d+1)C_d$), which are subject to the residual demand $E_d=(\ein-{\rm AET}_{sz})_+$. Filling and evaporation are both computed from the start-of-day storage $V$, with $s_d=\min(\max(V/C_d,0),1)$, $\delta_d=\max(1-s_d,10^{-7})$, $t_d=\tanh(E_d/C_d)$:
\begin{equation}\label{eq:dep}
\begin{aligned}
p_d&=\min\Big\{\Big[C_d\Big(\delta_d-\big(\delta_d^{1/(b_d+1)}-D/c_{d,\max}\big)_+^{\,b_d+1}\Big)\Big]_+,\ D\Big\},\qquad e_d=\min\Big\{\frac{C_d(1-\delta_d^2)\,t_d}{1+\delta_d\,t_d},\ V\Big\},\\
V'&=V-e_d+p_d,\qquad g_d=V'\Big[1-\big(1+k_d(V'/C_d)^4\big)^{-1/4}\Big],\qquad V^+=V'-g_d,
\end{aligned}
\end{equation}
giving the spill $D'=D-p_d$ to the unit hydrographs, evaporation $e_d$ and seepage $g_d$ to groundwater.

\paragraph{(9) Groundwater with uptake.} In \CaRTp{} the recharge is $G=(1-w)\,g+g_d$, with $\bar g$, $a_g$ and $W_{\rm ss}=\bar g\tau$ as in Eq.~\eqref{eq:gw} with this $G$. The residual demand is $E_r=(\ein-{\rm AET}_{sz}-e_d)_+$. Each day the inflow is applied first, then the uptake at a rate held fixed over the day, and then the exact quadratic release:
\begin{equation}\label{eq:uptake}
\begin{aligned}&W_m=W+G,\qquad r=\frac{\lambda_EE_r}{W_m+W_{\rm ss}},\qquad e_W=W_m\big(1-e^{-r}\big),\\ &W'=W_m-e_W,\qquad W^+=\frac{W'}{1+a_gW'},\qquad q_s=W'-W^+ .\end{aligned}
\end{equation}
The value $W_t=W^+$ is the end-of-day storage used in step (10).

\paragraph{(10) Hillslope wetness and connected fraction.} The mean wetness of the hillslope units (Appendix~C of the main text), written $\Theta=C_\theta\theta$, follows the zone equations \eqref{eq:ric}--\eqref{eq:bern} with $(C_\theta,\tau_\theta)$ in place of $(C_z,\tau_z)$. It uses the forcing $(P_n,E_n)=(x',\ein)$ shared with the zone, and carries no water. With the end-of-day $\theta_t$, $s_t$ and $W_t$,
\begin{equation}\label{eq:A}
A_t=A_{\min}+(1-A_{\min})\,\sg\!\Big(\frac{\theta_t-\theta_c+\lambda s_t+\lambda_W\,(W_t/W_{\rm ss}-1)}{s_\theta}\Big).
\end{equation}

\paragraph{(11) Bypass and unit hydrographs.} The bypass fraction $\gamma A_t$ is applied to the whole spill, including the runoff of the near-stream zone and the infiltration excess (Section~\ref{sec:S2bypass}). Then $D_{\rm by}=\gamma A_tD'$, $x_9=\text{UH1}*(D'-D_{\rm by})$, $x_1=\text{UH2}*D_{\rm by}$ (Eq.~\eqref{eq:uh} with $0.9D$, $0.1D$ replaced by $D'-D_{\rm by}$, $D_{\rm by}$).

\paragraph{(12) Quick store.} The log-multiplier of the release coefficient is
\begin{equation}\label{eq:m}
\begin{aligned}&m_t=m_{{\rm conn},t}+m_{\rm visc}(T_t)-a_F\,\sg\!\Big(\frac{T_F-T_t}{s_F}\Big),\\ &m_{\rm visc}(T)=-\ln(10)\;247.8\Big[\frac{1}{T_K-140}-\frac{1}{293.15-140}\Big],\end{aligned}
\end{equation}
$T_K=\max(T,0)+273.15$, with $T$ the catchment-mean air temperature, and the connectivity term is $m_{{\rm conn},t}=\ln A_t$. With $R$ at the start of the day, the exterior inflow $J$ and the dissipative loss $j\le0$ are applied once and split 0.9/0.1 like the direct runoff:
\begin{equation}\label{Seq:quick}
\begin{aligned}
j&=X_2\Big(\frac{R_+}{X_3}\Big)^{3.5},\qquad R_m=\big(R+x_9+0.9(J+j)\big)_+,\qquad Q_d=\big(x_1+0.1(J+j)\big)_+,\\
f_t&=\ln\frac{\max(R_m,10^{-3})}{X_3},\qquad \cA_t=a_f\exp\Big\{6\tanh\Big[\frac{m_t+k\,f_t}{6}\Big]\Big\},\qquad R^+=R_m\big(1+4\cA_tR_m^4\big)^{-1/4},
\end{aligned}
\end{equation}
$Q_r=R_m-R^+$, $a_f=1/(4X_3^4)$, $X_2\in(-20,0)$. The coefficient $\cA_t$ is written $A_t$ in Figure~2 of the main text, where $A$ without subscript is the connected fraction. It is evaluated once, at the storage $R_m$ after the inflow, and held fixed. Because of the bound $6\tanh(\cdot/6)$, the local elasticity of the coefficient is $\dd\ln\cA/\dd\ln R_m=k\,\mathrm{sech}^2[(m_t+kf_t)/6]$ above the floor $R_m=10^{-3}$\,mm of $f_t$ and zero below it, and it tends to $k$ when the argument is small. The realised exchange $(R_m-R-x_9-0.9J)+(Q_d-x_1-0.1J)$ and the inflow $J$ enter the water balance.

\paragraph{(13) Outlet and channel loss.} The outlet receives $Q_0=Q_r+Q_d+q_s+w\,q_z$, the transmission loss is $\ell=L_c\big(1-e^{-Q_0/L_c}\big)$, and the discharge is $Q=Q_0-\ell$.

\paragraph{Relation to the continuous laws.} Up to the truncation of the kernel at 2048 days, $v_t$ is the recursion $v_t=\alpha_vv_{t-1}+(1-\alpha_v)\sigma_t$ started at $\bar\sigma$, which is the exact one-day solution of $\dot v=(\sigma_t-v)/\tau_v$ for $\sigma$ fixed over the day. The seepage $g_d$ is the exact one-day solution of $\dot V=-(k_dC_d/4)(V/C_d)^5$ from $V'$. With $\cA_t$ held fixed, $R^+$ is the exact one-day solution of $\dot R=-\cA_tR^5$ starting from $R_m$, and with $\cA_t=a_f$ ($m_t=0$, $k=0$) this is the release of \CaRT.

\subsection{Parameters}\label{sec:S1par}

\CaRT{} has 11 per-catchment parameters (Table~\ref{tab:parCaRT}), while the network of \CaRTp{} has 35 per-catchment outputs, which are the parameters of Table~\ref{tab:parCaRTp}, and the model also has seven global constants (Table~\ref{tab:glob}). The lower bound $A_{\min}\ge0.5$ is imposed because with a smaller $A_{\min}$ the connection acted as a steep switch on $R^5$ and training became unstable.

\begin{table}[h]\centering\small
\caption{Per-catchment parameters of \CaRT. Range: open interval of Eq.~\eqref{eq:map}; map: linear in $p$ (lin) or in $\ln p$ (log); $p_0$: initial value.}\label{tab:parCaRT}
\begin{tabular}{@{}lp{7.6cm}llr@{}}\toprule
Symbol & Meaning (unit) & Range & Map & $p_0$\\\midrule
$C$ & production-store capacity (mm) & 25--1600 & log & 225\\
$r_K$ & percolation scale relative to GR4J, $K_p=r_KC(4/9)^4/4$ & 1/30--30 & log & 1\\
$X_3$ & quick-store scale (mm); $a_f=1/(4X_3^4)$ & 5--1000 & log & 33\\
$X_4$ & unit-hydrograph time base (d) & 0.5--6 & lin & 1.26\\
$\tau$ & groundwater residence time at $\bar g$ (d); $a_g=1/(\bar g\tau^2)$ & 3--1460 & log & 78\\
$X_2$ & exchange (loss) coefficient (mm\,d$^{-1}$); $j=X_2(R/X_3)^{3.5}$ & $-20$--0 & lin & $-0.35$\\
$T_T$ & rain--snow threshold temperature (\si{\celsius}) & $-2.5$--2.5 & lin & 0\\
$\cf$ & degree-day factor CFMAX (mm\,\si{\celsius}$^{-1}$\,d$^{-1}$) & 0.5--10 & log & 3\\
SFCF & snowfall correction factor & 0.5--1.5 & lin & 1\\
$\Gamma$ & temperature lapse rate across the bands (\si{\celsius}\,km$^{-1}$) & 3--9 & log & 6.5\\
$\kappa_s$ & sublimation as a fraction of PET & $10^{-4}$--1 & log & $10^{-3}$\\\bottomrule
\end{tabular}
\end{table}

\begin{table}[p]\centering\footnotesize
\caption{Per-catchment parameters of \CaRTp{} (35), with the physical bounds of the main text ($k,a_F,\lambda,\lambda_W\ge0$). Columns are as in Table~\ref{tab:parCaRT}, with the last column listing the parameter value at which the refinement vanishes (nested limit). The labels are ``open bound'' when the limit is the bound of the open interval and ``outside'' when it lies beyond the range.}\label{tab:parCaRTp}
\begin{tabular}{@{}l>{\raggedright\arraybackslash}p{6.1cm}llr>{\raggedright\arraybackslash}p{3.3cm}@{}}\toprule
Symbol & Meaning (unit) & Range & Map & $p_0$ & Nested limit\\\midrule
\multicolumn{6}{@{}l}{\emph{Rain--runoff parameters of \CaRT}}\\
$C$ & root-zone capacity (mm) & 25--1600 & log & 225 & --\\
$r_K$ & percolation scale relative to GR4J & 1/30--30 & log & 1 & --\\
$X_3$ & quick-store scale (mm) & 5--1000 & log & 33 & --\\
$X_4$ & unit-hydrograph time base (d) & 0.5--6 & lin & 1.26 & --\\
$\tau$ & groundwater residence time at $\bar g$ (d) & 3--1460 & log & 78 & --\\
$X_2$ & dissipative exchange coefficient (mm\,d$^{-1}$) & $-20$--0 & lin & $-0.35$ & --\\
\multicolumn{6}{@{}l}{\emph{Snow}}\\
$T_T$ & threshold temperature (\si{\celsius}) & $-2.5$--2.5 & lin & 0 & --\\
$\cf$ & degree-day factor (mm\,\si{\celsius}$^{-1}$\,d$^{-1}$) & 0.5--10 & log & 3 & --\\
SFCF & snowfall correction & 0.5--1.5 & lin & 1 & --\\
$\Gamma$ & lapse rate (\si{\celsius}\,km$^{-1}$) & 3--9 & log & 6.5 & --\\
$\kappa_s$ & sublimation fraction of PET & $10^{-4}$--1 & log & $10^{-3}$ & --\\
$r_p$ & melt per unit PET (mm\,mm$^{-1}$) & $10^{-3}$--10 & log & 0.01 & $r_p=0$ (outside)\\
\multicolumn{6}{@{}l}{\emph{Quick release: connected fraction, shape, frost}}\\
$A_{\min}$ & permanently connected fraction & 0.5--0.999 & lin & 0.75 & $A_{\min}=1$ (outside)\\
$C_\theta$ & capacity of the wetness state (mm) & 5--500 & log & 50 & (with $A_{\min}=1$)\\
$\tau_\theta$ & loss time of the wetness state (d) & 2--200 & log & 20 & (with $A_{\min}=1$)\\
$\lambda$ & threshold shift with root-zone wetness & 0--3 & lin & 0.05 & (with $A_{\min}=1$)\\
$\lambda_W$ & threshold shift with groundwater head & 0--3 & lin & 0.05 & $\lambda_W=0$ (open bound)\\
$k$ & release-shape exponent & 0--3 & lin & 0.05 & $k=0$ (open bound)\\
$a_F$ & frost step & 0--6 & lin & 1 & $a_F=0$ (open bound)\\
\multicolumn{6}{@{}l}{\emph{Near-stream zone, demand, uptake}}\\
$w$ & area fraction of the zone & 0--0.5 & lin & 0.05 & $w=0$ (open bound)\\
$C_z$ & zone capacity (mm) & 5--300 & log & 15 & (with $w=0$)\\
$\tau_z$ & zone drainage time (d) & 2--120 & log & 4 & (with $w=0$)\\
$\kappa$ & demand suppression by inactive vegetation & 0--1 & lin & 0.1 & $\kappa=0$ (open bound)\\
$\lambda_E$ & share of residual demand met from groundwater & 0--1 & lin & 0.05 & $\lambda_E=0$ (open bound)\\
\multicolumn{6}{@{}l}{\emph{Surface layer, partition, infiltration excess}}\\
$C_u$ & surface-layer capacity (mm) & 1--300 & log & 20 & (any, with $\tau_u\to0$)\\
$\tau_u$ & surface-layer drainage time (d) & 0.05--60 & log & 10 & $\tau_u\to0$ (outside)\\
$b$ & Pareto shape of root-zone capacities & 0.03--8 & log & 0.7 & none\\
$P^*$ & reference infiltration capacity (mm\,d$^{-1}$) & 2--3000 & log & 1000 & $P^*\to\infty$ (outside)\\
$g_F$ & wetness sensitivity of the capacity & 0--4 & lin & 0.3 & (with $P^*\to\infty$)\\
\multicolumn{6}{@{}l}{\emph{Routing, depressions, exchange, channel}}\\
$\gamma$ & bypass probability of connected units & 0--1 & lin & 0.13 & $\gamma=0.1$ with $A\equiv1$\\
$C_d$ & mean depression capacity (mm) & 0.02--300 & log & 0.2 & $C_d\to0$ (outside)\\
$b_d$ & Pareto shape of depression capacities & 0.05--8 & log & 1 & (with $C_d\to0$)\\
$k_d$ & depression seepage coefficient (d$^{-1}$) & $10^{-4}$--10 & log & 0.01 & (with $C_d\to0$)\\
$J$ & exterior inflow (mm\,d$^{-1}$) & $10^{-4}$--20 & log & $10^{-3}$ & $J=0$ (outside)\\
$L_c$ & mean channel loss capacity (mm\,d$^{-1}$) & $10^{-3}$--5 & log & $3\times10^{-3}$ & $L_c=0$ (outside)\\\bottomrule
\end{tabular}
\end{table}

\begin{table}[h]\centering\footnotesize\setlength\tabcolsep{4pt}
\caption{Global constants of \CaRTp{} (learned, one value for all catchments of a dataset; initial value and range of the trained values over three seeds) and fixed constants of both models.}\label{tab:glob}
\begin{tabular}{@{}llrlll@{}}\toprule
Symbol & Meaning & Initial & CAMELS-GB & CAMELS-US & CAMELS-AUS\\\midrule
$T_F$ & centre of the frost step (\si{\celsius}) & 0 & $-0.28$ to $0.07$ & $-0.11$ to $0.07$ & $-0.04$ to $-0.03$\\
$s_F$ & width of the frost step (\si{\celsius}; log) & 2.4 & 2.25--2.31 & 2.17--2.38 & 2.41--2.42\\
$\theta_c$ & centre of the connection thresholds & 0.4 & 0.49--0.52 & 0.54--0.61 & 0.67--0.70\\
$s_\theta$ & width of the connection thresholds (log) & 0.15 & 0.071--0.082 & 0.048--0.069 & 0.047--0.058\\
$T_v$ & activity threshold of vegetation (\si{\celsius}) & 8 & 8.14--8.22 & 7.67--7.98 & 8.27--8.37\\
$s_v$ & width of the activity threshold (\si{\celsius}; log) & 3 & 4.80--5.18 & 2.85--3.15 & 9.33--12.22\\
$\tau_v$ & time constant of vegetation activity (d; log) & 20 & 4.9--5.9 & 13.3--14.8 & 114--147\\\bottomrule
\end{tabular}\\[6pt]
\begin{tabular}{@{}p{15.4cm}@{}}
\emph{Fixed constants.} Number of bands $K=5$; release exponent 5 and exchange exponent 3.5 (GR4J); split 0.9/0.1 of direct runoff and exchange; $K_{p,\rm GR4J}=C(4/9)^4/4$; unit-hydrograph exponent 2.5 and lengths 6 and 13 days; bound $6\tanh(\cdot/6)$ of the release log-multiplier; viscosity constants 247.8\,K, 140\,K, reference \SI{20}{\celsius}, temperature clamped at \SI{0}{\celsius}; $\tau_\omega=3$\,d with a 36-day kernel; vegetation kernel 2048 days; lower bound 0.5 of $A_{\min}$; floor $10^{-3}$\,mm of $R_m$ in $f_t$; lower bound $F_{\min}$ of the infiltration capacity ($10^{-4}$\,mm\,d$^{-1}$ on CAMELS-AUS, 0 otherwise).
\end{tabular}
\end{table}

\subsection{Initial states and spin-up}\label{sec:S1spin}

Each store with a spin-up is first run over the first $n$ days of the record from the initial state in Table~\ref{tab:spin}, without gradient. The simulation then restarts on the first day of the record, using the final spin-up state as its initial state. The surface layer, the infiltration excess, the root zone, the near-stream zone and the wetness $\theta$ are spun up as a chain in which each store is driven by the spin-up output of the stores above it. The depressions and the quick store are spun up with the inputs of the simulation itself. With records starting on 3 December 1988 (CAMELS-GB) and 1 January 1980 (CAMELS-US and CAMELS-AUS), the training period (from 1 January 1989) is preceded by 29 days and by nine years of simulation, respectively.

\begin{table}[h]\centering\small
\caption{Initial states, with spin-up length $n$ in days (``--'': no spin-up).}\label{tab:spin}
\begin{tabular}{@{}lllc@{}}\toprule
State & Model & Initial state & $n$\\\midrule
snow $N_k$ & both & 0 & --\\
production store / root zone $S$ & both & $0.7C$ & 1460\\
surface layer $U$ & \CaRTp & $0.5C_u$ & 1460\\
near-stream zone $Z$ & \CaRTp & $0.3C_z$ & 1460\\
wetness $\Theta=C_\theta\theta$ & \CaRTp & $0.3C_\theta$ & 1460\\
depressions $V$ & \CaRTp & $0.5C_d$ & 1460\\
unit hydrographs $H$ & both & empty & --\\
quick store $R$ & both & $0.5X_3$ & 730\\
groundwater $W$ & both & $W_{\rm ss}=\bar g\tau$ (steady state at the training-mean recharge) & --\\
wetness index $\omega$ & \CaRTp & $\bar\omega$ at $t=0$ (Eq.~\ref{eq:omega}) & --\\
vegetation activity $v$ & \CaRTp & $\bar\sigma$ (Eq.~\ref{eq:veg}) & --\\\bottomrule
\end{tabular}
\end{table}

\subsection{Static-attribute network and training}\label{sec:S1net}

\paragraph{Inputs.} CAMELS-GB uses 21 attributes, standardised with the mean and standard deviation over the 604 catchments that are not held out as ungauged. The attributes are area, median elevation, sand, silt and clay fractions, porosity, hydraulic conductivity, soil depth, deciduous and evergreen woodland, crop and urban fractions, reservoir capacity, mean precipitation, mean PET, precipitation seasonality, snow fraction, and the frequency and duration of high- and low-precipitation days. CAMELS-US uses the 27 attributes of \citet{kratzert2019hess} (climate indices, mean elevation and slope, area, forest fraction, leaf-area and green-vegetation indices, soil depth, porosity, conductivity, water content and texture, carbonate fraction, permeability), standardised over the 671 basins. CAMELS-AUS uses 27 attributes that cover the same properties and are standardised over the 561 catchments, with the seasonal amplitude of net primary productivity in place of that of leaf area. The band elevations are used only by the snow store. During training, Gaussian noise of standard deviation 0.1 is added to every standardised attribute of every catchment, drawn anew at each iteration, and no noise is used for selection or evaluation.

\paragraph{Architecture.} The network has three fully connected hidden layers of 128 units with the (exact, error-function) GELU activation $\phi$. The hidden representation is $\bm h=\phi(\bm W_3\phi(\bm W_2\phi(\bm W_1\bm a+\bm b_1)+\bm b_2)+\bm b_3)$ and is followed by linear heads and the logistic map \eqref{eq:map}. \CaRT{} has one head of 11 outputs, whereas \CaRTp{} has one head of 28 outputs (rain--runoff, snow and the refinements other than the quick release) and another one of 7 (the quick-release parameters $A_{\min},C_\theta,\tau_\theta,\lambda,\lambda_W,k,a_F$). The heads are initialised with zero weights and with biases that give the initial values, while the hidden layers use the default uniform initialisation of the framework. For \CaRT{} the network has \num{37259} weights (CAMELS-GB) and \num{38027} (CAMELS-US and CAMELS-AUS, which have the same number of attributes), and for \CaRTp{} \num{40355} and \num{41123}, in addition to the seven global constants. The parameters are static, computed once per catchment and applied throughout the record.

\paragraph{Loss and optimisation.} Each iteration simulates all catchments jointly over the record from its first day to 31 December 2010. The loss is the mean of $1-\mathrm{NSE}$ over the gauged catchments with more than 730 observed days in the training period. Each NSE is computed over the observed days from 1 January 1989 to 31 December 2008. The optimiser is AdamW ($\beta=(0.9,0.999)$, $\epsilon=10^{-8}$, weight decay $10^{-4}$ on all trained quantities, including the global constants). It uses a one-cycle learning-rate schedule with maximum $2\times10^{-3}$, 5\,\% warm-up, cosine annealing, initial and final rates $2\times10^{-3}/25$ and $2\times10^{-3}/(25\times10^4)$, and the momentum cycling of the schedule ($\beta_1$ between 0.95 and 0.85). Training runs for 5000 iterations, each of which is one full-batch step, with gradients clipped to a global norm of 1 and steps skipped if the loss or gradient is non-finite. The steady storage $W_{\rm ss}$ is treated as a constant in the gradient (stop-gradient), in the initial state and wherever it enters the daily update (in the half-saturation of the uptake and in $W_t/W_{\rm ss}$ of the connected fraction). The gradient passes through the training-mean recharge $\bar g$ into $a_g=1/(\bar g\tau^2)$. Three seeds (0, 1, 2) set the initial weights and the attribute noise.

\paragraph{Selection and periods.} Every 50 iterations and at the last iteration the model is evaluated without noise. We keep the network with the largest median NSE over 1 January 2009 to 31 December 2010, taken over the gauged catchments with more than 365 observed days in that period. The selected network is then run over the whole record, with the same periods on all three datasets (training 1989--2008, selection 2009--2010 and test 2011--2019). In each dataset, 10\,\% of the catchments (67 on CAMELS-GB, 68 on CAMELS-US and 56 on CAMELS-AUS) are ungauged and, although simulated, never enter the loss or the selection. Two further CAMELS-GB catchments are excluded because they lack usable discharge.

\paragraph{Implementation.} No operation feeds back on an earlier one in this sequence, so the models are computed stage by stage over the whole record, and the result is identical to that of a daily loop in the order of Section~\ref{sec:S1order}. However, the groundwater coefficient depends on the mean recharge $\bar g$ over the training period (1 January 1989 to 31 December 2008) of the same simulation (Section~\ref{sec:S1cart}). A daily-loop implementation therefore needs a first pass of steps (1)--(8) (\CaRT: (1)--(2)) to obtain $\bar g$. The lower bound $F_{\min}$ of the infiltration capacity on CAMELS-AUS prevents an underflow of the capacity in single precision in the driest catchments, which made the gradients undefined. All daily recursions of the stores are implemented as fused GPU kernels (Triton) with adjoint recursions run backwards over the stored states. These recursions are either hand-written or generated from the one-day step by automatic differentiation of a single day. Every kernel was verified against float64 automatic differentiation of the unrolled loop (the largest relative gradient error in the maximum norm was below $3\times10^{-5}$ over the tested sequences and states). The fused kernels made the ablations 26--40 times faster than a compiled daily loop. The vegetation activity and the near-surface wetness, which are moving averages with kernels of 2048 and 36 days, are computed as convolutions in PyTorch, the former by FFT. Below $x=10^{-3}$, $1-e^{-x}$ is evaluated as $x-x^2/2$ in the infiltration excess and the channel loss and as $x-x^2/2+x^3/6$ in the groundwater uptake.

\subsection{Water-balance bookkeeping}\label{sec:S1bal}

Over the whole simulation of each catchment, the water that enters the model equals the water that leaves it plus the change in storage,
\begin{equation}\label{eq:bal}
\begin{aligned}
&\textstyle\sum P+\sum(\mathrm{SFCF}-1)\,P_{\rm snow}+\sum J\\
&\qquad=\textstyle\sum Q+\sum\big(u+e_u+{\rm AET}_{sz}+e_d+e_W\big)+\sum\ell-\sum X\\
&\qquad\quad+\Delta\Big(\textstyle\sum_ka_kN_k+U+(1-w)S+wZ+V+H+R+W\Big).
\end{aligned}
\end{equation}
The left side is the input, which besides the precipitation contains the snowfall added by the correction factor SFCF, with $P_{\rm snow}=\sum_ka_kc_kP$ the uncorrected snowfall, and the exterior inflow $J$. The right side contains the discharge, the losses to the atmosphere (sublimation, evaporation from the surface layer, the soil and the depressions, and uptake from groundwater), the channel loss $\ell$, the exchange loss $-X\ge0$ and the change in storage of all stores, including the water $H$ in transit in the unit hydrographs. The exchange enters as the amount actually removed, $X=(R_m-R-x_9-0.9J)+(Q_d-x_1-0.1J)$, which is smaller in magnitude than $j$ on the days a clamp acts (Section~\ref{sec:S1order}). The part of $j$ that the clamp removes is not taken from any store. For \CaRT{} the terms of the refinements vanish and ${\rm AET}_{sz}$ is replaced by $e_s+I$. The conditional variables $\theta$, $\omega$ and $v$, and the factors computed from them, carry no water and add no flow path, so they do not appear in the balance. For the final \CaRTp{} runs on CAMELS-GB and CAMELS-US (three seeds each), the residual of this balance divided by the accumulated precipitation is below $1.5\times10^{-14}$ in every catchment in float64 and below $10^{-5}$ in float32.

\section{Derivation of the refinements}\label{sec:S2}

For each refinement we state the fine-scale assumption, the derivation, the resulting law, the limit in which it vanishes and the origin of the derivation. The origin uses the categories of Section~4 and Table~4 of the main text.
\begin{itemize}
\item \emph{exact}: the aggregate of the stated fine-scale assumptions;
\item \emph{asymptotic}: an aggregate that holds only in a limit;
\item \emph{postulated}: a form chosen by analogy with a known law, including a new state and its balance;
\item \emph{physical}: the form of a known physical law;
\item \emph{heuristic}: a functional form with the expected sign and bounds.
\end{itemize} The operations (a)--(d) named below are those of Corollary~1 of the main text, and Table~\ref{tab:summary} summarises the results.

\subsection{Connected fraction of the quick release and the wetness $\theta$}\label{sec:S2conn}

\paragraph{Assumptions.} (A1) The hillslope consists of units $i$ with area weights $a_i$ that share one quick-flow depth $R$ (lateral redistribution is faster than release to the channel). (A2) A unit releases at the GR4J rate $a_fR^5$ only when it is connected, with indicator $\mathbf 1[\theta_i>\theta_{c,i}]$, where $\theta_i$ is its wetness and $\theta_{c,i}$ its connection threshold. A fraction $A_{\min}$ next to the channel is always connected, and the thresholds of the other units have the area distribution $F_\theta$. (A3) Where the forcing or the units vary in space, neighbouring units exchange wetness faster than the balance of A5 relaxes, and the exchange conserves the area-weighted mean of the wetness. (A4) The threshold distribution is logistic with centre $\theta_c$ and width $s_\theta$, and a wetter root zone lowers its centre linearly in $s$. (A5) The wetness of each unit obeys the balance of a production store with a linear loss, under the input $P_n$ and demand $E_n$ that reach the root zone and are shared by all units. The wetness carries no water.

\paragraph{Derivation.} By A5 the wetness of the units converges to its mean $\theta$, with or without the exchange of A3 (Appendix C, step 3). On the resulting slow manifold, by A1--A2, the release is $q_r=\sum_ia_iq_i=a_fR^5\big[A_{\min}+(1-A_{\min})\sum_{i}a'_i\mathbf 1[\theta>\theta_{c,i}]\big]$, where $a'_i=a_i/(1-A_{\min})$ are the weights within the non-permanent part, which sum to 1, and the indicator uses the common wetness $\theta_i=\theta$. The sum is the fraction of the area whose threshold lies below $\theta$, which is the area distribution of the thresholds at $\theta$, $F_\theta(\theta)$:
\begin{equation}\label{eq:S2A}
q_r=a_fA\,R^5,\qquad A=A_{\min}+(1-A_{\min})F_\theta(\theta),\qquad m_{\rm conn}=\ln A\in[\ln A_{\min},0].
\end{equation}
Here $m_{\rm conn}$ is the connectivity term of the log-multiplier $m_t$ of Eq.~\eqref{eq:m}. With A4 and $\lambda\ge0$, $F_\theta(\theta)=\sg((\theta-\theta_c+\lambda s)/s_\theta)$, which is Eq.~\eqref{eq:A} without the groundwater term (Section~\ref{sec:S2gwconn}). On this manifold A5 yields Eq.~(17) of the main text for the mean,
\begin{equation}\label{eq:S2theta}
C_\theta\dot\theta=P_n(1-\theta^2)-E_n\theta(2-\theta)-C_\theta\theta/\tau_\theta .
\end{equation}

\paragraph{Memory form.} The model integrates \eqref{eq:S2theta} as it stands, while its linearisation about the equilibrium under the mean forcing shows what kind of memory $\theta$ carries and on what time scale. Let $\bar\theta$ be the equilibrium of \eqref{eq:S2theta} under the mean forcing $(\bar P_n,\bar E_n)$. To first order in $\delta\theta=\theta-\bar\theta$, $\delta P_n=P_n-\bar P_n$, $\delta E_n=E_n-\bar E_n$,
\begin{equation}\label{eq:S2lin}
\dot{\delta\theta}=-\frac{\delta\theta}{\tau_{\rm eff}}+\frac{(1-\bar\theta^2)\,\delta P_n-\bar\theta(2-\bar\theta)\,\delta E_n}{C_\theta},\qquad
\frac1{\tau_{\rm eff}}=\frac1{\tau_\theta}+\frac{2\bar\theta\bar P_n+2(1-\bar\theta)\bar E_n}{C_\theta},
\end{equation}
whose solution is
\begin{equation}\label{eq:S2linsol}
\delta\theta(t)=\delta\theta(0)\,e^{-t/\tau_{\rm eff}}+\frac{\tau_{\rm eff}}{C_\theta}\int_0^t\frac{e^{-(t-s)/\tau_{\rm eff}}}{\tau_{\rm eff}}\,f(s)\,\dd s,\qquad f=(1-\bar\theta^2)\,\delta P_n-\bar\theta(2-\bar\theta)\,\delta E_n,
\end{equation}
so that $\delta\theta(t)$ is $\tau_{\rm eff}/C_\theta$ times an exponential moving average of the weighted net-input anomaly with time constant $\tau_{\rm eff}$, plus the decaying initial anomaly. At $\bar\theta=\tfrac12$ both weights are $\tfrac34$ and the driver is the anomaly of $P_n-E_n$. $\tau_{\rm eff}$ is set by $C_\theta$, $\tau_\theta$ and the mean forcing. Because $\bar\theta$ moves with the mean forcing, differentiating $1/\tau_{\rm eff}$ along the equilibrium of \eqref{eq:S2theta}, with $\dd\bar\theta/\dd\bar P_n$ from the equilibrium condition, gives
\begin{equation}\label{eq:S2taueff}
\frac{\dd(1/\tau_{\rm eff})}{\dd\bar P_n}=\frac{2\,\tau_{\rm eff}\,(2\bar P_n-\bar E_n)}{C_\theta^2},
\end{equation}
so $\tau_{\rm eff}$ shortens with $\bar P_n$ where $2\bar P_n>\bar E_n$ and lengthens with it where $2\bar P_n<\bar E_n$, that is, in drier climates. Away from the mean the instantaneous relaxation rate is $1/\tau_\theta+2P_n\theta/C_\theta$ on days without demand and $1/\tau_\theta+2E_n(1-\theta)/C_\theta$ on days with demand. It increases with $\theta$ while the hillslope wets up and decreases with $\theta$ while it dries. This moving-average form is the memory form of Law~3 of the main text, with $\theta$ the conditional variable and $A=e^{m_{\rm conn}}$ a bounded multiplier.

\paragraph{Law, limit and origin.} $q_r=a_fA_tR^5$ with $A_t$ of Eq.~\eqref{eq:A}, under operations (c) and (d). $A_{\min}\to1$ recovers $A\equiv1$ and the release of \CaRT. The sum over units \eqref{eq:S2A} is exact on the slow manifold of A5, whereas the logistic shape, the shift $\lambda s$ and the balance \eqref{eq:S2theta} are postulated (the balance by analogy with the production store). The moving-average form is the linearisation of \eqref{eq:S2theta}. Under a weak spatial pattern of the input and demand, or with units that differ, the exchange of A3 keeps the deviations small. These deviations shift the thresholds and add terms to the mean balance that the fitted parameters absorb as an empirical closure (Appendix~C, step~3). The construction draws on work on the threshold connectivity of hillslopes and channels \citep{jencso2009,bracken2007,western2001,tromp2006}, on variable source areas \citep{hewlett1967,dunne1970,bevenkirkby1979} and on memory from the elimination of states \citep{mori1965,zwanzig1973,chorin2002}.

\subsection{Release-shape exponent $k$}\label{sec:S2k}

\paragraph{Assumptions.} (B1) A unit releases only the quick-flow depth above a local depth threshold $x$, with the local GR4J law $a(R-x)_+^5$. (B2) The area distribution of the thresholds is a power law at small depth, $F_x(x)=c\,x^{\nu}$ with $\nu>0$, or a Dirac distribution at $x=0$ ($\nu=0$), so that every unit releases from the first millimetre of storage, over the range of $R$ considered.

\paragraph{Derivation.} For $\nu>0$, with $\dd F_x=c\,\nu\,x^{\nu-1}\dd x$ and the substitution $x=Ru$, where $u$ is the threshold as a fraction of the storage,
\begin{equation}\label{eq:S2k}
\begin{aligned}
\bar q&=a\int_0^R(R-x)^5\,\dd F_x=a\,c\,\nu\int_0^R(R-x)^5x^{\nu-1}\dd x\\
&=a\,c\,\nu\,R^{5+\nu}\int_0^1(1-u)^5u^{\nu-1}\dd u=\frac{5!\,a\,c\,R^{5+\nu}}{(\nu+1)(\nu+2)(\nu+3)(\nu+4)(\nu+5)},
\end{aligned}
\end{equation}
where the integral over $u$ follows from five integrations by parts. The exponent is $5+k$ with $k=\nu$, and the prefactor is absorbed into the scale $X_3$ through $a_f=1/(4X_3^4)$ (Eq.~\eqref{Seq:quick}), giving $q_r=a_fe^{m}R^5(R/X_3)^k$, which at $R=X_3$ equals the release with $k=0$ for any $k$ (Section~4.2 of the main text). As $\nu\to0$ the distribution tends to a Dirac distribution at $x=0$, which corresponds to $k=0$. If the thresholds have an upper end $x_{\max}<R$, the exponent falls back towards 5 for $R\gg x_{\max}$, because $(R-x)^5\approx R^5$ for all thresholds $x\le x_{\max}$.

\paragraph{Law, limit and origin.} Under operation (c), $k=0$ is the release of \CaRT, and the range $k\ge0$ corresponds to a power-law threshold distribution. The law is an asymptotic aggregate, which is exact while $R$ lies in the power-law range of $F_x$.

\subsection{Viscosity and frost}\label{sec:S2visc}

\paragraph{Assumptions.} (C1) The conductance of the quick pathway is inversely proportional to the dynamic viscosity of water, $\mu(T_w)$ (laminar flow, Law~2 of the main text). (C2) The water temperature equals the daily mean air temperature but does not fall below \SI{0}{\celsius}. (C3) On a frozen share of the area, which switches on near \SI{0}{\celsius}, frost raises all depth thresholds of B1 by a frozen depth $d_f$ of a few millimetres.

\paragraph{Derivation.} (C1--C2) give $m_{\rm visc}=-\ln[\mu(T)/\mu(\SI{20}{\celsius})]$. With the Vogel-type correlation $\mu\propto10^{B_V/(T_K-C_V)}$, where $B_V=247.8$\,K and $C_V=140$\,K, this is Eq.~\eqref{eq:m}. Between 0 and \SI{20}{\celsius} the multiplier changes by 0.56 in $m$. The correlation agrees with the reference values of \citet{huber2009} to within 2.2\,\% at \SI{0}{\celsius} and to 0.5\,\% above \SI{10}{\celsius}. (C3) With B1--B2 and a frozen share $\phi_F$ of the area, $\bar q=\bar q_0\big[(1-\phi_F)+\phi_F(1-d_f/R)^{5+\nu}\big]$, because on the frozen share $\bar q\propto(R-d_f)^{5+\nu}$ as in Eq.~\eqref{eq:S2k} with $R-d_f$ in place of $R$. For $R\gg d_f$ this gives $\ln(\bar q/\bar q_0)\approx-\phi_F(5+\nu)\,d_f/R$, so frost lowers the release most strongly at small storage. To simplify, the model replaces this storage-dependent reduction by the storage-independent step $-a_F\sg((T_F-T)/s_F)$ with $a_F\ge0$. The logistic term stands for the frozen share $\phi_F$ and switches with the same-day air temperature, and $a_F$ stands for a mean of $(5+\nu)\,d_f/R$.

\paragraph{Law, limit and origin.} For operation (d) (Eq.~\eqref{eq:m}), the viscosity term has no free parameter and no nested limit, while $a_F=0$ removes the frost term. The viscosity term follows a physical law \citep{huber2009,constantz1982}, whereas the frost step is a heuristic form of the frozen-depth argument and is represented by a storage-independent step without memory of the frost history \citep{stahli1999,shanley1999,bayard2005,iwata2010}.

\subsection{Near-stream saturated zone}\label{sec:S2zone}

\paragraph{Assumptions.} (Z1) A fraction $w$ of the area next to the channel has a small storage capacity $C_z$ and a water table close to the surface. (Z2) Its partition, evaporation and storage follow the laws of the production store with $z=Z/C_z$. (Z3) It drains linearly to the outlet with time constant $\tau_z$ and does not percolate. (Z4) It receives the same input and demand per unit area as the root zone.

\paragraph{Derivation.} The law is Z1--Z4 written as equations per unit area of the zone
\begin{equation}\label{eq:S2zone}
\dot Z=P_n(1-z^2)-E_nz(2-z)-Z/\tau_z,\qquad d_z=P_nz^2,\quad e_z=E_nz(2-z),\quad q_z=Z/\tau_z,
\end{equation}
integrated jointly (input and drainage together) by the exact daily solutions \eqref{eq:ric}--\eqref{eq:bern} (Section~\ref{sec:S2daily}). The family contains three limits. For $\tau_z\to0$ it is an impervious area that drains at once, for $C_z\to\infty$ at fixed storage $Z$ a linear interflow store, and for $\tau_z\to\infty$ a saturated area that varies with wetness, that is, the variable source area of \citet{dunne1970}. The allowed ranges, $\tau_z\in(2,120)$\,d and $C_z\in(5,300)$\,mm, lie between these limits.

\paragraph{Law, limit and origin.} Operations (a) and (b) make a parallel split of the soil volume with a new drainage to the outlet. $w\to0$ recovers the model without the zone, which is itself postulated by analogy with the production store \citep[cf.\ the probability-distributed store of][]{moore1985}.

\subsection{Surface layer}\label{sec:S2top}

\paragraph{Assumptions.} (T1) A thin layer $U$ (canopy, litter and top soil) evaporates at the demand rate, as interception and stage-one soil evaporation \citep{rutter1971,ritchie1972}. (T2) Water above its capacity $C_u$ passes the same day to the stores below. (T3) The rest drains linearly into the root zone with time constant $\tau_u$.

\paragraph{Derivation.} The law is T1--T3 written as equations for the layer. Eq.~\eqref{eq:top} applies evaporation, overflow and drainage in this order within the day. The drainage is the exact one-day solution $k_uU''$, with $k_u=1-e^{-1/\tau_u}$, of $\dot U=-U/\tau_u$ without other fluxes.

\paragraph{Law, limit and origin.} Merging $U$ and $S$ recovers the columns of \CaRT{} (operation a, series split). For $\tau_u\to0$, $k_u\to1$ and the layer empties daily, so from the second day on (once the initial content has drained) $x=(P_L-E_{\rm dem})_+$ and $\ein=(E_{\rm dem}-P_L)_+$, which corresponds to the same-day interception $\min(P,E)$ of \CaRT. This holds for any $C_u$ because the overflow also passes to the stores below. The layer is postulated \citep[cf.\ the interception store of][]{ficchi2019}.

\subsection{Pareto partition of the root zone}\label{sec:S2pareto}

\paragraph{Assumptions.} (P1) Point storage capacities $c$ have the distribution $F_c(c)=1-(1-c/c_{\max})^b$ on $[0,c_{\max}]$ with mean $C$, so $c_{\max}=(b+1)C$. (P2) Net rain is uniform in space. (P3) There is no lateral redistribution within the day, so all points with capacity below a critical capacity $C^*$ are full, and all remaining points hold $C^*$.

\paragraph{Derivation.} The storage is
\begin{equation}
S(C^*)=\int_0^{c_{\max}}\min(c,C^*)\,\dd F_c=\int_0^{C^*}\big(1-F_c(c)\big)\dd c=C\Big[1-\big(1-C^*/c_{\max}\big)^{b+1}\Big],
\end{equation}
so $\delta\equiv1-S/C=(1-C^*/c_{\max})^{b+1}$. A net rain depth $P_n$ raises $C^*$ by $P_n$, up to $c_{\max}$, and the water that enters the store is
\begin{equation}\label{eq:S2pareto}
p_s=S(C^*+P_n)-S(C^*)=C\Big[\delta-\big(\delta^{1/(b+1)}-P_n/c_{\max}\big)_+^{\,b+1}\Big],
\end{equation}
which is Eq.~(9) of the main text, with $P_n-p_s$ as saturation excess. The configuration of P3 is preserved by filling but not by evaporation between events, so \eqref{eq:S2pareto} is exact for a filling step from a state of that form. The infiltrating fraction is $\dd S/\dd C^*=1-F_c(C^*)$, the area of the points that are not yet full, which for the Pareto distribution is $(1-C^*/c_{\max})^b=\delta^{b/(b+1)}$. The GR4J partition belongs to the same class with $F_c(c)=\tanh^2(c/C)$. Then $1-F_c=\mathrm{sech}^2(c/C)$, $S=C\tanh(C^*/C)$, and the infiltrating fraction $1-F_c(C^*)$ is $1-s^2$, which is the integrand of Eq.~\eqref{eq:prod}. The two partitions coincide at $s=\tfrac12$ for $b\approx0.71$, where $\delta^{b/(b+1)}=0.5^{b/(b+1)}=0.75$ equals $1-s^2$.

\paragraph{Law, limit and origin.} Under operation (c) the refinement keeps the evaporation and percolation of GR4J (Eq.~\ref{Seq:pareto}). No value of $b$ produces $1-s^2$, so the refinement does not nest \CaRT. However, filling is exact under P1--P3 \citep{moore1985,moore2007,zhao1992}.

\subsection{Infiltration excess}\label{sec:S2ie}

\paragraph{Assumptions.} (I1) The water $x$ that leaves the surface layer on a day is uniform in space and over the day. (I2) The local daily infiltration capacity (rate times duration) is exponentially distributed in space with mean $F$ \citep{hawkins1987}. (I3) Input above the local capacity becomes direct runoff without run-on \citep{horton1933,philip1957}. (I4) A near-surface wetness $m$ that is shared by all points scales every local capacity by the same factor $g(m)$, with $g(0)=1$ and $P^*$ the mean capacity at zero wetness. (I5) At every point $m$ obeys the linear balance $\tau_\omega\dot m=x-m$, driven by the water $x$ that leaves the surface layer before the infiltration excess is removed, with $\tau_\omega=3$\,d. (I6) $g(m)=\exp(-g_Fm/\bar\omega)$, with the sensitivity $g_F\ge0$ learned per catchment and the scale $\bar\omega$ of Section~\ref{sec:S1cartp} (4).

\paragraph{Derivation.} By I1--I3,
\begin{equation}\label{eq:S2ie}
i_e=\int_0^\infty(x-f)_+\,\frac{e^{-f/F}}{F}\,\dd f=x\big(1-e^{-x/F}\big)-F\Big[1-e^{-x/F}\big(1+x/F\big)\Big]=x-F\big(1-e^{-x/F}\big),
\end{equation}
which is quadratic in $x$ for small input ($i_e\approx x^2/2F$) and linear for large input ($i_e\approx x-F$). Under I4 the capacities stay exponentially distributed with mean $F=P^*g(m)$, so the same expression holds. Under I5 points with the same initial wetness keep a common $m$, and different initial values converge at the rate $1/\tau_\omega$ (Appendix~C, step~3). With $x$ held fixed over each day, the daily solution of I5 for the start-of-day state is $m_t=\alpha m_{t-1}+(1-\alpha)x_{t-1}$ with $\alpha=e^{-1/\tau_\omega}$, which is Eq.~\eqref{eq:omega} with $m=\omega$ up to its truncation at 36 days.

\paragraph{Law, limit and origin.} The refinement is operation (c), with I4 a modulation of type (d), and $P^*\to\infty$ removes the excess. The result is exact under I1--I5, with the balance I5 postulated and the scale function I6 heuristic.

\subsection{Vegetation demand}\label{sec:S2veg}

\paragraph{Assumptions.} (V1) The vegetation consists of patches $i$ with area weights $a_i$ and temperature thresholds $T_{v,i}$ whose area distribution is $F_v(T)=\sg((T-T_v)/s_v)$, with centre $T_v$ and width $s_v$ shared by all catchments of a dataset. (V2) A patch becomes active at the rate $1/\tau_v$ while the daily mean air temperature $T$ exceeds its threshold and dormant at the same rate while $T$ lies below it. The temperature is uniform over the catchment and constant within the day. (V3) Active patches have a demand of $E$ and dormant ones $(1-\kappa)E$, with $\kappa\le1$ the share of the demand that depends on the vegetation.

\paragraph{Derivation.} By V2 an inactive patch becomes active at the rate $1/\tau_v$ while $T>T_{v,i}$ and an active one becomes dormant at that rate while $T\le T_{v,i}$, so the probability $p_i$ that patch $i$ is active obeys $\tau_v\dot p_i=\mathbf 1[T>T_{v,i}](1-p_i)-\mathbf 1[T\le T_{v,i}]\,p_i$. This is the linear equation $\tau_v\dot p_i=\mathbf 1[T>T_{v,i}]-p_i$, so the area average $v=\sum_ia_ip_i$ obeys $\tau_v\dot v=F_v(T)-v$ exactly. With $\sigma_t=F_v(T_t)$ fixed over day $t$ the exact one-day solution is $v_t=e^{-1/\tau_v}v_{t-1}+(1-e^{-1/\tau_v})\sigma_t$, which Eq.~\eqref{eq:veg} implements from $v=\bar\sigma$ before the first day, up to the truncation of its kernel at 2048 days. By V3 the area-averaged demand is $E_{\rm dem}=E[v+(1-\kappa)(1-v)]=E[1-\kappa(1-v)]$, and it is evaluated with the activity $v_t$ at the end of the day.

\paragraph{Law, limit and origin.} Under operation (d) a conditional variable modulates the boundary fluxes to the atmosphere, with the limit $\kappa=0$. The aggregate is exact under V1--V3, and the common rate for activation and dormancy closes it in $v$. The switching law and the logistic distribution are postulated and are motivated by the seasonal lag of transpiration behind air temperature \citep{oudin2005,chuine2000}.

\subsection{Groundwater uptake}\label{sec:S2uptake}

\paragraph{Assumptions.} (U1) Riparian vegetation meets a share $\lambda_E\le1$ of the residual demand $E_r$ from groundwater \citep{loheide2005}. (U2) The uptake saturates with storage, with half-saturation at the uptake-free steady storage $W_{\rm ss}$.

\paragraph{Derivation.} The law is U1--U2 written as equations for the uptake, $e_W=\lambda_EE_rW/(W+W_{\rm ss})$, which is integrated as in Eq.~\eqref{eq:uptake}. For the quadratic store, $\dd\ln q_s/\dd t=2\dot W/W$, so the uptake adds $2e_W/W=2\lambda_EE_r/(W+W_{\rm ss})$ to the recession rate of baseflow. Recessions therefore steepen with evaporative demand, as observed in dry-weather low flows \citep{federer1973,wittenberg1999}.

\paragraph{Law, limit and origin.} The uptake is a new boundary flux from $W$ to the atmosphere under operation (b). $\lambda_E=0$ recovers the quadratic store exactly, while the saturating form is heuristic.

\subsection{Connected bypass}\label{sec:S2bypass}

\paragraph{Assumptions.} The assumptions of Section~\ref{sec:S2conn} apply, together with:
\begin{itemize}
\item[(R1)] Surface runoff produced on a connected unit reaches the channel directly with probability $\gamma$ and otherwise enters the unit's quick-flow layer.
\item[(R2)] Surface runoff on a disconnected unit re-infiltrates downslope into the quick path \citep{bracken2007}.
\item[(R3)] Runoff production is independent of the connection thresholds.
\end{itemize}

\paragraph{Derivation.} The bypass is $\sum_ia_id_i\,\gamma\,\mathbf 1[\text{unit $i$ connected}]=\gamma\,\mathbb E[d\,\mathbf 1]=\gamma\bar d\,A_t$ under R3, which is a fraction $\gamma A_t$ of the hillslope runoff, while the always-connected near-stream zone has fraction $\gamma$. In \CaRTp{} the runoff of all sources is mixed in the depressions before routing, and the fraction $\gamma A_t$ is applied to the whole spill. This choice treats the runoff of the near-stream zone and the infiltration excess as hillslope runoff, so the implemented split is exact for the hillslope part under R1--R3.

\paragraph{Law, limit and origin.} For operation (d), the limit $A\equiv1$ with $\gamma=0.1$ recovers the GR4J split. Because $A_t$ is shared with the release, this limit recovers the preceding model (variable $A$ in the release, fixed 0.1 bypass) only where the release is fully connected, so the refinement is not strictly nested. The split is exact for the hillslope runoff under R1--R3 and is applied to all runoff.

\subsection{Depression storage}\label{sec:S2dep}

\paragraph{Assumptions.} (Dp1) Direct runoff of all sources first fills surface depressions whose capacities follow the Pareto distribution of P1 with mean $C_d$ and shape $b_d$. (Dp2) Fill and spill follow P2--P3. (Dp3) Water held in depressions evaporates at the residual demand with the evaporation ratio of the production store. (Dp4) It seeps to the groundwater store with the percolation law of the production store and the coefficient $k_d$.

\paragraph{Derivation.} The spill is $D-p_d$ with $p_d$ given by Eq.~\eqref{eq:S2pareto} with $(C_d,b_d)$ and the input $D$, which is exact under Dp1--Dp2 for a filling step from a state of the form P3 (Section~\ref{sec:S2pareto}). By Dp3 the evaporation is the GR4J evaporation of Eq.~\eqref{eq:prod} with $s=1-\delta_d$ and the residual demand $E_d$, which yields $e_d=C_d(1-\delta_d^2)\,t_d/(1+\delta_d\,t_d)$. By Dp4 the seepage is the exact one-day solution of $\dot V=-(k_dC_d/4)(V/C_d)^5$ from $V'=V-e_d+p_d$, with $k_d$ in the role of $\kappa_q$ of the production store. Filling and evaporation are both evaluated from the start-of-day storage (Eq.~\eqref{eq:dep}).

\paragraph{Law, limit and origin.} Operations (a) and (b) apply, as the depressions are a series split between runoff generation and routing, the seepage is an internal flux to $W$, and the evaporation is a new boundary flux. $C_d\to0$ recovers the model without depressions. The spill is exact while the depressions fill, whereas the losses are postulated \citep{shook2011,moore1985}.

\subsection{Exterior inflow}\label{sec:S2exch}

\paragraph{Assumptions.} (X1) The flux across the catchment boundary is driven by the head difference with an exterior of fixed head, $j_{\rm ext}=L[H_{\rm ext}-g(R/X_3)]$, with the conductance $L\ge0$ (Law~2 of the main text). (X2) The head function is that of GR4J, $g(r)=r^{3.5}$.

\paragraph{Derivation.} The flux $j_{\rm ext}=-L\,g(R/X_3)+LH_{\rm ext}$ with $H_{\rm ext}=0$ is the exchange of \CaRT, namely $X_2(R/X_3)^{3.5}$ with $X_2=-L\le0$, and a positive exterior head adds the constant inflow $J=LH_{\rm ext}\ge0$. In the product form the inflow vanishes with the dissipative loss and has a saddle at $L=H_{\rm ext}=0$, so the inflow $J$ is parameterised separately from $X_2$. The total
\begin{equation}\label{eq:S2exch}
j_{\rm ext}=J+X_2\,(R/X_3)^{3.5},\qquad X_2\le0,\ J\ge0,\qquad \partial j_{\rm ext}/\partial R\le0,
\end{equation}
is a stabilising feedback, with inflow when the store is low and outflow when it is high \citep{lemoine2007,pushpalatha2011}.

\paragraph{Law, limit and origin.} The inflow uses the connection between $R$ and the exterior that the loss of \CaRT{} already has, so it changes the law of that flux (operation c). $J\to0$ recovers the exchange of \CaRT{}. The law takes the form of a physical law because the flux is driven by a head difference.

\subsection{Groundwater-controlled connectivity}\label{sec:S2gwconn}

\paragraph{Assumptions.} (G1) In chalk and limestone catchments the stream network expands when the water table rises above the beds of headwater reaches \citep{sear1999}. (G2) The connection thresholds of Section~\ref{sec:S2conn} are lowered linearly by the groundwater-head anomaly. The model's groundwater storage relative to its uptake-free steady value, $W/W_{\rm ss}-1$, stands for that anomaly.

\paragraph{Derivation.} The law is G2 written as a shift of the argument of $F_\theta$, which gains the term $\lambda_W(W_t/W_{\rm ss}-1)$ with $\lambda_W\ge0$ (Eq.~\ref{eq:A}), where $W_t$ is the end-of-day storage for that day.

\paragraph{Law, limit and origin.} The shift is operation (d), with $\lambda_W=0$ the model without it, and the linear shift is postulated.

\subsection{Snow store and melt energy term}\label{sec:S2snow}

\paragraph{Snow store.} A snowpack is a volume inserted on the precipitation (operation a) that loses water to the atmosphere through sublimation (operation b). Because snow accumulation has a temperature threshold, the spread of temperature with elevation is an $O(1)$ heterogeneity, and the store is split into five elevation bands with temperatures $T_k=T-\Gamma\Delta h_k$ (operation a). In each band the degree-day law of HBV \citep{bergstrom1995,seibert2012} accumulates snowfall below $T_T$ and melts $\cf(T_k-T_T)$ above it. Sublimation $u_k=\min(\kappa_sE,N_{k,m})$, with $N_{k,m}$ the pack after the day's snowfall, removes snow to the atmosphere at a fraction of the demand. This loss is absent from the degree-day store, and on the prairies it removes a large part of the pack as blowing snow \citep{pomeroy1993}. The store is postulated and has no nested limit because it is already present in \CaRT, whereas sublimation vanishes for $\kappa_s\to0$.

\paragraph{Melt energy term.} Assumptions. (M1) The degree-day factor represents the surface energy balance. (M2) In spring the melt energy rises with radiation faster than with air temperature. (M3) Being the only energy variable among the inputs, the potential evapotranspiration serves as an index of the available energy. Adding an energy term to the degree-day law gives $M_k=\min\{N_k',\cf(T_k-T_T)+r_pE\}$, with $N_k'$ the pack after sublimation, on days with $T_k\ge T_T$ (operation d), in the form of temperature-index models with a radiation term \citep{hock2003,kustas1994,pellicciotti2005}. For a given melt, a larger energy term lowers the temperature sensitivity of the store, because $\partial M_k/\partial T_k=\cf$ and the degree-day factor then carries a smaller part of the melt. $r_p\to0$ is the degree-day law. The energy argument M1--M2 is physical, while the additive form and the use of PET as the energy index are heuristic.

\subsection{Channel transmission loss}\label{sec:S2chan}

\paragraph{Assumptions.} (L1) The catchment consists of sub-areas whose flow per unit area equals the catchment-mean flow $Q_0$ before the loss on that day. (L2) The flow of each sub-area passes one loss site whose capacity $c$ (mm\,d$^{-1}$ per unit area) is exponentially distributed with mean $L_c$, and the site takes $\min(c,Q_0)$. (L3) There is no storage in the channel and no further loss downstream of a site.

\paragraph{Derivation.} The aggregate loss is
\begin{equation}\label{eq:S2chan}
\ell=\mathbb E[\min(c,Q_0)]=\int_0^{Q_0}\big(1-F_c(c)\big)\,\dd c=\int_0^{Q_0}e^{-c/L_c}\,\dd c=L_c\big(1-e^{-Q_0/L_c}\big)\le\min(Q_0,L_c),
\end{equation}
which takes nearly all of a small flow and at most $L_c$ of a large one. Under L1--L3 the loss sites act in parallel, whereas sites in series would be offered flows reduced by the losses upstream, and the aggregate would not have this form.

\paragraph{Law, limit and origin.} The loss is a flux from the outlet to the exterior \citep{shanafield2014}, a connection that the exchange of \CaRT{} already uses on the direct path, and its law follows by aggregation over thresholds (operation c). $L_c\to0$ is the model without it, and the result is exact under L1--L3.

\subsection{The exact one-day solutions}\label{sec:S2daily}

\paragraph{Power-law stores.} For $\dot V=-aV^p$, $p>1$, $V^{1-p}$ grows linearly in time, $V(1)^{1-p}=V_0^{1-p}+(p-1)a$, i.e.\ $V^+=V_0[1+(p-1)aV_0^{p-1}]^{-1/(p-1)}$. With $p=5$ this is the percolation of Eq.~\eqref{eq:prod} ($a=K_p/C^5$), the depression seepage and the quick release ($a=\cA_t$), while with $p=2$ it is the groundwater release, $1/W^+=1/W_m+a_g$.

\paragraph{Production-store partition.} On a day with input, with $\chi$ the net input accumulated since the start of the day, $\dd s/\dd \chi=(1-s^2)/C$ integrates to $s=\tanh(\chi/C+\operatorname{artanh}s_0)$, so with $t_P=\tanh(P_n/C)$, $s_1=(s_0+t_P)/(1+s_0t_P)$ and $C(s_1-s_0)=C(1-s_0^2)t_P/(1+s_0t_P)=p_s$. On a day with demand, with $\chi$ the demand accumulated since the start of the day, $\dd s/\dd \chi=-s(2-s)/C$, and with $u=1-s$, $\dd u/\dd \chi=(1-u^2)/C$, so with $t_E=\tanh(E_n/C)$, $s_1=s_0(1-t_E)/(1+(1-s_0)t_E)$ and $C(s_0-s_1)=Cs_0(2-s_0)t_E/(1+(1-s_0)t_E)=e_s$.

\paragraph{Wetness and near-stream zone.} On a day without demand, $\dot z=a-\beta z-az^2=-a(z-z_+)(z-z_-)$ with roots $z_\pm=(-\beta\pm\Delta)/(2a)$, $\Delta=\sqrt{\beta^2+4a^2}$, $z_+=2a/(\beta+\Delta)\in[0,1)$. The ratio $\rho=(z-z_+)/(z-z_-)$ obeys $\dot\rho=\dot z\,(z_+-z_-)/(z-z_-)^2=-a(z_+-z_-)\rho=-\Delta\rho$, since $a(z_+-z_-)=\Delta$, so $\rho_1=\rho_0e^{-\Delta}$, and $z_1-z_+=\rho_1(z_+-z_-)/(1-\rho_1)$ with $\rho_0(z_+-z_-)=(z_0-z_+)(1-\rho_0)$ reproduces Eq.~\eqref{eq:ric}. The identity $z_0-z_-=(2az_0+\beta+\Delta)/(2a)$ provides the cancellation-free $\rho_0$. Substituting $r=\rho_0e^{-\Delta t}$ gives $\int_0^1e^{-\Delta t}/(1-\rho_0e^{-\Delta t})\,\dd t=(\rho_0\Delta)^{-1}\ln[(1-\rho_0e^{-\Delta})/(1-\rho_0)]$. With $z(t)=z_++(z_+-z_-)\rho_0e^{-\Delta t}/(1-\rho_0e^{-\Delta t})$ this gives
\begin{equation}\label{eq:S2ricint}
\int_0^1z\,\dd t=z_++\frac{z_+-z_-}{\Delta}\ln(1+y)=z_++\frac{(z_0-z_+)(1-e^{-\Delta})}{\Delta}\,\psi(y),\qquad y=\frac{\rho_0(1-e^{-\Delta})}{1-\rho_0},
\end{equation}
using $\ln(1+y)=y\,\psi(y)$ and $\rho_0(z_+-z_-)=(z_0-z_+)(1-\rho_0)$, which is the integral of Eq.~\eqref{eq:ric}. On a day with demand, $\dot z=-\alpha_zz+ez^2$ (Bernoulli), and $y=1/z$ obeys $\dot y=\alpha_zy-e$, so $y(t)=e/\alpha_z+(y_0-e/\alpha_z)e^{\alpha_zt}$. With $\xi(t)=z_0(e/\alpha_z)(1-e^{-\alpha_zt})$ this gives $z(t)=1/y(t)=z_0e^{-\alpha_zt}/(1-\xi(t))$, which at $t=1$ is the state of Eq.~\eqref{eq:bern}. Since $\dot\xi=z_0e\,e^{-\alpha_zt}$, the relation $\dd\ln(1-\xi(t))/\dd t=-\dot\xi/(1-\xi)=-ez(t)$ gives
\begin{equation}\label{eq:S2bernint}
\int_0^1z\,\dd t=-\frac{\ln(1-\xi)}{e}=z_0\,\frac{1-e^{-\alpha_z}}{\alpha_z}\,\varphi(\xi),\qquad\xi=\xi(1),
\end{equation}
which is the integral in that equation. Both solutions keep $z$ in $[0,1]$ and agree with a fourth-order Runge--Kutta integration to $10^{-14}$.

\paragraph{Operator splitting.} Within a day the stores are advanced in sequence, with coefficients held at their value after the day's inflow (quick release, uptake) or at the start of the day (exchange, depressions). The quick-store update is the exact solution of the bounded law only for $k=0$ and $m_t$ constant over the day, whereas for $k>0$ it is the exact solution for the coefficient frozen at $R_m$. This regularised, operator-split model is the one that was trained and evaluated.

\begin{table}[h]\centering\footnotesize
\caption{Summary of the refinements of \CaRTp. Op.: operation of Corollary~1 of the main text. Nested limit: in the range (open bound) or outside it (residual effect at the bound in brackets).}\label{tab:summary}
\begin{tabular}{@{}>{\raggedright\arraybackslash}p{3.3cm}>{\raggedright\arraybackslash}p{0.9cm}>{\raggedright\arraybackslash}p{4.6cm}>{\raggedright\arraybackslash}p{5.6cm}@{}}\toprule
Refinement & Op. & Origin & Nested limit\\\midrule
Connected fraction $A(\theta)$ & c, d & exact on the slow manifold; logistic shape, shift and $\theta$ balance postulated & $A_{\min}\to1$, outside ($A_{\min}\le0.999$)\\
Release exponent & c & asymptotic & $k=0$, open bound\\
Viscosity & d & physical & none (no parameter)\\
Frost & d & heuristic & $a_F=0$, open bound\\
Near-stream zone & a, b & postulated & $w=0$, open bound\\
Surface layer & a & postulated & $\tau_u\to0$, outside ($k_u\le1-e^{-20}$)\\
Pareto partition & c & exact for filling & none\\
Infiltration excess & c, d & exact; wetness balance postulated, scale function heuristic & $P^*\to\infty$, outside ($i_e\approx x^2e^{g_F\omega/\bar\omega}/6000$)\\
Vegetation demand & d & exact; switching law and distribution postulated & $\kappa=0$, open bound\\
Groundwater uptake & b & heuristic & $\lambda_E=0$, open bound\\
Connected bypass & d & exact for hillslope runoff; applied to all runoff & $A\equiv1$, $\gamma=0.1$; not strictly nested\\
Depression storage & a, b & spill exact while filling; losses postulated & $C_d\to0$, outside ($C_d\ge0.02$\,mm)\\
Exterior inflow & c & physical & $J\to0$, outside ($J\ge10^{-4}$\,mm\,d$^{-1}$)\\
Groundwater connectivity & d & postulated & $\lambda_W=0$, open bound\\
Snow store (in \CaRT) & a, b & postulated & none (boundary volume); $\kappa_s\to0$ ($\ge10^{-4}$)\\
Melt energy term & d & heuristic; energy argument physical & $r_p\to0$, outside ($r_p\ge10^{-3}$)\\
Channel loss & c & exact & $L_c\to0$, outside ($L_c\ge10^{-3}$\,mm\,d$^{-1}$)\\\bottomrule
\end{tabular}
\end{table}
\end{document}